\pdfoutput=1

\documentclass[11pt]{article}

\usepackage{EMNLP2022}

\usepackage{times}
\usepackage{latexsym}

\usepackage[T1]{fontenc}

\usepackage[utf8]{inputenc}

\usepackage{microtype}

\usepackage{inconsolata}

\usepackage{graphicx}
\usepackage{mathtools}
\usepackage{multirow}
\usepackage{adjustbox}
\usepackage{subfigure}  
\usepackage{booktabs}
\usepackage[ruled,linesnumbered]{algorithm2e}
\usepackage{listings}
\usepackage{enumitem}
\newcommand*{\affaddr}[1]{#1} 
\newcommand{\uskg}{\textsc{Unified}SKG\xspace}
\newcommand{\skg}{SKG\xspace}
\newcommand{\ntasks}{21\xspace}  

\definecolor{amber}{rgb}{1.0, 0.75, 0.0}

\newcommand{\todo}[1]{\textcolor{purple}{[todo: #1]}}
\newcommand{\tao}[1]{\textcolor{blue}{[Tao: #1]}}

\newcommand{\running}{\textcolor{purple}{Running}}

\newcommand{\ziyu}[1]{\textcolor{magenta}{\bf\small [Ziyu: #1]}}

\newcommand{\na}{\multicolumn{1}{c}{---}}

\title{
\uskg: Unifying and Multi-Tasking Structured Knowledge Grounding with Text-to-Text Language Models
}

\author{
{\normalsize 
Tianbao Xie\thanks{\ \ Equal contributions. 
Author contributions in App.~\ref{sec:contributions}.}\ \ $^{1}$ ~ Chen Henry Wu\footnotemark[1] $\ ^2$ ~ Peng Shi$^3$ ~ Ruiqi Zhong$^4$ ~ Torsten Scholak$^5$ \vspace{0.2mm} 
}\\ 
{\normalsize\bf Michihiro Yasunaga$^6$ ~ Chien-Sheng Wu$^7$ ~ Ming Zhong$^8$ ~ Pengcheng Yin$^{9}$ ~ Sida I. Wang$^{10}$ \vspace{0.2mm}
}\\
{\normalsize\bf Victor Zhong$^{17}$ ~ Bailin Wang$^{11}$ ~ Chengzu Li$^{12}$ ~ Connor Boyle$^{17}$ ~ Ansong Ni$^{13}$ ~ Ziyu Yao$^{14}$ 
\vspace{0.2mm}
}\\ 
{\normalsize\bf Dragomir Radev$^{13}$ ~ Caiming Xiong$^{7}$ ~ Lingpeng Kong$^{1,12}$ ~ Rui Zhang$^{15}$
\vspace{0.2mm}
}\\
{\normalsize\bf Noah A. Smith$^{16,17}$ ~ Luke Zettlemoyer$^{10,17}$ ~ Tao Yu$^{1,17}$
\vspace{0.2mm}
}
\\
\affaddr{\normalsize$^{1}$The University of Hong Kong \ \ \ \ } 
\affaddr{\normalsize$^2$Carnegie Mellon University\ \ \ \ }
\affaddr{\normalsize$^3$University of Waterloo} \\
\affaddr{\normalsize$^4$UC Berkeley\ \ \ \ }
\affaddr{\normalsize$^5$ServiceNow Research\ \ \ \ }
\affaddr{\normalsize$^6$Stanford University\ \ \ \ } 
\affaddr{\normalsize$^7$Salesforce Research} \\
\affaddr{\normalsize$^8$UIUC\ \ \ \ }
\affaddr{\normalsize$^{9}$Google Research\ \ \ \ }
\affaddr{\normalsize$^{10}$Facebook AI Research\ \ \ \ } 
\affaddr{\normalsize$^{11}$University of Edinburgh} \\
\affaddr{\normalsize$^{12}$Shanghai AI Lab\ \ }
\affaddr{\normalsize$^{13}$Yale University\ \ }
\affaddr{\normalsize$^{14}$George Mason University\ \ } 
\affaddr{\normalsize$^{15}$Penn State University} \\
\affaddr{\normalsize$^{16}$Allen Institute for Artificial Intelligence \ \ \ \ } 
\affaddr{\normalsize$^{17}$University of Washington}
}

\begin{document}
\maketitle

\begin{abstract}

Structured knowledge grounding (SKG) leverages structured knowledge to complete user requests, such as semantic parsing over databases and question answering over knowledge bases. 
Since the inputs and outputs of SKG tasks are heterogeneous, they have been studied separately by different communities, which limits systematic and compatible research on \skg. 
In this paper, we overcome this limitation by proposing the \uskg framework, which unifies \ntasks SKG tasks into a text-to-text format, aiming to promote systematic SKG research, instead of being exclusive to a single task, \mbox{domain}, or dataset.
We use \uskg to benchmark T5 with different sizes and show that T5, with simple modifications when necessary, achieves state-of-the-art performance on almost all of the \ntasks tasks. 
We further demonstrate that multi-task prefix-tuning improves the performance on most tasks, largely improving the overall performance. 
\uskg also facilitates the investigation of zero-shot and few-shot learning, and we show that T0, GPT-3, and Codex struggle in zero-shot and few-shot learning for SKG. 
We also use \uskg to conduct a series of controlled experiments on structured knowledge encoding variants across \skg tasks. 
\uskg is easily extensible to more tasks, and it is open-sourced at \url{https://github.com/hkunlp/unifiedskg}.\footnote{Latest collections at \url{https://unifiedskg.com}.}

\end{abstract}

\section{Introduction}

\begin{figure*}[!t]
    \centering
    \includegraphics[width=0.75\linewidth]{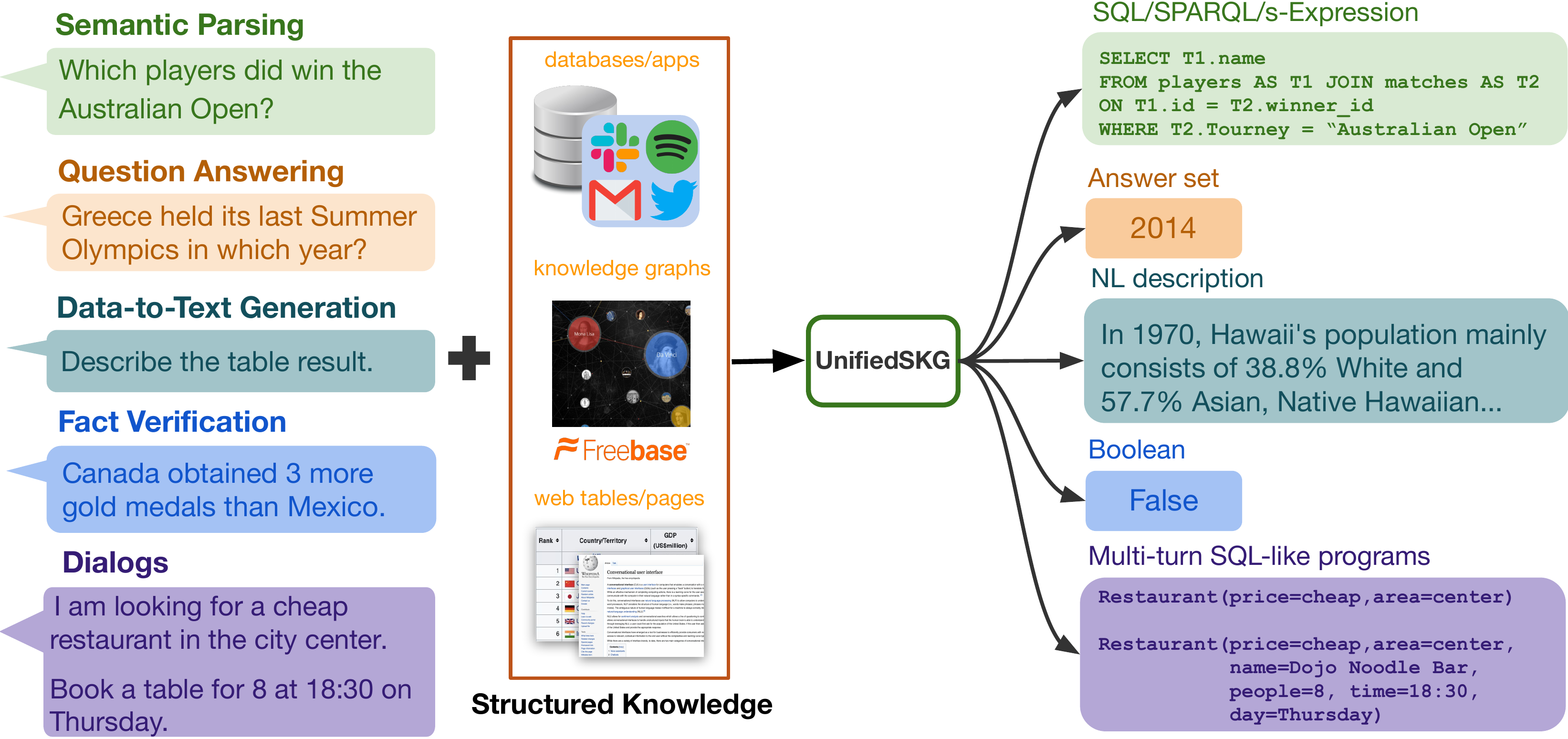}
    \caption{Structured knowledge grounding (SKG) leverages structured knowledge to complete user requests.
    By casting inputs and outputs into the text-to-text format, \uskg standardizes datasets, models, code, experiments, and metrics for \ntasks \skg tasks.
    }
\label{fig:skg}
\end{figure*}

Structured knowledge (e.g., web tables, knowledge graphs, and databases) stores large amounts of data in organized structures, forming a basis for a wide range of applications, e.g., medical diagnosis, personal assistants, and customer relations management.
Accessing and searching data in structured knowledge typically requires mastering query languages through professional training.
To promote the efficiency of data access, structured knowledge grounding (\skg) systems ground user requests in structured knowledge and produce various outputs, including computer programs (e.g., SQL and SPARQL), table cell values, and natural language responses (Figure \ref{fig:skg}). 
For example, semantic parsing~\cite{ZelleM96,Zettlemoyer05} converts natural language questions into formal programs;
knowledge-base question answering \cite{BerantCFL13} derives answers from tables or knowledge graphs.

\skg has attracted significant interest and has been studied through different tasks defined by different communities.
Recent developments in tasks, models, and datasets for \skg have led to task-specific modeling advances, making each task's progress seemingly unique and incompatible.
A main reason is that \skg tasks are \textit{heterogeneous}.
Different types of structured knowledge, such as databases or knowledge graphs, lead to highly specialized encoders~\citep{kagnet-emnlp19,Herzig2020tapas,Wang2020RATSQLRS,yasunaga-etal-2021-qa}.
Some SKG tasks, e.g., semantic parsing, use customized decoders to generate programs~\cite{YinN18,ren2021lego}.
Therefore, instead of solving common challenges in \skg research, improvements in \skg have been prone to be exclusive to a single task, domain, or dataset.

In this paper, we propose the \uskg framework to advocate for a unifying view of \ntasks \skg tasks across six task families and multiple data domains (Table \ref{tab:tasks-included}).
\uskg standardizes datasets, models, code, experiments, and evaluation metrics into a single framework.
By casting user requests, structured knowledge, and outputs into the text-to-text format \cite{2020t5}, it promotes model advances where new tasks can be framed with our standardized abstraction, and new models can be easily applied to diverse \skg tasks.
While previous works also cast \skg tasks into the text-to-text format \cite{hosseini2020simple,Shaw2021CompositionalGA,liu2021tapex}, their independent choices of pretrained language models (PLMs), input-output formats, and frameworks make our unification non-trivial. 
\uskg is easily extensible to more \skg tasks, and it is open-sourced to promote community-wide progress.

Using \uskg as a benchmark, we show that finetuning T5 (with constrained decoding or reranking when necessary) on individual tasks achieves state-of-the-art (sota) results on almost all of the \ntasks tasks, establishing a powerful and reproducible starting point for \skg research. T5 performance also increases with size on most tasks. 

\uskg facilitates multi-task learning on \skg, enabling knowledge sharing and cross-task generalization. 
Although simple multi-task learning has mixed results, we show that multi-task learning with prefix-tuning \citep{li2021prefixtuning} benefits most tasks and largely improves the overall performance, on both T5-base and T5-large.

\uskg is a challenging testbed for few-shot \cite{brown2020language,Ye2021CrossFitAF} and zero-shot learning \cite{Zhong2021AdaptingLM,Jason2021,sanh2021multitask} with PLMs.
Our experiments show that models like T0 \cite{sanh2021multitask} struggle in zero-shot learning on \skg tasks, and GPT-3 \cite{brown2020language} and Codex \cite{chen2021evaluating} struggle in few-shot learning on \skg tasks.

\uskg enables a series of controlled experiments on structured knowledge encoding.
We find that T5 is sensitive to encoding variations, and the sensitivity varies across tasks.
\uskg aims to facilitate more general and robust structured knowledge encoding methods. 
Finally, we conduct a comprehensive error analysis across \skg tasks.
Although the errors made by PLMs decrease with the model size, T5-3B may still generate invalid outputs. 

In summary, we 1) unify and benchmark 21 \skg tasks under the \uskg framework to evaluate diverse grounding goals and structured knowledge sources, 2) demonstrate (near) sota performance of T5 on all the unified \skg tasks, using a single, general-purpose approach, 3) show the benefit of knowledge sharing across \skg tasks via multi-task prefix-tuning, and 4) analyze recent modeling contributions (zero-shot, few-shot, and structured knowledge encoding) on these tasks. 
We hope \uskg enables the design of new models and learning algorithms that generalize to diverse \skg tasks and to identify their challenges.

\section{Related Work}
\label{sec:related-work}

\begin{table*}[ht]
    \small
	\centering
	\begin{adjustbox}{width=0.96\linewidth}
		\begin{tabular}{@{}ccccc@{}}
			\toprule
			Task Family & Task & Knowledge Input & User Input & Output \\
			\midrule
	        \multirow{4}*{\textit{Semantic Parsing}} & Spider \cite{Yu18c} & Database & Question & SQL \\
            & GrailQA \cite{gu2021beyond} & Knowledge Graph & Question & s-Expression \\
            & WebQSP \cite{yih-etal-2016-value} & Knowledge Graph & Question & s-Expression \\
            & MTOP \cite{li-etal-2021-mtop} & API Calls & Question &  TOP Representation\\
			\midrule
			\multirow{6}*{\textit{Question Answering}} & WikiSQL \cite{zhongSeq2SQL2017} & Table & Question & Answer \\
            & WikiTQ \cite{pasupat-liang-2015-compositional} & Table & Question & Answer \\
            & CompWebQ \cite{talmor18compwebq} & Knowledge Graph & Question & Answer \\
            & HybridQA \cite{chen2020hybridqa} & Table + Text Passage & Question & Answer \\
            & MultiModalQA \cite{talmor2021multimodalqa} & Table + Text + Image & Question & Answer \\
            & FeTaQA \cite{nan2021feta} & Table & Question & Free-Form Answer  \\
			\midrule
			\multirow{2}*{\textit{Data-to-Text}} & DART \cite{nan2021dart} & Triple & None & Text \\
			& ToTTo \cite{parikh2020totto} & Highlighted Table & None & Text \\
			\midrule 
			& MultiWoZ \cite{budzianowski2018large} & Ontology & Dialog & Dialog State \\
			& KVRET \cite{Eric2017KeyValueRN} & Table & Dialog & Response \\
			\textit{Conversational} & SParC \cite{Yu19} & Database & Multi turn & SQL \\
			& CoSQL \cite{yu-etal-2019-cosql} & Database & Dialog & SQL \\
			& SQA \cite{iyyer-etal-2017-search} & Table & Multi turn & Answer \\
			\midrule
			\multirow{2}*{\textit{Fact Verification}} 
			& TabFact \cite{2019TabFactA} & Table & Statement & Boolean \\
			& FEVEROUS \cite{aly2021fact} & Table + Text & Statement & Boolean \\
			\midrule
			\multirow{2}*{\textit{Formal-Language-to-Text}} & SQL2Text \cite{shu-etal-2021-logic} & Optional Database & SQL & Text \\
			& Logic2Text \cite{chen-etal-2020-logic2text} & Table Schema & Python-like program & Text \\
			\bottomrule
		\end{tabular}
	\end{adjustbox}
	\caption{We unify \ntasks SKG tasks with different knowledge input, user input, and output, covering six task families. 
	}
	\label{tab:tasks-included}
	\vspace{-3mm}
\end{table*}

\noindent\textbf{SKG with PLMs \ \ }
PLMs have been applied to several SKG tasks. 
To encode structured knowledge, prior work linearized the structured knowledge and concatenated it with the text \cite{Hwang2019ACE,weijie2019kbert,hosseini2020simple,liu2021tapex}, which has been augmented by positional encoding (e.g., row/column embedding) \cite{Herzig2020tapas,yin20tabert} and template-based linearization \cite{chen2020logical,2019TabFactA,oguz2021unik}, and planning \cite{Su2021PlanthenGenerateCD}.
Recently, cell-column alignment is modeled by manipulating the attention matrix of transformers \cite{zhang2020table,eisenschlos2021mate}.
Hierarchical encoding is another way to represent the structure, e.g., 
\citet{wang2021tuta} used tree-based transformers to represent the structure of the tables;
\citet{iida2021tabbie} used transformers to encode row and column representations; \citet{chen2020hitter} used hierarchical transformers to encode KG triples. 
\skg's outputs include, but are not limited to, structured meaning representations (e.g., logic forms, SQL), dialogue states, natural language, answer sets, and Boolean values. Among them, structured meaning representation is challenging for PLMs because they are originally trained on natural language. 
To bridge this gap, \citet{shin2021constrained} adopted the insights from \citet{berant2014semantic} and \citet{marzoev2020unnatural} and proposed to convert formal language into an English-like representation, decode with GPT-3, and map back to formal language automatically.
We do not focus on these techniques in this work; instead, we unify all tasks and systematically compare them.

\noindent\textbf{Task format unification \ \ }
Recent years witnessed the trend of unifying related but different tasks into a shared format.
\citet{abs-1806-08730} unified various tasks as question answering. \citet{YinRRSX20} and \citet{abs-2104-14690} unified few-shot learning as textual entailment. PLUR \cite{chen2021plur} unified program learning, understanding, and repair tasks into a graph-to-sequence format. In this paper, we focus on the text-to-text format \cite{2020t5} due to its flexibility. 
Different from unifying tasks that only take text as input, a core challenge in unifying \skg tasks into the text-to-text format is to linearize structured knowledge.
Notably, UnifiedQA \cite{2020unifiedqa} unified QA tasks, while \uskg covers a broader scope of six task families for systematic exploration. 

\noindent\textbf{Cross-task generalization with PLMs \ \ } 
Multi-task learning and transfer learning go beyond task boundaries, view different tasks as related, and have been shown to outperform single-task learning \cite{muppet,vu2021spot}.
Large PLMs show potential for zero-shot and few-shot learning, e.g., GPT-2 \cite{radford2019language} and GPT-3 \cite{brown2020language}, which can be improved by multi-task learning \cite{Zhong2021AdaptingLM}, e.g., FLAN \cite{Jason2021}, T0 \cite{sanh2021multitask}, and CrossFit \cite{Ye2021CrossFitAF}. ExT5 \cite{aribandi2021ext5} shows that scaling up multi-task learning helps improve pretraining efficiency and downstream performances.
\uskg facilitates the investigation of multi-task, zero-shot, and few-shot learning on \skg tasks.

\begin{figure*}[t]
    \centering
    \includegraphics[width=0.9\linewidth]{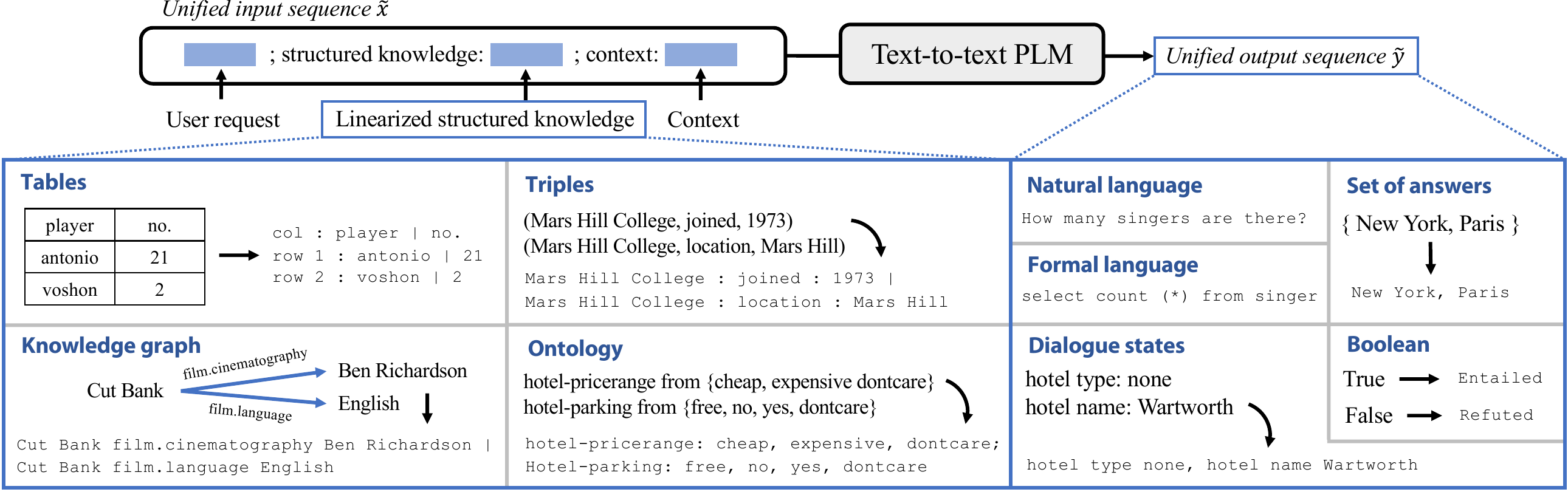}
    \caption{We unify SKG tasks with heterogeneous inputs and outputs into the text-to-text format. }
\label{fig:input-output-format}
\end{figure*}

\section{The \uskg Framework}
\label{sec:method}


\subsection{Task Unification}
\label{subsec:task_unification}
The guiding principle of \uskg{}'s task selection is diversity. We unify \ntasks SKG tasks across six task families and multiple domains (Table \ref{tab:tasks-included}). Our task families include:
\begin{itemize}[leftmargin=*]
    \item \textbf{Semantic parsing} converts questions to logical forms \cite{ZelleM96,Zettlemoyer05}.
    \item \textbf{Question answering} derives answers to natural language questions based on structured data \cite{BerantCFL13}.
    \item \textbf{Data-to-text generation} describes structured data in natural language \cite{NovikovaDR17}.
    \item \textbf{Fact verification} checks if a statement is true based on the structured data \cite{2019TabFactA}.
    \item \textbf{Conversational tasks} require understanding of not only the user's last request but also the full interaction history between users and machines \cite{budzianowski2018large, eric2019multiwoz, yu-etal-2019-cosql}.
    \item \textbf{Formal language to text translation} describes formal language in natural language \cite{chen-etal-2020-logic2text}. 
\end{itemize}

All these tasks take as input $x$ a user request, a structured knowledge input, and an optional (dialogue) context to predict an output $y$.
Figure \ref{fig:input-output-format} illustrates how we convert the input $x$ to an input sequence $\tilde{x}$ and the output $y$ to an output sequence $\tilde{y}$ by means of ``linearization'' \cite{liu2021tapex}, enabling the unification of diverse forms of structured knowledge. We provide more details, examples, and input length analysis in the Appendices \ref{app:task-unification} and \ref{app:examples}.
Our code implementation uses Hugging Face's Transformers \cite{wolf-etal-2020-transformers} and Datasets \cite{lhoest2021datasets} toolkits. 

\subsection{Modeling}
The simplest usage of \uskg is to train on individual tasks. In this case, we minimize the negative log-likelihood loss averaged over tokens in each batch. 
For decoding, we use beam search by default.
\uskg also facilitates exploration of multi-task learning, few-shot, and zero-shot learning with PLMs, and details are presented in the corresponding parts in Section \ref{sec:results}.

\begin{table*}[ht]
	\begin{minipage}{\textwidth}
	    \small
        \centering
	    \begin{adjustbox}{width=0.95\linewidth}
    		\begin{tabular}{@{}l@{}ccccl|l@{}}
    			\toprule
    			& Metric & T5-base
    			& T5-large
    			& T5-3B & Previous sota (w/o extra) & Previous sota (w/ extra)
    			\\ 
    			\midrule
    	        \multirow{1}*{Spider (dev.)} 
                & Match & 58.12 & 66.63 & 71.76 & \textbf{75.5}$^{+}$ \cite{Scholak2021:PICARD} & 74.7  \cite{rubin-berant-2021-smbop} \\
                \multirow{1}*{GrailQA}
                & Match & 62.39	& 67.30 & 70.11 & 
                \textbf{83.8}$^{+}$ \cite{ye2021rng} 
                & \na \\
                \multirow{1}*{WebQSP}
                & F1 & 78.83 & 79.45 & 80.70 & 
                \textbf{83.6}$^{+}$ \cite{ye2021rng} 
                & \na \\
                MTOP & Match & 85.49 & 86.17 & \textbf{86.78} & 86.36 \cite{pasupat-etal-2021-controllable} & \na \\
    			\midrule
                \multirow{1}*{WikiTQ}
                & Acc & 35.76 & 43.22 & \bf 49.29 & 44.5 \cite{wang-etal-2019-learning} & 57.5  \cite{liu2021tapex}\\
                \multirow{1}*{WikiSQL}
                & Acc & 82.63 & 84.80 & \textbf{85.96} & 85.8 \cite{liu2021tapex} & 89.5 \cite{liu2021tapex}\\
                \multirow{1}*{CompWebQ}
                & Acc & 68.43 & 71.38 & 73.26 & 70.4$^{\ddag}$ \cite{das2021case} & \na \\
                \multirow{1}*{HybridQA (dev.)}
                & Acc & 54.07 & 56.95 & 59.41 & 60.8$^\ddag$ \cite{eisenschlos2021mate} & 63.4$^\ddag$  \cite{eisenschlos2021mate} \\
                \multirow{1}*{MultiModalQA (dev.)}
                & F1 & 75.51	& 81.84	& \bf 85.28 &    82.7 \cite{yoran2021turning} &  83.8 \cite{yoran2021turning} \\
                \multirow{1}*{FeTaQA}
                & BLEU & 29.91 & 32.45 & \bf 33.44 & 30.54 \cite{nan2021feta} & \na \\
    			\midrule
    			\multirow{1}*{DART}
    			& BLEU & 46.22 & \textbf{46.89} & 46.66 & 46.89 \cite{nan2021dart} & 47.2  \cite{aghajanyan2021htlm} \\
    			\multirow{1}*{ToTTo (dev.)}
    			& BLEU & 48.29 & \bf 48.95 & \bf 48.95 & 
    			48.95 \cite{KaleR20a} & \na \\
                \midrule
    			\multirow{1}*{MultiWoZ2.1}
    			& Joint Acc & 54.64 & 54.45 & 55.42 & \textbf{60.61}$^*$\cite{dai-etal-2021-preview} & 60.48 \cite{yu2021SCoRE}\\
    			\multirow{1}*{KVRET}
    			& Micro F1 & 66.45 & 65.85 & \bf 67.88 & 63.6 \cite{gou2021contextualize} & \na \\
    			\multirow{1}*{SParC (dev.)}
    			& Match & 50.54 & 56.69 & \bf 61.51 & 54.1 \cite{hui2021dynamic} & 62.2 \cite{yu2021SCoRE}\\
    			\multirow{1}*{CoSQL (dev.)}
    			& Match & 42.30 & 48.26 & 54.08 & \textbf{56.9}$^{+}$  \cite{Scholak2021:PICARD} & 52.1 \cite{yu2021SCoRE}\\
    			\multirow{1}*{SQA}
    			& Overall Acc  & 52.91 & 61.28 & \bf 62.37 & 58.6 \cite{liu2021tapex} & 74.5 \cite{liu2021tapex}\\
    			\midrule
    			\multirow{1}*{TabFact}
    			& Acc & 76.13 & 80.85 & \bf 83.68 & 74.4 \cite{yang-etal-2020-program} & 84.2 \cite{liu2021tapex}\\
    			\multirow{1}*{FEVEROUS (dev.)}
    			& Acc & 75.05 & 79.81 & \bf 82.40 & 82.38 \cite{aly2021fact} & \na \\
    			\midrule
    			\multirow{1}*{SQL2Text}
    			& BLEC & 93.52 & 93.68 & \textbf{94.78} & 93.7 \cite{shu-etal-2021-logic} & \na \\
    			\multirow{1}*{Logic2Text}
    			& BLEC & 90.66	& 90.57	& \bf 91.39 & 88.6 \cite{shu-etal-2021-logic} & \na \\
    			\bottomrule
    		\end{tabular}
    	\end{adjustbox}
	    \caption{Test or development (dev.) set performance of models trained on individual tasks.
	    Vanilla T5 or T5 with simple modifications (e.g., $^{+}$constrained decoding or reranking) achieve sota on nearly all tasks.
	    The best result without extra pretraining is shown in \textbf{bold}.
	    More detailed results and result variances can be found in Tables \ref{tab:baseline-full-dev} and \ref{tab:baseline-full-test} in Appendix.
	    Human evaluation for generation tasks is in Section \ref{subsec:human-eval}.
	    \textit{w/ (w/o) extra} means with (without) extra pretraining on unsupervised structured data (e.g., web tables).\footnotemark
	    }
	\label{tab:baseline-simple-version}
	\end{minipage}
\end{table*}
\begin{table}[t]
\small
	\centering
	\begin{adjustbox}{width=0.98\linewidth}
		\begin{tabular}{@{}l@{}c@{}c@{}c@{}c@{}c@{}c@{}}
			\toprule
			& Spider \ & \ WikiTQ \ & \ \ DART \ \ &  \ MWoZ \ \ &  \ TabFact \ \ & SQL2Text \\ 
			\midrule 
		    T5-3B \ \ \ & \bf 71.76 & \bf 50.65 & \bf 50.38 & 58.46 & 83.97 & 92.71 \\
            T0-3B & 68.09 & 50.62 & 50.16 & \bf 60.20 & \bf 85.51 & \bf 92.93 \\
			\bottomrule
		\end{tabular}
	\end{adjustbox}
	\caption{Comparison between T5-3B and T0-3B. T0-3B is initialized from LM-adapted T5 and further pretrained on a large number of non-\skg tasks. We finetune both models on individual tasks.
	T0-3B under-performs T5-3B on semantic parsing (Spider) and outperforms T5-3B on dialogue state tracking (MWoZ) and fact verification (TabFact). We report results on the dev. set. }
	\vspace{-3mm}
	\label{tab:t0}
\end{table}

\section{Experiments and Analysis}
\label{sec:results}

\subsection{Results on Individual Tasks} 
\label{subsec:baseline-performance}
We apply T5 models \cite{2020t5} on each individual task in \uskg.
For model training, we set the maximum number of epochs as 50--200, depending on the dataset size. We use early stopping and model selection on the development set. 
More details are shown in Appendix \ref{subapp:implementation-details}. 
For each task, we report one commonly used metric in Table~\ref{tab:baseline-simple-version}. See Appendix \ref{app:all-results} for all metrics. 

\noindent\textbf{Comparison with previous sota \ \ }
Table~\ref{tab:baseline-simple-version} shows that vanilla T5-3B outperforms most previous sota models not trained on extra unsupervised in-domain data.  
Some semantic parsing sota models, denoted as $^{+}$ in Table \ref{tab:baseline-simple-version}, are also T5 with constrained decoding \cite{Scholak2021:PICARD} or reranking \cite{ye2021rng}. 
This shows that a generalist architecture like T5, when scaled up to a certain size, can be as good as task-specific architectures for \skg, suggesting the potential of larger PLMs. 

\noindent\textbf{Model scalability \ \ } 
In general, T5 performance increases with the model size, but this trend varies across task families.
Semantic parsing, QA, and fact verification tasks get large benefits from increased sizes, while text generation does not. 
See Section \ref{subsec:human-eval} for a human evaluation for text generation tasks.
Also, the gap between T5-base (220M) and T5-large (770M) is larger than the gap between T5-large (770M) and T5-3B (3B). 

\noindent\textbf{Effect of pretraining on structured knowledge} \ \
Some smaller models pretrained on structured knowledge \cite{liu2021tapex} show competitive performance as T5-3B, suggesting that pretraining with structured data is beneficial for \skg. This result calls for structured knowledge pretraining that generalizes to different \skg tasks across domains, which can be systematically explored using \uskg. 

\noindent\textbf{Effect of pretraining on non-\skg tasks \ \ }
T0-3B \cite{sanh2021multitask} is initialized from T5-3B and pretrained on multiple tasks that (in most cases) do not use structured knowledge as input (non-\skg tasks). 
Exploring the performance of T0-3B on \skg tasks helps us understand the relationship between \skg tasks and non-\skg tasks. 
Table \ref{tab:t0} shows that T0-3B under-performs T5-3B on semantic parsing and outperforms T5-3B on dialogue state tracking and fact verification. 
We note that T0-3B is pretrained on dialogue QA, dialogue summarization, and NLI tasks; therefore, pretraining on non-\skg tasks might not be useful for \skg unless we add similar \skg tasks to pretraining.

\footnotetext{%
For GrailQA and WebQSP, we run T5 and rerun the previous sota model~\cite{ye2021rng} using the gold entities. 
For MultiModalQA and FEVEROUS, we report performance of T5 and the previous sota models on the dev. samples with at least one table (samples with image input are further excluded for MultiModalQA);
The gold table and text candidates are used for both T5 and previous sota~(for MultiModelQA, numbers are from~\cite{yoran2021turning}, and for FEVEROUS, we rerun the available model~\cite{aly2021fact} on gold candidates to obtain the number).
We use sacreBLEU to report all BLEU results.
$^\ddag$We use gold entity linking, but the previous sota does not, which makes the results not directly comparable; therefore, we do not bold any numbers for CompWebQ and HybridQA.
$^*$T5-base with the independent output scheme \cite{lee-etal-2021-dialogue} achieves 56.66 on MWoZ2.1, higher than our sequence output scheme. 
For WebQSP, as the original dataset does not have a dev.~set, we split the original train set into in-house train/dev.~sets (90\%/10\%), following prior practice (e.g. \citet{ren2021lego}). Similarly, for CompWebQ, as the test set is not publicly available, we split the original dev.~set into in-house dev./test sets (20\%/80\%). For GrailQA, we split the original dev.~set into in-house dev./test sets (5\%/95\%).
}

\subsection{Multi-Task Learning}
\uskg facilitates the exploration of multi-task learning. In this part, we systematically study multi-task learning on all \ntasks unified tasks. We find that \skg benefits from multi-task prefix-tuning on both T5-base and T5-large, showing that the benefits from multi-task learning is scalable in terms of the model size. The baselines we use include:

\noindent\textbf{Single-task finetuning (ST-F)}, which is finetuning on individual tasks, same as Section \ref{subsec:baseline-performance}.

\noindent\textbf{Single-task prefix-tuning (ST-P;} \citealp{li2021prefixtuning}\textbf{)}, which learns lightweight task-specific parameters while keeping the PLM fixed. We set the prefix length as 10. \citet{Clive2021ControlPF} also used prefix-tuning on T5 for data-to-text generation. 

\noindent\textbf{Multi-task finetuning (MT-F)}, which combines the training data of all tasks with temperature mixing (\citealp{2020t5}; after hyperparameter tuning with a few steps, we set the temperature as 2). We select model weights based on the average metric on all tasks' development set.

Table~\ref{tab:multitask-dev-simple} shows that ST-P is comparable to ST-F on nearly all tasks. 
However, we find that it takes about 5--10 times as many training steps (See Appendix \ref{app:step}), which is similarly observed for prompt-tuning \cite{LesterAC21}. 
We also observe that MT-F leads to mixed results. 
For many tasks, MT-F is even worse than ST-F.

\noindent\textbf{Multi-task prefix-tuning (MT-P) \ \ } 
Our explanation for the mixed results of MT-F is that the inputs of SKG tasks contain different structured knowledge from diverse domains, making it difficult to learn shared parameters effectively. 
To address this challenge, we first pretrain a prefix on all tasks, freezing T5 and using the same temperature mixing as MT-F. In the second step, we initialize each task's prefix with this pretrained prefix and optimize the prefix while freezing T5. 
This initialization step is similar to the prompt transfer explored in \citet{vu2021spot}.
Following ST-P, we set the prefix length as 10. 

Table~\ref{tab:multitask-dev-simple} shows that multi-task prefix-tuning outperforms single-task finetuning and single-task prefix-tuning on most tasks, and it largely outperforms the naive multi-task learning baseline.
It demonstrates that SKG tasks can be studied together to share data and knowledge.

\begin{table}[t!]
	\centering
	\begin{adjustbox}{width=0.95\linewidth}
		\begin{tabular}{@{}lcccc|cc@{}}
			\toprule
			\multicolumn{1}{c}{}  & \multicolumn{4}{c|}{T5-base} & \multicolumn{2}{c}{T5-large} \\
		    & ST-F & ST-P & MT-F & MT-P & ST-F & MT-P
			\\ 
			\midrule
	        \multirow{1}*{Spider} 
            & 58.12 & 58.61 & 58.90 & \bf 59.86 & 66.63 & \bf 67.60 \\
            \multirow{1}*{GrailQA}
            & 60.00 & 61.33 & 56.00 & \bf 62.67 & \bf 67.00 & 65.33 \\
            \multirow{1}*{WebQSP}
            & 72.50 & 73.81 & 67.25 & \bf 74.77 & 73.96 & \bf 74.92 \\
            \multirow{1}*{MTOP}
            & \bf 83.89 & 82.93 & 78.79 & 82.77 & \bf 84.70 & 84.34 \\
			\midrule
            \multirow{1}*{WikiTQ}
            & 36.94 & 36.42 & \bf 41.15 & 39.74 & 43.30 & \bf 50.90 \\
            \multirow{1}*{WikiSQL}
            & \bf 84.50 & 83.09 & 81.85 & 84.44 & 86.27	& \bf 87.45 \\
            \multirow{1}*{CompWQ}
            & 66.71 & 67.85 & 68.28 & \bf 69.70 & 68.85	& \bf 71.27 \\
            \multirow{1}*{HybridQA}
            & 54.07 & \bf 54.93 & 53.52 & 54.88 & 56.95	& \bf 57.33 \\
            \multirow{1}*{MMQA}
            & 75.51 & 75.50 & \bf 76.63 & 76.40 & 81.84	& \bf 84.59 \\
            \multirow{1}*{FeTaQA}
            & 29.00 & 28.03 & \bf 31.85 & 29.33 & 30.94	& \bf 32.48 \\
			\midrule 
			\multirow{1}*{DART} 
			& 50.62 & 50.33 & 49.74 & \bf 50.68 & \bf 51.72	& 50.82 \\
			\multirow{1}*{ToTTo}
			& \bf 48.29 & 45.70 & 45.29 & 45.21 & \bf 48.95	& 47.90 \\
            \midrule
			\multirow{1}*{MWoZ2.1}
			& \bf 57.52 & 56.67 & 53.19 & 57.06 & 58.23	& \bf 59.24 \\
			\multirow{1}*{KVRET} 
			& 20.04 & 19.68 & 18.53 & \bf 21.32 & 18.84	& \bf 20.76 \\
			\multirow{1}*{SParC}
			& 50.54 & 51.04 & \bf 51.70 & 51.29 & 56.69	& \bf 59.02 \\
			\multirow{1}*{CoSQL}
			& 42.30 & 44.39 & 43.59 & \bf 45.68 & 48.26	& \bf 51.64 \\
			\multirow{1}*{SQA}
		    & 49.49 & 44.81 & \bf 51.48 & 48.43 & \bf 59.12	& 58.15 \\
			\midrule
			\multirow{1}*{TabFact}
			& 76.34 & 75.74 & 71.19 & \bf 77.86 & 81.40 & \bf 83.62 \\
            \multirow{1}*{FEVER.}
            & 75.05 & 75.33 & 76.85 & \bf 78.02 & 79.81	& \bf 82.05 \\
			\midrule
			\multirow{1}*{SQL2Text}
			& 93.69 & \bf 94.50 & 93.57 & 93.79 & 93.35	& \bf 93.93 \\
			\multirow{1}*{Logic2Text}
			& 92.15 & \bf 95.25 & 92.24 & 94.70 & 92.88	& \bf 93.61\\
			\midrule
			Total para. & $21T$ & $T + 21P$ & $T$ & $T + 21P$ & $21T$ & $T + 21P$ \\
			Avg. score & 60.82 & 60.76 & 60.08 & \bf 61.84 & 64.27 & \bf 65.57 \\
			\bottomrule
		\end{tabular}
		\end{adjustbox}
	\caption{Multi-task learning results. 
	ST and MT stand for single-task and multi-task. F and P stand for finetuning and prefix-tuning. 
	For total parameters, $T$ and $P$ are the numbers of T5 and prefix parameters ($P \ll T$). 
    Multi-task learning with prefix improves the performance on most tasks, largely improving the overall performance. We report results on the dev. set. }
	\label{tab:multitask-dev-simple}
	\vspace{-3mm}
\end{table}


\begin{table}[!t]
	\centering
	\begin{adjustbox}{width=0.85\linewidth}
		\begin{tabular}{@{}ccccc@{}}
			\toprule
			Task A & Task B & Type & B only & A to B \\
			\midrule
			\multirow{1}*{WikiSQL}
			& \multirow{1}*{TabFact}
			& same source & 81.43 & 82.76 \\
			\multirow{1}*{TabFact}
			& \multirow{1}*{WikiTQ}
			& same source & 43.30 & 45.88 \\
			\multirow{1}*{WikiSQL}
			& \multirow{1}*{FeTaQA}
	        & same source & 30.94 & 31.19 \\
			\multirow{1}*{Spider}
			& \multirow{1}*{GrailQA}
	        & parallel tasks & 67.00 & 67.00 \\
			\multirow{1}*{Spider}
			& \multirow{1}*{WikiTQ}
            & subtask & 43.30 & 41.68 \\
			\multirow{1}*{Spider}
			& \multirow{1}*{TabFact}
			& weakly related & 81.43 & 80.39 \\
	        \bottomrule
		\end{tabular}
		\end{adjustbox}
	\caption{Task knowledge transfer. We use T5-large here. \textit{B only} means training the model on task B; \textit{A to B} means to train the model on task A and then to finetune the model on task B. In both settings, we report task B's development set performance. We find that tasks benefit from other tasks with the same data source. }
	\label{tab:task-relationship-main}
\end{table}

\noindent\textbf{Exploring task knowledge transfer \ \ } 
\uskg facilitates studying knowledge transfer between \skg tasks. Given two tasks, \textit{task A} and \textit{task B}, we first train the model on task A and then continue training on task B. Table \ref{tab:task-relationship-main} shows that tasks benefit from other tasks with the same data source (e.g., tasks that all use Wikipedia tables as structured knowledge). We do not observe positive transfer between \textit{parallel tasks} (e.g., semantic parsing tasks with different structured knowledge and different output) and \textit{subtask} (e.g., question answering can be viewed as the execution semantic parses) when data sources are different.  
Compared to the positive results in Table \ref{tab:multitask-dev-simple}, results in this part indicate that manually selecting source and target tasks may not be efficient for multi-task learning. 

\subsection{Zero-Shot and Few-Shot Learning}
\label{subsec:pretrained-models-analysis}
The text-to-text unification of \uskg enables us to investigate zero/few-shot learning on SKG with large PLMs. 

\noindent\textbf{Zero-shot learning setting \ \ } 
Zero-shot learning enables models to solve tasks with natural language descriptions without training samples. We follow T0 \cite{sanh2021multitask} to create similar natural language instructions for the unseen tasks. Our instructions are provided in Appendix \ref{subapp:t0_zero}.

\begin{table}[t]
    \small
	\centering
	\begin{adjustbox}{width=\columnwidth}
		\begin{tabular}{@{}l@{}cccccc@{}}
			\toprule
			& T5-3B &
			\multicolumn{1}{c}{T0 3B} &
			\multicolumn{2}{c}{GPT-3 175B} & \multicolumn{2}{c}{Codex 175B} \\ 
			& \textit{finetune} & \textit{zero-shot} & \textit{select} & \textit{random} & \textit{select} & \textit{random} \\
			\midrule
			\multirow{1}*{Spider} 
            & 71.76 & \ \ 0.00 & 20.00 & 18.33$_{3.78}$ & 40.72 & 43.23$_{4.16}$ \\
            \multirow{1}*{WikiTQ}
            & 50.65 & 12.68 & 32.00 & 29.33$_{9.04}$ & 26.21 & 20.46$_{4.21}$ \\
            \multirow{1}*{DART} 
			& 50.38 & 23.42 & 40.23 & 34.21$_{4.50}$ & 42.13 & 36.54$_{1.67}$ \\
			\multirow{1}*{MWoZ}
			& 58.46 & \ \ 0.00 & 18.00 & \ \ 0.02$_{0.02}$ & 23.47 & \ \ 0.06$_{0.03}$ \\
			\multirow{1}*{TabFact}
			& 83.97 & 52.45 & 51.00 & 49.67$_{3.79}$ & 50.97 & 51.58$_{1.59}$ \\
			\multirow{1}*{SQL2Text}
			& 92.71 & 39.64 & 94.00 & 85.00$_{2.65}$ & 90.64 & 88.31$_{1.61}$ \\
			\bottomrule
		\end{tabular}
	\end{adjustbox}
	\caption{Zero-shot and few-shot learning for SKG. Subscripts show the standard deviation with three runs. 
	\textit{select} means to select the most similar training samples as few-shot examples, while \textit{random} means to randomly select training samples as few-shot examples. 
	T0 performs poorly on all the tasks in the zero-shot setting.
	Codex outperforms GPT-3 on tasks that generate structured programs (Spider and MultiWoZ).
	}
	\label{tab:effect-of-models}
\end{table}

\noindent\textbf{Few-shot learning settings \ \ } \citet{brown2020language} showed that large PLMs could be few-shot learners by encoding a few training samples as ``context'' to learn without gradient updates. 
We use GPT-3 \cite{brown2020language} and Codex \cite{chen2021evaluating} to explore such few-shot learning for SKG. To stay within our budget, for GPT-3, we report the performance on 100 random dev. set samples. We explore two settings for few-shot learning. 

In the first setting, we randomly sample few-shot examples from the training set; these examples are shared by all dev. set samples, denoted as \textit{random} in Table \ref{tab:effect-of-models}. 
For sequences that are too long for Codex (4096) and GPT-3 (2048), we use as many examples as possible and make sure that there is at least one example (truncated if needed). 

In the second setting, we follow \citet{gao2021making} to select few-shot examples from the training set. 
We call this setting \textit{few-shot with example selection}, denoted as \textit{select} in Table \ref{tab:effect-of-models}. 
We use the pretrained SBERT \cite{reimers-2020-multilingual-sentence-bert} for sentence embeddings of the user request input (for tasks that only have structured input, we embed the linearized structured input) and sample five most similar examples measured by cosine similarity. Further details (e.g., prompts and task instructions) are provided in Appendix \ref{subapp:gpt3_codex}.

\noindent\textbf{SKG is challenging for zero/few-shot learning. }
Table~\ref{tab:effect-of-models} shows that zero-shot performance is very poor on most tasks (Spider and MultiWoZ are even 0). 
It also shows a large gap between few-shot learning and finetuning for Spider, WikiTQ, MWoZ, and TabFact, while the gap is smaller for generation tasks. 
For few-shot learning, example selection based on similarity outperforms random selection, but the gap is usually smaller than 10 points out of 100. 
It is also interesting to compare the results between \textit{synthesis} tasks (Spider), which requires predicting programs, and \textit{induction} tasks (WikiTQ and TabFact), where a model directly outputs answers \citep{Devlin2017RobustFillNP}. 
We find that PLMs generally struggle more when adapting to induction tasks (e.g., close to random-guess on the binary classification task TabFact), reminiscent of recent attempts in program synthesis and induction using PLMs~\citep{Austin2021ProgramSW}.
For GPT-3 and Codex, 
better zero-shot performances can be expected by better prompt design. 

\subsection{Structured Knowledge Encoding}
\label{subsec:structured-encoding-analysis}
Structured knowledge encoding has been widely explored (\citealp{Bogin2019GlobalRO,kagnet-emnlp19,agarwal2020knowledge,saxena2020improving,yasunaga2020graph,yasunaga2022dragon}; and others detailed in Section \ref{sec:related-work}). We hope that \uskg can promote systematic study of \textit{general} structured knowledge encoding. 
To this end, this part focuses on the linearization of structured knowledge. 

\begin{table}[t]
	\centering
	\small
	\begin{adjustbox}{width=0.95\linewidth}
		\begin{tabular}{@{}lcccc@{}}
		    \toprule
		    & Spider & WikiTQ & MultiWoZ2.1 & TabFact\\
			\midrule
			 \textit{rs(c)} & \quad \ \ 66.63$_{2.31}$ & \quad \ \ 43.30$_{0.25}$ & \quad \ \ 58.23$_{0.39}$ & \quad \ \ 81.43$_{0.16}$ \\
			 \textit{sr} & 64.12 & 38.78 & \na & 80.98 \\ \textit{rcs} & \na & \na & 58.89 & \na \\
			\bottomrule
		\end{tabular}
	\end{adjustbox}
	\caption{Ordering of inputs. Subscripts show the standard deviation with three runs. \textit{s}, \textit{r}, and \textit{c} stand for the structured knowledge, request input, and context. Placing \textit{r} before \textit{s} is always better, and placing \textit{c} between \textit{r} and \textit{s} is better for dialogue state tracking (MultiWoZ2.1).
	}
	\label{tab:sk_txt_ordering}
\end{table}

\begin{table}[t]
    \small
	\centering
	\begin{adjustbox}{width=1\linewidth}
		\begin{tabular}{@{}l@{}cccc@{}}
		    \toprule
			& Spider & WikiTQ & DART & MultiWoZ2.1 \\
			\midrule
			Same Order & \quad \ \ 66.63$_{2.31}$ & \quad \ \ 43.30$_{0.25}$ & \quad \ \ 51.72$_{0.15}$ & \quad \ \ 58.23$_{0.39}$ \\
			Reversed Order & 64.80 & 37.80 & 48.47 & 13.59 \\
			\bottomrule
		\end{tabular}
	\end{adjustbox}
	\caption{Order-sensitivity of structured knowledge. Subscripts show the standard deviation with three runs. \textit{Same Order} is the default benchmark setting. \textit{Reversed Order} means to reverse the structured knowledge ordering on the development set (but not the training set). 
	Tasks with cross-domain tables (in WikiTQ), databases (in Spider), and triples (in DART) are less order-sensitive, while pre-defined ontology (in MultiWoZ2.1) is highly order-sensitive. }
	\label{tab:sk_ordering}
\end{table}

\noindent\textbf{Does the order of user input, structured knowledge, and context matter?} 
To explore the effect of the order of user input, structured knowledge, and context, we rerun the single-task experiments while switching the order of these components in both the training and development set. 
Table \ref{tab:sk_txt_ordering} shows that placing the text before structured knowledge (\textit{rs}) is better than the opposite (\textit{sr}), which is consistent across \skg tasks.
Our explanation is that the position of the text is relatively fixed in \textit{rs}, helping the decoder to learn stable attention over the text. 
Also, placing the context in between the text and structured knowledge yields better results.

\noindent\textbf{Is T5 sensitive to structured knowledge ordering?}
Order-insensitivity is common for most structured knowledge, e.g., permutation of columns in a table preserves the meaning. 
To study this insensitivity, we evaluate T5-large on a manipulated development set where the order of schema (for database), column (for table), or slots and values (for ontology) is reversed. 
Table \ref{tab:sk_ordering} shows that tasks with cross-domain tables and databases are less order-sensitive, while models are very sensitive to the order of ontology. 
Other types of robustness (e.g., robustness to cell values irrelevant to the answer) remain an open question in \uskg. 

\begin{table}[t]
	\centering
	\small
	\begin{adjustbox}{width=0.85\linewidth}
		\begin{tabular}{@{}lccc@{}}
		    \toprule
			& Spider & WikiSQL & TabFact \\
			\midrule
			Linearization  & 40.23 & 59.21 & 58.77  \\
			Natural Language & 38.59 & 63.16 & 58.56  \\
			\bottomrule
		\end{tabular}
	\end{adjustbox}
	\caption{Converting structured knowledge into natural language for low-resource learning. 
	A large improvement is observed on question answering (WikiSQL), but not on text2SQL semantic parsing (Spider) and fact verification (TabFact). }
	\label{tab:convert-to-nl}
\end{table}
\noindent\textbf{Is it beneficial to represent structured knowledge as natural language?}
SKG data is not typically used to pretrain PLMs. Given ample training data, PLMs adapt well to SKG tasks, as shown in Table \ref{tab:baseline-simple-version}. However, under the low-resource setting, converting structured data to natural language might be helpful. 
For Spider, we use a shared template to convert structured data to natural language. 
For TabFact and WikiSQL, we randomly selected 236 tables shared by both datasets and manually labeled templates to convert each row into a sentence. Examples of the templates are shown in Appendix \ref{app:examples_template}. These templates produce about 1000 samples for each task, divided into training and test sets. We find that, in WikiSQL, the conversion to natural language stabilizes and accelerates the training process. 
Table \ref{tab:convert-to-nl} shows that conversion to natural language improves the performance on WikiSQL, has no significant influence on TabFact, and slightly degrades the performance on Spider.

\begin{table}[t]
	\centering
	\small
	\begin{adjustbox}{width=0.85\linewidth}
	\begin{tabular}{@{}lllll@{}}
			\toprule
			& Metric & T5-base
			& T5-large
			& T5-3B
			\\ 
			\midrule
            \multirow{2}*{FeTaQA}
            & BLEU & 29.00 & 30.94 & 31.73 \\
            & Human$^{*\dag}$ & 36.0\% & 51.3\% & 57.3\% \\
			\midrule 
			\multirow{2}*{DART} 
			& BLEU & 50.62 & 51.72 & 50.38 \\
			& Human & 90.7\% & 91.7\% & 87.7\% \\
            \midrule
			\multirow{2}*{ToTTo}
			& BLEU & 48.29 & 48.95 & 48.95 \\
			& Human & 78.7\% & 80.0\% & 81.3\% \\
            \midrule
			\multirow{2}*{KVRET}
			& BLEU & 20.04 & 18.84 & 17.75 \\
			& Human$^{\dag}$ & 72.3\% & 66.3\% & 75.0\% \\
            \midrule
			\multirow{2}*{SQL2Text}
			& BLEC & 93.69 & 93.35 & 92.71 \\
			& Human$^{*}$ & 83.7\% & 90.3\% & 84.7\% \\
            \midrule
			\multirow{2}*{Logic2Text}
			& BLEC & 92.15 & 92.88 & 91.69 \\
			& Human$^{\dag}$ & 77.2\% & 81.5\% & 84.2\% \\
			\bottomrule
	\end{tabular}
	\end{adjustbox}
	\caption{Automatic metrics and human evaluation on the development set of generation tasks. $^{*}p<0.05$ for ``the rank-1 model is better than the rank-2 model''. $^{\dag}p<0.05$ for ``the rank-2 model is better than the rank-3 model''. Automatic metrics do not always reflect human evaluation. Larger models are not always better. }
	\label{tab:human-eval}
\end{table}

\begin{figure}[t]
\centering
    \subfigure[Spider]{
        \includegraphics[width=0.45\linewidth]{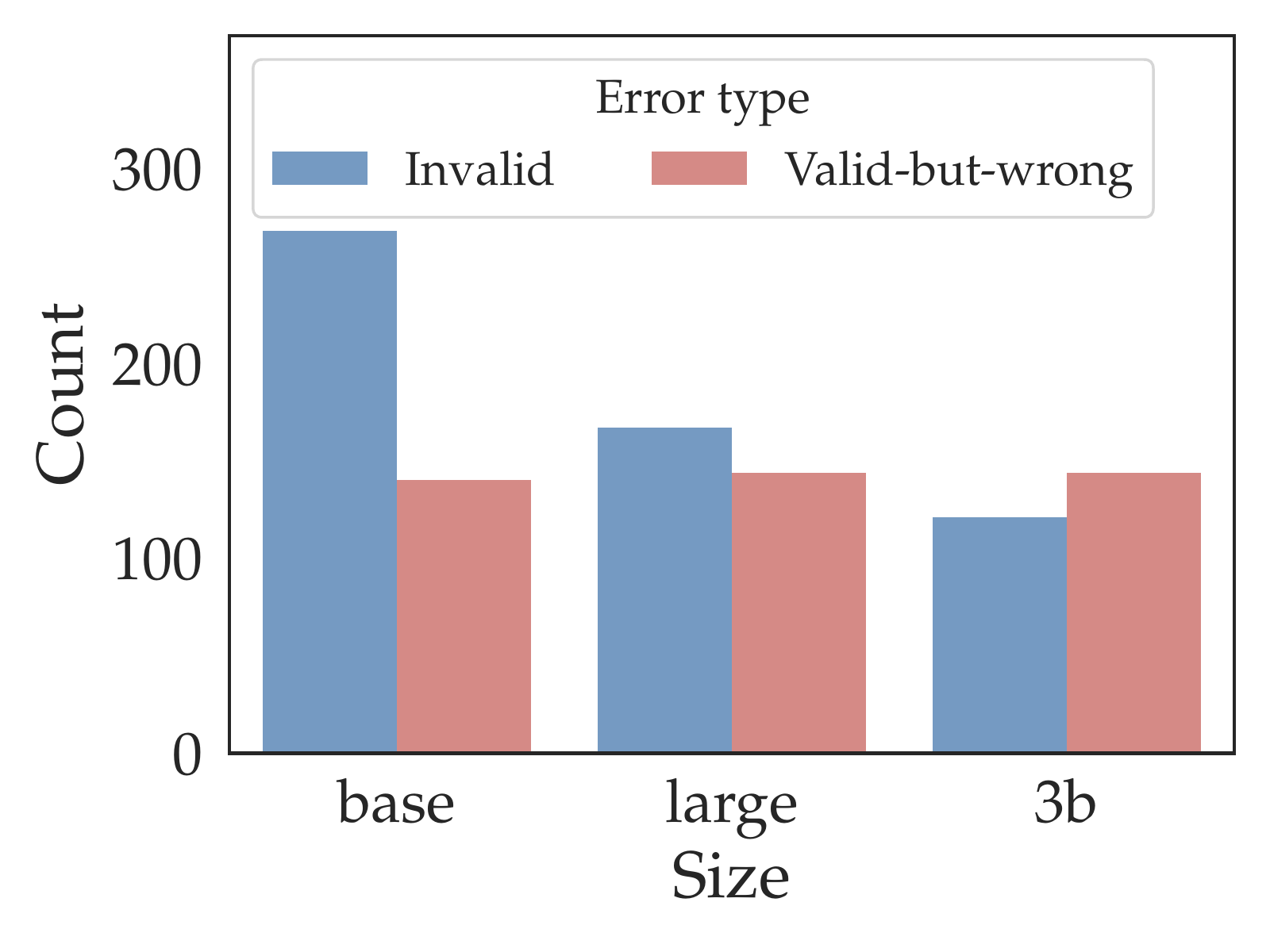}
    }
    \subfigure[GrailQA]{
        \includegraphics[width=0.45\linewidth]{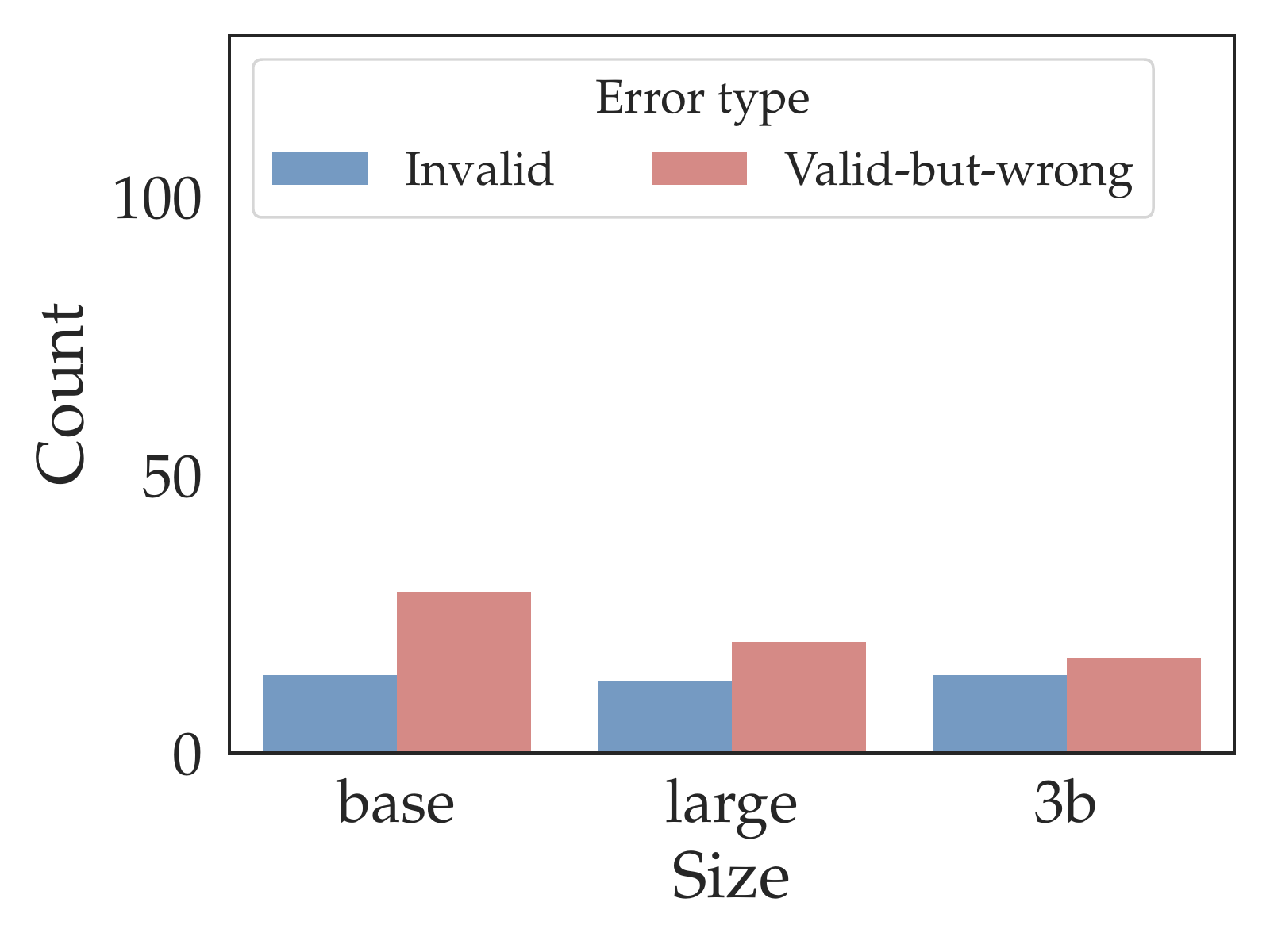}
    }
    \subfigure[FeTaQA]{
        \includegraphics[width=0.45\linewidth]{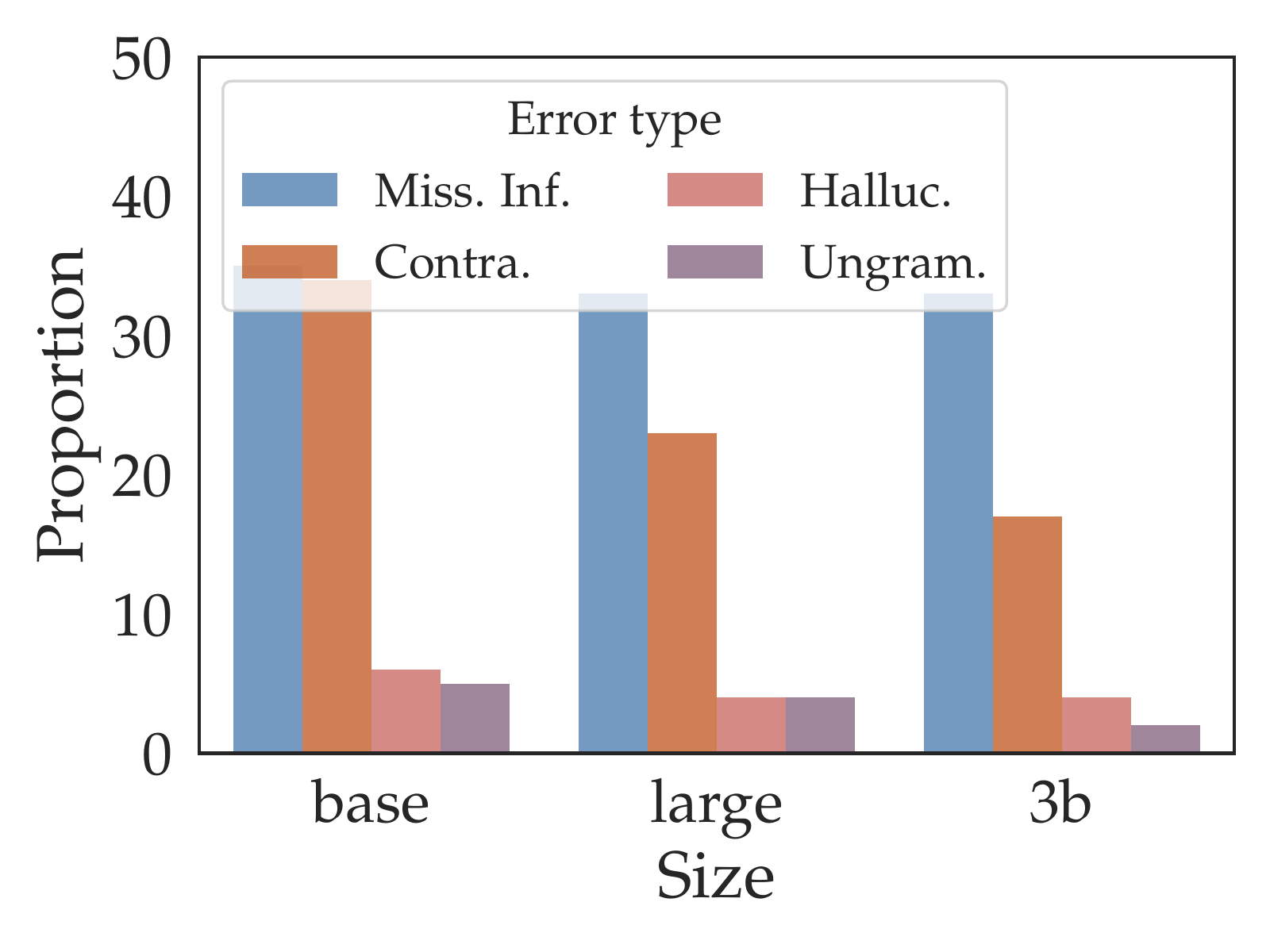}
    }
    \subfigure[ToTTo]{
        \includegraphics[width=0.45\linewidth]{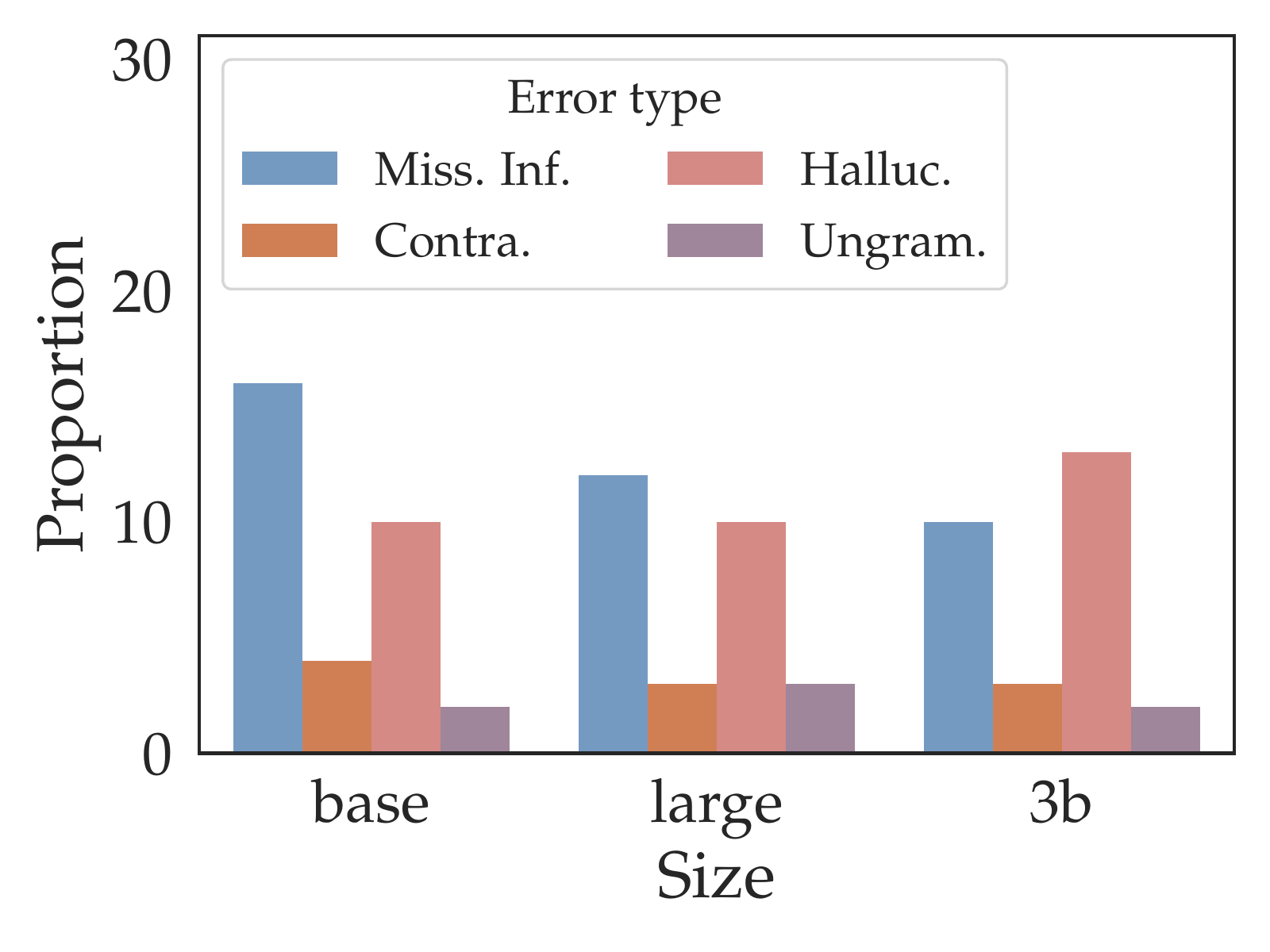}
    }
\caption{\label{fig:error-analysis} Error analysis. For semantic parsing, we plot the number of invalid/valid-but-wrong predictions. For generation, we plot the proportion of missing-information/contradiction/hallucination/ungrammatical errors among all predictions (one prediction may have multiple errors). Full visualization is in Appendix \ref{app:all-results}. } 
\end{figure}

\subsection{Human Evaluation for Generation Tasks}
\label{subsec:human-eval}
For each generation task, we randomly sample 100 development set samples and ask human annotators to judge the correctness of each output, using a 0-1 score. Details are provided in Appendix \ref{subapp:human-evaluation}. 
Table \ref{tab:human-eval} shows that automatic metrics do not always reflect human evaluation, calling for better automatic metrics to truly reflect the model's ability on generation tasks. 
Larger models are not always better, and detailed error analysis is provided below. 

\subsection{Error Analysis}
\label{subsec:error}

\noindent\textbf{Error analysis based on output validity \ \ }  
Unconstrained decoding from PLMs may generate \textit{invalid outputs}. 
For semantic parsing, we divide wrong outputs into \textit{invalid outputs} (i.e., not executable when the output is SQL, and not parse-able when the output is s-expression or TOP-representation) and \textit{valid but wrong answers}. 
Figure \ref{fig:error-analysis} shows that, for SQL semantic parsing, a large number of errors are caused by invalid outputs, and the number of invalid outputs gradually decreases with the increase of model size. This phenomenon is also observed by \citet{Scholak2021:PICARD}, who used constrained decoding to improve the validity, largely improving the parsing performance.
For s-expression semantic parsing, invalid outputs take up 30--50\% of all wrong outputs, and increasing the model size does not reduce invalidity significantly. 
For fact verification tasks, valid outputs are ``entailed'' and ``refuted''. 
We observe that T5 always generates valid outputs. 
For question answering, we do not include the validity analysis since the validity check for an answer is non-trivial and could be imprecise. 

\noindent\textbf{Error analysis for text generation tasks \ \ } 
For generation tasks, we consider four types of errors: \textit{missing information} (required information is not shown in the output), \textit{contradiction} (the output is contradictory to the input), 3) \textit{hallucination} (the output contains information that cannot be verified by the input), and 4) \textit{ungrammatical}. 
Figure \ref{fig:error-analysis} shows that the proportion of ungrammatical outputs is generally less than 5\%. 
Missing information and contradiction are common errors made by T5, and performance gains generally come from reducing contradiction. 
Hallucination is not a common error made by T5 except for the highlighted-table-to-text task (ToTTo), where T5 tends to output information of non-highlighted cell values.

\noindent\textbf{Case study \ \ }
We summarize some interesting observations about the model output (more in Appendix~\ref{app:case-study}).
Compared with T5-base and T5-large, T5-3B's outputs for text generation tasks tend to be more diverse and creative as shown in Appendix \ref{app:fetaqa_case} and \ref{app:totto_case}.
Also, T5-3B sometimes leverages domain knowledge to summarize facts in some tasks such as DART (e.g., describing \textit{rating 5 out of 5} as \textit{low}), while the other two copy the original expressions in the input, as shown in Appendix \ref{app:dart_case} and \ref{app:logic2text_case}.
However, this ability puts T5-3B in the risk of manipulating information and meaning of user request as shown in Appendix \ref{app:kvret_case2} and \ref{app:sql2text_case}. 

\section{Conclusions}
In this paper, we propose the \uskg framework to promote systematic research on structured knowledge grounding by unifying \ntasks \skg tasks. Using \uskg as a benchmark, we demonstrate that finetuning T5 on individual tasks achieves state-of-the-art results on almost all \ntasks tasks. 
We show that multi-task prefix-tuning benefits most \skg tasks, largely improving the overall performance. 
For structured knowledge encoding, we find that the effectiveness of encoding variations varies across tasks. 
Moreover, \uskg is a challenging testbed for zero-shot and few-shot learning, shown by the poor results of large PLMs. 

\section{Limitations}

\uskg establishes a powerful and reproducible starting point for \skg research. New models can be easily applied to diverse \skg tasks, and new tasks can be easily framed based on our standardized abstraction.
\uskg promotes a systematic study on more general and robust advances in structured knowledge encoding, multi-task learning, zero-shot learning, and few-shot learning for \skg tasks.
It also would be interesting to explore general pretraining methods within \uskg, which potentially benefit all the unified tasks. 
When the structured knowledge is too large for GPU memory, we truncate them based on heuristic rules,
calling for future study on 1) incorporating retrieval component in \skg, 
2) designing sparse attention in T5 for structured knowledge or other means to improve model efficiency. 

\uskg currently provides the correct type of structured knowledge for each task. 
However, how a system searches for the correct structured knowledge resources, takes appropriate action, and integrates information and results from multiple structured sources given a user request is still under-explored, which are a prerequisite for building a unified multi-purpose \skg system. 

Since we select popular tasks from each task family, we risk disproportionality in terms of the data language,  domain and population, and we actively welcome diverse, multi-lingual tasks to be added into \uskg. 
Also, the error analysis of \skg can more fine-grained, and we hope our findings can promote future work on systematically studying and decomposing the behavior of PLMs on \skg tasks. 
Furthermore, training and evaluation data should reflect the intents and linguistic phenomena in the real world \cite{Vries2020TowardsEV}, suggesting more realistic tasks to be added into \uskg. 

\clearpage
\bibliography{anthology,custom}

\begin{thebibliography}{104}
\expandafter\ifx\csname natexlab\endcsname\relax\def\natexlab#1{#1}\fi

\bibitem[{Agarwal et~al.(2020)Agarwal, Ge, Shakeri, and
  Al-Rfou}]{agarwal2020knowledge}
Oshin Agarwal, Heming Ge, Siamak Shakeri, and Rami Al-Rfou. 2020.
\newblock Knowledge graph based synthetic corpus generation for
  knowledge-enhanced language model pre-training.
\newblock \emph{arXiv preprint arXiv:2010.12688}.

\bibitem[{Aghajanyan et~al.(2021{\natexlab{a}})Aghajanyan, Gupta, Shrivastava,
  Chen, Zettlemoyer, and Gupta}]{muppet}
Armen Aghajanyan, Anchit Gupta, Akshat Shrivastava, Xilun Chen, Luke
  Zettlemoyer, and Sonal Gupta. 2021{\natexlab{a}}.
\newblock Muppet: Massive multi-task representations with pre-finetuning.
\newblock In \emph{Proceedings of EMNLP 2021}, pages 5799--5811, Online and
  Punta Cana, Dominican Republic.

\bibitem[{Aghajanyan et~al.(2021{\natexlab{b}})Aghajanyan, Okhonko, Lewis,
  Joshi, Xu, Ghosh, and Zettlemoyer}]{aghajanyan2021htlm}
Armen Aghajanyan, Dmytro Okhonko, Mike Lewis, Mandar Joshi, Hu~Xu, Gargi Ghosh,
  and Luke Zettlemoyer. 2021{\natexlab{b}}.
\newblock Htlm: Hyper-text pre-training and prompting of language models.

\bibitem[{Aly et~al.(2021)Aly, Guo, Schlichtkrull, Thorne, Vlachos,
  Christodoulopoulos, Cocarascu, and Mittal}]{aly2021fact}
Rami Aly, Zhijiang Guo, Michael~Sejr Schlichtkrull, James Thorne, Andreas
  Vlachos, Christos Christodoulopoulos, Oana Cocarascu, and Arpit Mittal. 2021.
\newblock The fact extraction and {VER}ification over unstructured and
  structured information ({FEVEROUS}) shared task.
\newblock In \emph{Proceedings of the Fourth Workshop on Fact Extraction and
  VERification (FEVER)}, pages 1--13.

\bibitem[{Aribandi et~al.(2021)Aribandi, Tay, Schuster, Rao, Zheng, Mehta,
  Zhuang, Tran, Bahri, Ni, Gupta, Hui, Ruder, and Metzler}]{aribandi2021ext5}
Vamsi Aribandi, Yi~Tay, Tal Schuster, Jinfeng Rao, Huaixiu~Steven Zheng,
  Sanket~Vaibhav Mehta, Honglei Zhuang, Vinh~Q. Tran, Dara Bahri, Jianmo Ni,
  Jai Gupta, Kai Hui, Sebastian Ruder, and Donald Metzler. 2021.
\newblock Ext5: Towards extreme multi-task scaling for transfer learning.

\bibitem[{Austin et~al.(2021)Austin, Odena, Nye, Bosma, Michalewski, Dohan,
  Jiang, Cai, Terry, Le, and Sutton}]{Austin2021ProgramSW}
Jacob Austin, Augustus Odena, Maxwell Nye, Maarten Bosma, Henryk Michalewski,
  David Dohan, Ellen Jiang, Carrie Cai, Michael Terry, Quoc~V. Le, and Charles
  Sutton. 2021.
\newblock Program synthesis with large language models.
\newblock \emph{ArXiv}, abs/2108.07732.

\bibitem[{Berant et~al.(2013)Berant, Chou, Frostig, and Liang}]{BerantCFL13}
Jonathan Berant, Andrew Chou, Roy Frostig, and Percy Liang. 2013.
\newblock Semantic parsing on freebase from question-answer pairs.
\newblock In \emph{{EMNLP} 2013}, pages 1533--1544.

\bibitem[{Berant and Liang(2014)}]{berant2014semantic}
Jonathan Berant and Percy Liang. 2014.
\newblock Semantic parsing via paraphrasing.
\newblock In \emph{Proceedings of ACL 2014}, pages 1415--1425.

\bibitem[{Bogin et~al.(2019)Bogin, Gardner, and Berant}]{Bogin2019GlobalRO}
Ben Bogin, Matt Gardner, and Jonathan Berant. 2019.
\newblock Global reasoning over database structures for text-to-sql parsing.
\newblock In \emph{Proceedings of EMNLP 2019}.

\bibitem[{Brown et~al.(2020)Brown, Mann, Ryder, Subbiah, Kaplan, Dhariwal,
  Neelakantan, Shyam, Sastry, Askell, Agarwal, Herbert-Voss, Krueger, Henighan,
  Child, Ramesh, Ziegler, Wu, Winter, Hesse, Chen, Sigler, Litwin, Gray, Chess,
  Clark, Berner, McCandlish, Radford, Sutskever, and
  Amodei}]{brown2020language}
Tom Brown, Benjamin Mann, Nick Ryder, Melanie Subbiah, Jared~D Kaplan, Prafulla
  Dhariwal, Arvind Neelakantan, Pranav Shyam, Girish Sastry, Amanda Askell,
  Sandhini Agarwal, Ariel Herbert-Voss, Gretchen Krueger, Tom Henighan, Rewon
  Child, Aditya Ramesh, Daniel Ziegler, Jeffrey Wu, Clemens Winter, Chris
  Hesse, Mark Chen, Eric Sigler, Mateusz Litwin, Scott Gray, Benjamin Chess,
  Jack Clark, Christopher Berner, Sam McCandlish, Alec Radford, Ilya Sutskever,
  and Dario Amodei. 2020.
\newblock Language {Models} are {Few}-{Shot} {Learners}.
\newblock In \emph{Advances in {Neural} {Information} {Processing} {Systems}},
  volume~33, pages 1877--1901.

\bibitem[{Budzianowski et~al.(2018)Budzianowski, Wen, Tseng, Casanueva, Stefan,
  Osman, and Ga{\v{s}}i\'c}]{budzianowski2018large}
Pawe{\l} Budzianowski, Tsung-Hsien Wen, Bo-Hsiang Tseng, I{\~n}igo Casanueva,
  Ultes Stefan, Ramadan Osman, and Milica Ga{\v{s}}i\'c. 2018.
\newblock Multiwoz - a large-scale multi-domain wizard-of-oz dataset for
  task-oriented dialogue modelling.
\newblock In \emph{Proceedings of the 2018 Conference on Empirical Methods in
  Natural Language Processing (EMNLP)}.

\bibitem[{Chen et~al.(2021{\natexlab{a}})Chen, Tworek, Jun, Yuan,
  de~Oliveira~Pinto, Kaplan, Edwards, Burda, Joseph, Brockman, Ray, Puri,
  Krueger, Petrov, Khlaaf, Sastry, Mishkin, Chan, Gray, Ryder, Pavlov, Power,
  Kaiser, Bavarian, Winter, Tillet, Such, Cummings, Plappert, Chantzis, Barnes,
  Herbert-Voss, Guss, Nichol, Paino, Tezak, Tang, Babuschkin, Balaji, Jain,
  Saunders, Hesse, Carr, Leike, Achiam, Misra, Morikawa, Radford, Knight,
  Brundage, Murati, Mayer, Welinder, McGrew, Amodei, McCandlish, Sutskever, and
  Zaremba}]{chen2021evaluating}
Mark Chen, Jerry Tworek, Heewoo Jun, Qiming Yuan, Henrique~Ponde
  de~Oliveira~Pinto, Jared Kaplan, Harri Edwards, Yuri Burda, Nicholas Joseph,
  Greg Brockman, Alex Ray, Raul Puri, Gretchen Krueger, Michael Petrov, Heidy
  Khlaaf, Girish Sastry, Pamela Mishkin, Brooke Chan, Scott Gray, Nick Ryder,
  Mikhail Pavlov, Alethea Power, Lukasz Kaiser, Mohammad Bavarian, Clemens
  Winter, Philippe Tillet, Felipe~Petroski Such, Dave Cummings, Matthias
  Plappert, Fotios Chantzis, Elizabeth Barnes, Ariel Herbert-Voss,
  William~Hebgen Guss, Alex Nichol, Alex Paino, Nikolas Tezak, Jie Tang, Igor
  Babuschkin, Suchir Balaji, Shantanu Jain, William Saunders, Christopher
  Hesse, Andrew~N. Carr, Jan Leike, Josh Achiam, Vedant Misra, Evan Morikawa,
  Alec Radford, Matthew Knight, Miles Brundage, Mira Murati, Katie Mayer, Peter
  Welinder, Bob McGrew, Dario Amodei, Sam McCandlish, Ilya Sutskever, and
  Wojciech Zaremba. 2021{\natexlab{a}}.
\newblock Evaluating large language models trained on code.

\bibitem[{Chen et~al.(2021{\natexlab{b}})Chen, Liu, Gao, Jiao, Zhang, and
  Ji}]{chen2020hitter}
Sanxing Chen, Xiaodong Liu, Jianfeng Gao, Jian Jiao, Ruofei Zhang, and Yangfeng
  Ji. 2021{\natexlab{b}}.
\newblock {H}itt{ER}: Hierarchical transformers for knowledge graph embeddings.
\newblock In \emph{Proceedings of the 2021 Conference on Empirical Methods in
  Natural Language Processing}, pages 10395--10407.

\bibitem[{Chen et~al.(2020{\natexlab{a}})Chen, Chen, Su, Chen, and
  Wang}]{chen2020logical}
Wenhu Chen, Jianshu Chen, Yu~Su, Zhiyu Chen, and William~Yang Wang.
  2020{\natexlab{a}}.
\newblock Logical natural language generation from open-domain tables.
\newblock In \emph{Proceedings of the 58th Annual Meeting of the Association
  for Computational Linguistics}, pages 7929--7942, Online. Association for
  Computational Linguistics.

\bibitem[{Chen et~al.(2020{\natexlab{b}})Chen, Wang, Chen, Zhang, Wang, Li,
  Zhou, and Wang}]{2019TabFactA}
Wenhu Chen, Hongmin Wang, Jianshu Chen, Yunkai Zhang, Hong Wang, Shiyang Li,
  Xiyou Zhou, and William~Yang Wang. 2020{\natexlab{b}}.
\newblock Tabfact : A large-scale dataset for table-based fact verification.
\newblock In \emph{International Conference on Learning Representations
  (ICLR)}, Addis Ababa, Ethiopia.

\bibitem[{Chen et~al.(2020{\natexlab{c}})Chen, Zha, Chen, Xiong, Wang, and
  Wang}]{chen2020hybridqa}
Wenhu Chen, Hanwen Zha, Zhiyu Chen, Wenhan Xiong, Hong Wang, and William Wang.
  2020{\natexlab{c}}.
\newblock Hybridqa: A dataset of multi-hop question answering over tabular and
  textual data.
\newblock \emph{Findings of EMNLP 2020}.

\bibitem[{Chen et~al.(2020{\natexlab{d}})Chen, Chen, Zha, Zhou, Zhang,
  Sundaresan, and Wang}]{chen-etal-2020-logic2text}
Zhiyu Chen, Wenhu Chen, Hanwen Zha, Xiyou Zhou, Yunkai Zhang, Sairam
  Sundaresan, and William~Yang Wang. 2020{\natexlab{d}}.
\newblock {L}ogic2{T}ext: High-fidelity natural language generation from
  logical forms.
\newblock In \emph{Findings of the Association for Computational Linguistics:
  EMNLP 2020}.

\bibitem[{Chen et~al.(2021{\natexlab{c}})Chen, Hellendoorn, Lamblin, Maniatis,
  Manzagol, Tarlow, and Moitra}]{chen2021plur}
Zimin Chen, Vincent~Josua Hellendoorn, Pascal Lamblin, Petros Maniatis,
  Pierre-Antoine Manzagol, Daniel Tarlow, and Subhodeep Moitra.
  2021{\natexlab{c}}.
\newblock {PLUR}: A unifying, graph-based view of program learning,
  understanding, and repair.
\newblock In \emph{Thirty-Fifth Conference on Neural Information Processing
  Systems}.

\bibitem[{Clive et~al.(2021)Clive, Cao, and Rei}]{Clive2021ControlPF}
Jordan Clive, Kris Cao, and Marek Rei. 2021.
\newblock Control prefixes for parameter-efficient text generation.

\bibitem[{Dai et~al.(2021)Dai, Li, Li, Sun, Huang, Si, and
  Zhu}]{dai-etal-2021-preview}
Yinpei Dai, Hangyu Li, Yongbin Li, Jian Sun, Fei Huang, Luo Si, and Xiaodan
  Zhu. 2021.
\newblock Preview, attend and review: Schema-aware curriculum learning for
  multi-domain dialogue state tracking.
\newblock In \emph{Proceedings ACL-IJCNLP 2021 (Volume 2: Short Papers)}, pages
  879--885, Online.

\bibitem[{Das et~al.(2021)Das, Zaheer, Thai, Godbole, Perez, Lee, Tan,
  Polymenakos, and McCallum}]{das2021case}
Rajarshi Das, Manzil Zaheer, Dung Thai, Ameya Godbole, Ethan Perez, Jay~Yoon
  Lee, Lizhen Tan, Lazaros Polymenakos, and Andrew McCallum. 2021.
\newblock Case-based reasoning for natural language queries over knowledge
  bases.
\newblock In \emph{Proceedings of EMNLP 2021}, pages 9594--9611, Online and
  Punta Cana, Dominican Republic.

\bibitem[{de~Vries et~al.(2020)de~Vries, Bahdanau, and
  Manning}]{Vries2020TowardsEV}
Harm de~Vries, Dzmitry Bahdanau, and Christopher~D. Manning. 2020.
\newblock Towards ecologically valid research on language user interfaces.
\newblock \emph{ArXiv}.

\bibitem[{Devlin et~al.(2017)Devlin, Uesato, Bhupatiraju, Singh, rahman
  Mohamed, and Kohli}]{Devlin2017RobustFillNP}
Jacob Devlin, Jonathan Uesato, Surya Bhupatiraju, Rishabh Singh, Abdel rahman
  Mohamed, and Pushmeet Kohli. 2017.
\newblock Robustfill: Neural program learning under noisy i/o.
\newblock In \emph{ICML}.

\bibitem[{Eisenschlos et~al.(2021)Eisenschlos, Gor, M{\"u}ller, and
  Cohen}]{eisenschlos2021mate}
Julian~Martin Eisenschlos, Maharshi Gor, Thomas M{\"u}ller, and William~W
  Cohen. 2021.
\newblock Mate: Multi-view attention for table transformer efficiency.
\newblock \emph{arXiv preprint arXiv:2109.04312}.

\bibitem[{Eric et~al.(2019)Eric, Goel, Paul, Sethi, Agarwal, Gao, and
  Hakkani-Tur}]{eric2019multiwoz}
Mihail Eric, Rahul Goel, Shachi Paul, Abhishek Sethi, Sanchit Agarwal, Shuyag
  Gao, and Dilek Hakkani-Tur. 2019.
\newblock Multiwoz 2.1: Multi-domain dialogue state corrections and state
  tracking baselines.
\newblock \emph{arXiv preprint arXiv:1907.01669}.

\bibitem[{Eric et~al.(2017)Eric, Krishnan, Charette, and
  Manning}]{Eric2017KeyValueRN}
Mihail Eric, Lakshmi. Krishnan, François Charette, and Christopher~D. Manning.
  2017.
\newblock Key-value retrieval networks for task-oriented dialogue.
\newblock In \emph{SIGDIAL Conference}.

\bibitem[{Gao et~al.(2021)Gao, Fisch, and Chen}]{gao2021making}
Tianyu Gao, Adam Fisch, and Danqi Chen. 2021.
\newblock Making pre-trained language models better few-shot learners.
\newblock In \emph{Association for Computational Linguistics (ACL)}.

\bibitem[{Gou et~al.(2021)Gou, Lei, Liu, Dai, and Shen}]{gou2021contextualize}
Yanjie Gou, Yinjie Lei, Lingqiao Liu, Yong Dai, and Chunxu Shen. 2021.
\newblock Contextualize knowledge bases with transformer for end-to-end
  task-oriented dialogue systems.
\newblock In \emph{Proceedings of the EMNLP 2021}, pages 4300--4310, Online and
  Punta Cana, Dominican Republic.

\bibitem[{Gu et~al.(2021)Gu, Kase, Vanni, Sadler, Liang, Yan, and
  Su}]{gu2021beyond}
Yu~Gu, Sue Kase, Michelle Vanni, Brian Sadler, Percy Liang, Xifeng Yan, and
  Yu~Su. 2021.
\newblock Beyond iid: three levels of generalization for question answering on
  knowledge bases.
\newblock In \emph{Proceedings of the Web Conference 2021}.

\bibitem[{Herzig et~al.(2020)Herzig, Nowak, M{\"u}ller, Piccinno, and
  Eisenschlos}]{Herzig2020tapas}
Jonathan Herzig, P.~Nowak, Thomas M{\"u}ller, Francesco Piccinno, and
  Julian~Martin Eisenschlos. 2020.
\newblock Tapas: Weakly supervised table parsing via pre-training.
\newblock In \emph{Proceedings of ACL}.

\bibitem[{Hosseini-Asl et~al.(2020)Hosseini-Asl, McCann, Wu, Yavuz, and
  Socher}]{hosseini2020simple}
Ehsan Hosseini-Asl, Bryan McCann, Chien-Sheng Wu, Semih Yavuz, and Richard
  Socher. 2020.
\newblock A simple language model for task-oriented dialogue.
\newblock In \emph{Proceedings of Conference on Neural Information Processing
  Systems (NeurIPS)}.

\bibitem[{Hui et~al.(2021)Hui, Geng, Ren, Li, Li, Sun, Huang, Si, Zhu, and
  Zhu}]{hui2021dynamic}
Binyuan Hui, Ruiying Geng, Qiyu Ren, Binhua Li, Yongbin Li, Jian Sun, Fei
  Huang, Luo Si, Pengfei Zhu, and Xiaodan Zhu. 2021.
\newblock Dynamic hybrid relation network for cross-domain context-dependent
  semantic parsing.

\bibitem[{Hwang et~al.(2019)Hwang, Yim, Park, and Seo}]{Hwang2019ACE}
Wonseok Hwang, Jinyeung Yim, Seunghyun Park, and Minjoon Seo. 2019.
\newblock A comprehensive exploration on wikisql with table-aware word
  contextualization.
\newblock \emph{ArXiv}, abs/1902.01069.

\bibitem[{Iida et~al.(2021)Iida, Thai, Manjunatha, and Iyyer}]{iida2021tabbie}
Hiroshi Iida, Dung Thai, Varun Manjunatha, and Mohit Iyyer. 2021.
\newblock Tabbie: Pretrained representations of tabular data.
\newblock \emph{arXiv preprint arXiv:2105.02584}.

\bibitem[{Iyyer et~al.(2017)Iyyer, Yih, and Chang}]{iyyer-etal-2017-search}
Mohit Iyyer, Wen-tau Yih, and Ming-Wei Chang. 2017.
\newblock Search-based neural structured learning for sequential question
  answering.
\newblock In \emph{Proceedings of the 55th Annual Meeting of the Association
  for Computational Linguistics (Volume 1: Long Papers)}, pages 1821--1831,
  Vancouver, Canada.

\bibitem[{Kale and Rastogi(2020)}]{KaleR20a}
Mihir Kale and Abhinav Rastogi. 2020.
\newblock Text-to-text pre-training for data-to-text tasks.
\newblock In \emph{Proceedings of {INLG} 2020, Dublin, Ireland, December 15-18,
  2020}, pages 97--102.

\bibitem[{Khashabi et~al.(2020)Khashabi, Min, Khot, Sabhwaral, Tafjord, Clark,
  and Hajishirzi}]{2020unifiedqa}
D.~Khashabi, S.~Min, T.~Khot, A.~Sabhwaral, O.~Tafjord, P.~Clark, and
  H.~Hajishirzi. 2020.
\newblock Unifiedqa: Crossing format boundaries with a single qa system.
\newblock \emph{EMNLP - findings}.

\bibitem[{Lee et~al.(2021)Lee, Cheng, and Ostendorf}]{lee-etal-2021-dialogue}
Chia-Hsuan Lee, Hao Cheng, and Mari Ostendorf. 2021.
\newblock Dialogue state tracking with a language model using schema-driven
  prompting.
\newblock In \emph{Proceedings of the 2021 Conference on Empirical Methods in
  Natural Language Processing}, pages 4937--4949.

\bibitem[{Lester et~al.(2021)Lester, Al{-}Rfou, and Constant}]{LesterAC21}
Brian Lester, Rami Al{-}Rfou, and Noah Constant. 2021.
\newblock The power of scale for parameter-efficient prompt tuning.
\newblock In \emph{Proceedings of {EMNLP} 2021, Virtual Event / Punta Cana,
  Dominican Republic, 7-11 November, 2021}, pages 3045--3059.

\bibitem[{Lhoest et~al.(2021)Lhoest, del Moral, Jernite, Thakur, von Platen,
  Patil, Chaumond, Drame, Plu, Tunstall, Davison, Šaško, Chhablani, Malik,
  Brandeis, Scao, Sanh, Xu, Patry, McMillan-Major, Schmid, Gugger, Delangue,
  Matussière, Debut, Bekman, Cistac, Goehringer, Mustar, Lagunas, Rush, and
  Wolf}]{lhoest2021datasets}
Quentin Lhoest, Albert~Villanova del Moral, Yacine Jernite, Abhishek Thakur,
  Patrick von Platen, Suraj Patil, Julien Chaumond, Mariama Drame, Julien Plu,
  Lewis Tunstall, Joe Davison, Mario Šaško, Gunjan Chhablani, Bhavitvya
  Malik, Simon Brandeis, Teven~Le Scao, Victor Sanh, Canwen Xu, Nicolas Patry,
  Angelina McMillan-Major, Philipp Schmid, Sylvain Gugger, Clément Delangue,
  Théo Matussière, Lysandre Debut, Stas Bekman, Pierric Cistac, Thibault
  Goehringer, Victor Mustar, François Lagunas, Alexander~M. Rush, and Thomas
  Wolf. 2021.
\newblock Datasets: A community library for natural language processing.

\bibitem[{Li et~al.(2021)Li, Arora, Chen, Gupta, Gupta, and
  Mehdad}]{li-etal-2021-mtop}
Haoran Li, Abhinav Arora, Shuohui Chen, Anchit Gupta, Sonal Gupta, and Yashar
  Mehdad. 2021.
\newblock {MTOP}: A comprehensive multilingual task-oriented semantic parsing
  benchmark.
\newblock In \emph{Proceedings of the 16th Conference of the European Chapter
  of the Association for Computational Linguistics: Main Volume}, pages
  2950--2962, Online.

\bibitem[{Li and Liang(2021)}]{li2021prefixtuning}
Xiang~Lisa Li and Percy Liang. 2021.
\newblock Prefix-tuning: Optimizing continuous prompts for generation.
\newblock In \emph{Proceedings of the 59th Annual Meeting of the Association
  for Computational Linguistics and the 11th International Joint Conference on
  Natural Language Processing (Volume 1: Long Papers)}, pages 4582--4597,
  Online.

\bibitem[{Lin et~al.(2019)Lin, Chen, Chen, and Ren}]{kagnet-emnlp19}
Bill~Yuchen Lin, Xinyue Chen, Jamin Chen, and Xiang Ren. 2019.
\newblock Kagnet: Knowledge-aware graph networks for commonsense reasoning.
\newblock In \emph{Proceedings of EMNLP-IJCNLP}.

\bibitem[{Lin et~al.(2021)Lin, Liu, Moon, Crook, Zhou, Wang, Yu, Madotto, Cho,
  and Subba}]{lin2021leveraging}
Zhaojiang Lin, Bing Liu, Seungwhan Moon, Paul~A Crook, Zhenpeng Zhou, Zhiguang
  Wang, Zhou Yu, Andrea Madotto, Eunjoon Cho, and Rajen Subba. 2021.
\newblock Leveraging slot descriptions for zero-shot cross-domain dialogue
  statetracking.
\newblock In \emph{Proceedings of NAACL 2021}, pages 5640--5648.

\bibitem[{Liu et~al.(2021)Liu, Chen, Guo, Lin, and guang Lou}]{liu2021tapex}
Qian Liu, Bei Chen, Jiaqi Guo, Zeqi Lin, and Jian guang Lou. 2021.
\newblock Tapex: Table pre-training via learning a neural sql executor.

\bibitem[{Liu et~al.(2020)Liu, Zhou, Zhao, Wang, Ju, Deng, and
  Wang}]{weijie2019kbert}
Weijie Liu, Peng Zhou, Zhe Zhao, Zhiruo Wang, Qi~Ju, Haotang Deng, and Ping
  Wang. 2020.
\newblock K-bert: Enabling language representation with knowledge graph.
\newblock In \emph{AAAI}.

\bibitem[{Madotto et~al.(2018)Madotto, Wu, and Fung}]{P18-1136}
Andrea Madotto, Chien-Sheng Wu, and Pascale Fung. 2018.
\newblock Mem2seq: Effectively incorporating knowledge bases into end-to-end
  task-oriented dialog systems.
\newblock In \emph{Proceedings of the 56th Annual Meeting of the Association
  for Computational Linguistics (Volume 1: Long Papers)}, pages 1468--1478.

\bibitem[{Marzoev et~al.(2020)Marzoev, Madden, Kaashoek, Cafarella, and
  Andreas}]{marzoev2020unnatural}
Alana Marzoev, Samuel Madden, M~Frans Kaashoek, Michael Cafarella, and Jacob
  Andreas. 2020.
\newblock Unnatural language processing: Bridging the gap between synthetic and
  natural language data.
\newblock \emph{arXiv preprint arXiv:2004.13645}.

\bibitem[{McCann et~al.(2018)McCann, Keskar, Xiong, and
  Socher}]{abs-1806-08730}
Bryan McCann, Nitish~Shirish Keskar, Caiming Xiong, and Richard Socher. 2018.
\newblock The natural language decathlon: Multitask learning as question
  answering.
\newblock \emph{CoRR}, abs/1806.08730.

\bibitem[{Nan et~al.(2021{\natexlab{a}})Nan, Hsieh, Mao, Lin, Verma, Zhang,
  Kryściński, Schoelkopf, Kong, Tang, Mutuma, Rosand, Trindade, Bandaru,
  Cunningham, Xiong, and Radev}]{nan2021feta}
Linyong Nan, Chiachun Hsieh, Ziming Mao, Xi~Victoria Lin, Neha Verma, Rui
  Zhang, Wojciech Kryściński, Nick Schoelkopf, Riley Kong, Xiangru Tang,
  Murori Mutuma, Ben Rosand, Isabel Trindade, Renusree Bandaru, Jacob
  Cunningham, Caiming Xiong, and Dragomir Radev. 2021{\natexlab{a}}.
\newblock Fetaqa: Free-form table question answering.
\newblock \emph{TACL}.

\bibitem[{Nan et~al.(2021{\natexlab{b}})Nan, Radev, Zhang, Rau, Sivaprasad,
  Hsieh, Tang, Vyas, Verma, Krishna, Liu, Irwanto, Pan, Rahman, Zaidi, Mutuma,
  Tarabar, Gupta, Yu, Tan, Lin, Xiong, Socher, and Rajani}]{nan2021dart}
Linyong Nan, Dragomir Radev, Rui Zhang, Amrit Rau, Abhinand Sivaprasad,
  Chiachun Hsieh, Xiangru Tang, Aadit Vyas, Neha Verma, Pranav Krishna,
  Yangxiaokang Liu, Nadia Irwanto, Jessica Pan, Faiaz Rahman, Ahmad Zaidi,
  Murori Mutuma, Yasin Tarabar, Ankit Gupta, Tao Yu, Yi~Chern Tan, Xi~Victoria
  Lin, Caiming Xiong, Richard Socher, and Nazneen~Fatema Rajani.
  2021{\natexlab{b}}.
\newblock Dart: Open-domain structured data record to text generation.
\newblock In \emph{NAACL}.

\bibitem[{Novikova et~al.(2017)Novikova, Dusek, and Rieser}]{NovikovaDR17}
Jekaterina Novikova, Ondrej Dusek, and Verena Rieser. 2017.
\newblock The {E2E} dataset: New challenges for end-to-end generation.
\newblock In \emph{SIGDial 2017}, pages 201--206.

\bibitem[{Oguz et~al.(2021)Oguz, Chen, Karpukhin, Peshterliev, Okhonko,
  Schlichtkrull, Gupta, Mehdad, and Yih}]{oguz2021unik}
Barlas Oguz, Xilun Chen, Vladimir Karpukhin, Stan Peshterliev, Dmytro Okhonko,
  Michael Schlichtkrull, Sonal Gupta, Yashar Mehdad, and Scott Yih. 2021.
\newblock Unik-qa: Unified representations of structured and unstructured
  knowledge for open-domain question answering.
\newblock \emph{arXiv preprint arXiv:2012.14610}.

\bibitem[{Parikh et~al.(2020)Parikh, Wang, Gehrmann, Faruqui, Dhingra, Yang,
  and Das}]{parikh2020totto}
Ankur~P Parikh, Xuezhi Wang, Sebastian Gehrmann, Manaal Faruqui, Bhuwan
  Dhingra, Diyi Yang, and Dipanjan Das. 2020.
\newblock {ToTTo}: A controlled table-to-text generation dataset.
\newblock In \emph{Proceedings of EMNLP}.

\bibitem[{Pasupat and Liang(2015)}]{pasupat-liang-2015-compositional}
Panupong Pasupat and Percy Liang. 2015.
\newblock Compositional semantic parsing on semi-structured tables.
\newblock In \emph{Proceedings of the 53rd Annual Meeting of the Association
  for Computational Linguistics and the 7th International Joint Conference on
  Natural Language Processing (Volume 1: Long Papers)}, pages 1470--1480,
  Beijing, China.

\bibitem[{Pasupat et~al.(2021)Pasupat, Zhang, and
  Guu}]{pasupat-etal-2021-controllable}
Panupong Pasupat, Yuan Zhang, and Kelvin Guu. 2021.
\newblock Controllable semantic parsing via retrieval augmentation.
\newblock In \emph{Proceedings of EMNLP 2021}, pages 7683--7698, Online and
  Punta Cana, Dominican Republic.

\bibitem[{Post(2018)}]{post-2018-call}
Matt Post. 2018.
\newblock A call for clarity in reporting {BLEU} scores.
\newblock In \emph{Proceedings of the Third Conference on Machine Translation:
  Research Papers}, pages 186--191, Belgium, Brussels.

\bibitem[{Qin et~al.(2020)Qin, Xu, Che, Zhang, and Liu}]{qin-etal-2020-dynamic}
Libo Qin, Xiao Xu, Wanxiang Che, Yue Zhang, and Ting Liu. 2020.
\newblock Dynamic fusion network for multi-domain end-to-end task-oriented
  dialog.
\newblock In \emph{Proceedings of the 58th Annual Meeting of the Association
  for Computational Linguistics}, pages 6344--6354, Online.

\bibitem[{Radford et~al.(2019)Radford, Wu, Child, Luan, Amodei, and
  Sutskever}]{radford2019language}
Alec Radford, Jeff Wu, Rewon Child, David Luan, Dario Amodei, and Ilya
  Sutskever. 2019.
\newblock Language models are unsupervised multitask learners.

\bibitem[{Raffel et~al.(2020)Raffel, Shazeer, Roberts, Lee, Narang, Matena,
  Zhou, Li, and Liu}]{2020t5}
Colin Raffel, Noam Shazeer, Adam Roberts, Katherine Lee, Sharan Narang, Michael
  Matena, Yanqi Zhou, Wei Li, and Peter~J. Liu. 2020.
\newblock Exploring the limits of transfer learning with a unified text-to-text
  transformer.
\newblock \emph{Journal of Machine Learning Research}, 21(140):1--67.

\bibitem[{Reimers and Gurevych(2020)}]{reimers-2020-multilingual-sentence-bert}
Nils Reimers and Iryna Gurevych. 2020.
\newblock Making monolingual sentence embeddings multilingual using knowledge
  distillation.
\newblock In \emph{Proceedings of EMNLP 2020}.

\bibitem[{Ren et~al.(2021)Ren, Dai, Dai, Chen, Yasunaga, Sun, Schuurmans,
  Leskovec, and Zhou}]{ren2021lego}
Hongyu Ren, Hanjun Dai, Bo~Dai, Xinyun Chen, Michihiro Yasunaga, Haitian Sun,
  Dale Schuurmans, Jure Leskovec, and Denny Zhou. 2021.
\newblock Lego: Latent execution-guided reasoning for multi-hop question
  answering on knowledge graphs.
\newblock In \emph{International Conference on Machine Learning (ICML)}.

\bibitem[{Rubin and Berant(2021)}]{rubin-berant-2021-smbop}
Ohad Rubin and Jonathan Berant. 2021.
\newblock {S}m{B}o{P}: Semi-autoregressive bottom-up semantic parsing.
\newblock In \emph{Proceedings of NAACL 2021}, pages 311--324, Online.

\bibitem[{Sanh et~al.(2021)Sanh, Webson, Raffel, Bach, Sutawika, Alyafeai,
  Chaffin, Stiegler, Scao, Raja, Dey, Bari, Xu, Thakker, Sharma, Szczechla,
  Kim, Chhablani, Nayak, Datta, Chang, Jiang, Wang, Manica, Shen, Yong, Pandey,
  Bawden, Wang, Neeraj, Rozen, Sharma, Santilli, Fevry, Fries, Teehan,
  Biderman, Gao, Bers, Wolf, and Rush}]{sanh2021multitask}
Victor Sanh, Albert Webson, Colin Raffel, Stephen~H. Bach, Lintang Sutawika,
  Zaid Alyafeai, Antoine Chaffin, Arnaud Stiegler, Teven~Le Scao, Arun Raja,
  Manan Dey, M~Saiful Bari, Canwen Xu, Urmish Thakker, Shanya~Sharma Sharma,
  Eliza Szczechla, Taewoon Kim, Gunjan Chhablani, Nihal Nayak, Debajyoti Datta,
  Jonathan Chang, Mike Tian-Jian Jiang, Han Wang, Matteo Manica, Sheng Shen,
  Zheng~Xin Yong, Harshit Pandey, Rachel Bawden, Thomas Wang, Trishala Neeraj,
  Jos Rozen, Abheesht Sharma, Andrea Santilli, Thibault Fevry, Jason~Alan
  Fries, Ryan Teehan, Stella Biderman, Leo Gao, Tali Bers, Thomas Wolf, and
  Alexander~M. Rush. 2021.
\newblock Multitask prompted training enables zero-shot task generalization.

\bibitem[{Saxena et~al.(2020)Saxena, Tripathi, and
  Talukdar}]{saxena2020improving}
Apoorv Saxena, Aditay Tripathi, and Partha Talukdar. 2020.
\newblock Improving multi-hop question answering over knowledge graphs using
  knowledge base embeddings.
\newblock In \emph{Association for Computational Linguistics (ACL)}.

\bibitem[{Scholak et~al.(2021)Scholak, Schucher, and
  Bahdanau}]{Scholak2021:PICARD}
Torsten Scholak, Nathan Schucher, and Dzmitry Bahdanau. 2021.
\newblock {PICARD}: Parsing incrementally for constrained auto-regressive
  decoding from language models.
\newblock In \emph{Proceedings of EMNLP 2021}, pages 9895--9901.

\bibitem[{Shaw et~al.(2021)Shaw, Chang, Pasupat, and
  Toutanova}]{Shaw2021CompositionalGA}
Peter Shaw, Ming-Wei Chang, Panupong Pasupat, and Kristina Toutanova. 2021.
\newblock Compositional generalization and natural language variation: Can a
  semantic parsing approach handle both?
\newblock In \emph{ACL/IJCNLP}.

\bibitem[{Shin et~al.(2021)Shin, Lin, Thomson, Chen, Roy, Platanios, Pauls,
  Klein, Eisner, and Van~Durme}]{shin2021constrained}
Richard Shin, Christopher~H Lin, Sam Thomson, Charles Chen, Subhro Roy,
  Emmanouil~Antonios Platanios, Adam Pauls, Dan Klein, Jason Eisner, and
  Benjamin Van~Durme. 2021.
\newblock Constrained language models yield few-shot semantic parsers.
\newblock \emph{arXiv preprint arXiv:2104.08768}.

\bibitem[{Shu et~al.(2021)Shu, Zhang, Dong, Shi, Yu, and
  Zhang}]{shu-etal-2021-logic}
Chang Shu, Yusen Zhang, Xiangyu Dong, Peng Shi, Tao Yu, and Rui Zhang. 2021.
\newblock Logic-consistency text generation from semantic parses.
\newblock In \emph{Findings of the Association for Computational Linguistics:
  ACL-IJCNLP 2021}.

\bibitem[{Su et~al.(2021)Su, Vandyke, Wang, Fang, and
  Collier}]{Su2021PlanthenGenerateCD}
Yixuan Su, David Vandyke, Sihui Wang, Yimai Fang, and Nigel Collier. 2021.
\newblock Plan-then-generate: Controlled data-to-text generation via planning.
\newblock In \emph{EMNLP}.

\bibitem[{Talmor and Berant(2018)}]{talmor18compwebq}
A.~Talmor and J.~Berant. 2018.
\newblock The web as a knowledge-base for answering complex questions.
\newblock In \emph{North American Association for Computational Linguistics
  (NAACL)}.

\bibitem[{Talmor et~al.(2021)Talmor, Yoran, Catav, Lahav, Wang, Asai, Ilharco,
  Hajishirzi, and Berant}]{talmor2021multimodalqa}
Alon Talmor, Ori Yoran, Amnon Catav, Dan Lahav, Yizhong Wang, Akari Asai,
  Gabriel Ilharco, Hannaneh Hajishirzi, and Jonathan Berant. 2021.
\newblock Multimodal{\{}qa{\}}: complex question answering over text, tables
  and images.
\newblock In \emph{International Conference on Learning Representations}.

\bibitem[{Vu et~al.(2021)Vu, Lester, Constant, Al-Rfou, and Cer}]{vu2021spot}
Tu~Vu, Brian Lester, Noah Constant, Rami Al-Rfou, and Daniel Cer. 2021.
\newblock Spot: Better frozen model adaptation through soft prompt transfer.

\bibitem[{Wang et~al.(2020)Wang, Shin, Liu, Polozov, and
  Richardson}]{Wang2020RATSQLRS}
Bailin Wang, Richard Shin, Xiaodong Liu, Oleksandr Polozov, and Matthew
  Richardson. 2020.
\newblock Rat-sql: Relation-aware schema encoding and linking for text-to-sql
  parsers.
\newblock In \emph{ACL}.

\bibitem[{Wang et~al.(2019)Wang, Titov, and Lapata}]{wang-etal-2019-learning}
Bailin Wang, Ivan Titov, and Mirella Lapata. 2019.
\newblock Learning semantic parsers from denotations with latent structured
  alignments and abstract programs.
\newblock In \emph{Proceedings of EMNLP-IJCNLP 2019}, pages 3774--3785, Hong
  Kong, China.

\bibitem[{Wang et~al.(2017)Wang, Mao, Wang, and Guo}]{wang2017survey}
Quan Wang, Zhendong Mao, Bin Wang, and Li~Guo. 2017.
\newblock Knowledge graph embedding: A survey of approaches and applications.
\newblock \emph{IEEE Transactions on Knowledge and Data Engineering},
  29(12):2724--2743.

\bibitem[{Wang et~al.(2021{\natexlab{a}})Wang, Fang, Khabsa, Mao, and
  Ma}]{abs-2104-14690}
Sinong Wang, Han Fang, Madian Khabsa, Hanzi Mao, and Hao Ma.
  2021{\natexlab{a}}.
\newblock Entailment as few-shot learner.
\newblock \emph{CoRR}, abs/2104.14690.

\bibitem[{Wang et~al.(2021{\natexlab{b}})Wang, Dong, Jia, Li, Fu, Han, and
  Zhang}]{wang2021tuta}
Zhiruo Wang, Haoyu Dong, Ran Jia, Jia Li, Zhiyi Fu, Shi Han, and Dongmei Zhang.
  2021{\natexlab{b}}.
\newblock Tuta: Tree-based transformers for generally structured table
  pre-training.
\newblock In \emph{Proceedings of the 27th ACM SIGKDD Conference on Knowledge
  Discovery \& Data Mining}, pages 1780--1790.

\bibitem[{Wei et~al.(2021)Wei, Bosma, Zhao, Guu, Yu, Lester, Du, Dai, and
  Le}]{Jason2021}
Jason Wei, Maarten Bosma, Vincent~Y. Zhao, Kelvin Guu, Adams~Wei Yu, Brian
  Lester, Nan Du, Andrew~M. Dai, and Quoc~V. Le. 2021.
\newblock Finetuned language models are zero-shot learners.
\newblock \emph{arXiv preprint}.

\bibitem[{Williams et~al.(2016)Williams, Raux, and
  Henderson}]{williams2016dialog}
Jason~D Williams, Antoine Raux, and Matthew Henderson. 2016.
\newblock The dialog state tracking challenge series: A review.
\newblock \emph{Dialogue \& Discourse}, 7(3):4--33.

\bibitem[{Wolf et~al.(2020)Wolf, Debut, Sanh, Chaumond, Delangue, Moi, Cistac,
  Rault, Louf, Funtowicz, Davison, Shleifer, von Platen, Ma, Jernite, Plu, Xu,
  Scao, Gugger, Drame, Lhoest, and Rush}]{wolf-etal-2020-transformers}
Thomas Wolf, Lysandre Debut, Victor Sanh, Julien Chaumond, Clement Delangue,
  Anthony Moi, Pierric Cistac, Tim Rault, Rémi Louf, Morgan Funtowicz, Joe
  Davison, Sam Shleifer, Patrick von Platen, Clara Ma, Yacine Jernite, Julien
  Plu, Canwen Xu, Teven~Le Scao, Sylvain Gugger, Mariama Drame, Quentin Lhoest,
  and Alexander~M. Rush. 2020.
\newblock Transformers: State-of-the-art natural language processing.
\newblock In \emph{Proceedings of EMNLP 2020: System Demonstrations}, pages
  38--45, Online.

\bibitem[{Wu et~al.(2019)Wu, Socher, and Xiong}]{wu2019global}
Chien-Sheng Wu, Richard Socher, and Caiming Xiong. 2019.
\newblock Global-to-local memory pointer networks for task-oriented dialogue.
\newblock In \emph{Proceedings of the International Conference on Learning
  Representations (ICLR)}.

\bibitem[{Yang et~al.(2020)Yang, Nie, Feng, Liu, Chen, and
  Zhu}]{yang-etal-2020-program}
Xiaoyu Yang, Feng Nie, Yufei Feng, Quan Liu, Zhigang Chen, and Xiaodan Zhu.
  2020.
\newblock Program enhanced fact verification with verbalization and graph
  attention network.
\newblock In \emph{Proceedings of EMNLP 2020}, pages 7810--7825, Online.

\bibitem[{Yasunaga et~al.(2022)Yasunaga, Bosselut, Ren, Zhang, Manning, Liang,
  and Leskovec}]{yasunaga2022dragon}
Michihiro Yasunaga, Antoine Bosselut, Hongyu Ren, Xikun Zhang, Christopher~D.
  Manning, Percy Liang, and Jure Leskovec. 2022.
\newblock Deep bidirectional language-knowledge graph pretraining.
\newblock In \emph{Neural Information Processing Systems (NeurIPS)}.

\bibitem[{Yasunaga and Liang(2020)}]{yasunaga2020graph}
Michihiro Yasunaga and Percy Liang. 2020.
\newblock Graph-based, self-supervised program repair from diagnostic feedback.
\newblock In \emph{International Conference on Machine Learning (ICML)}.

\bibitem[{Yasunaga et~al.(2021)Yasunaga, Ren, Bosselut, Liang, and
  Leskovec}]{yasunaga-etal-2021-qa}
Michihiro Yasunaga, Hongyu Ren, Antoine Bosselut, Percy Liang, and Jure
  Leskovec. 2021.
\newblock {QA}-{GNN}: Reasoning with language models and knowledge graphs for
  question answering.
\newblock In \emph{North American Chapter of the Association for Computational
  Linguistics: Human Language Technologies (NAACL)}. Association for
  Computational Linguistics.

\bibitem[{Ye et~al.(2021{\natexlab{a}})Ye, Lin, and Ren}]{Ye2021CrossFitAF}
Qinyuan Ye, Bill~Yuchen Lin, and Xiang Ren. 2021{\natexlab{a}}.
\newblock Crossfit: A few-shot learning challenge for cross-task generalization
  in nlp.
\newblock In \emph{Proceedings of EMNLP}.

\bibitem[{Ye et~al.(2021{\natexlab{b}})Ye, Yavuz, Hashimoto, Zhou, and
  Xiong}]{ye2021rng}
Xi~Ye, Semih Yavuz, Kazuma Hashimoto, Yingbo Zhou, and Caiming Xiong.
  2021{\natexlab{b}}.
\newblock Rng-kbqa: Generation augmented iterative ranking for knowledge base
  question answering.
\newblock \emph{arXiv preprint arXiv:2109.08678}.

\bibitem[{Yih et~al.(2016)Yih, Richardson, Meek, Chang, and
  Suh}]{yih-etal-2016-value}
Wen-tau Yih, Matthew Richardson, Chris Meek, Ming-Wei Chang, and Jina Suh.
  2016.
\newblock The value of semantic parse labeling for knowledge base question
  answering.
\newblock In \emph{Proceedings of the 54th Annual Meeting of the Association
  for Computational Linguistics (Volume 2: Short Papers)}.

\bibitem[{Yin and Neubig(2018)}]{YinN18}
Pengcheng Yin and Graham Neubig. 2018.
\newblock {TRANX:} {A} transition-based neural abstract syntax parser for
  semantic parsing and code generation.
\newblock In \emph{{EMNLP} 2018: System Demonstrations}, pages 7--12.

\bibitem[{Yin et~al.(2020{\natexlab{a}})Yin, Neubig, Yih, and
  Riedel}]{yin20tabert}
Pengcheng Yin, Graham Neubig, Wen-tau Yih, and Sebastian Riedel.
  2020{\natexlab{a}}.
\newblock {T}a{BERT}: Pretraining for joint understanding of textual and
  tabular data.
\newblock In \emph{Proceedings of the 58th Annual Meeting of the Association
  for Computational Linguistics}. Association for Computational Linguistics.

\bibitem[{Yin et~al.(2020{\natexlab{b}})Yin, Rajani, Radev, Socher, and
  Xiong}]{YinRRSX20}
Wenpeng Yin, Nazneen~Fatema Rajani, Dragomir~R. Radev, Richard Socher, and
  Caiming Xiong. 2020{\natexlab{b}}.
\newblock Universal natural language processing with limited annotations: Try
  few-shot textual entailment as a start.
\newblock In \emph{Proceedings of {EMNLP} 2020, Online, November 16-20, 2020},
  pages 8229--8239.

\bibitem[{Yoran et~al.(2021)Yoran, Talmor, and Berant}]{yoran2021turning}
Ori Yoran, Alon Talmor, and Jonathan Berant. 2021.
\newblock Turning tables: Generating examples from semi-structured tables for
  endowing language models with reasoning skills.
\newblock \emph{arXiv preprint arXiv:2107.07261}.

\bibitem[{Yu et~al.(2019{\natexlab{a}})Yu, Zhang, Er, Li, Xue, Pang, Lin, Tan,
  Shi, Li, Jiang, Yasunaga, Shim, Chen, Fabbri, Li, Chen, Zhang, Dixit, Zhang,
  Xiong, Socher, Lasecki, and Radev}]{yu-etal-2019-cosql}
Tao Yu, Rui Zhang, Heyang Er, Suyi Li, Eric Xue, Bo~Pang, Xi~Victoria Lin,
  Yi~Chern Tan, Tianze Shi, Zihan Li, Youxuan Jiang, Michihiro Yasunaga,
  Sungrok Shim, Tao Chen, Alexander Fabbri, Zifan Li, Luyao Chen, Yuwen Zhang,
  Shreya Dixit, Vincent Zhang, Caiming Xiong, Richard Socher, Walter Lasecki,
  and Dragomir Radev. 2019{\natexlab{a}}.
\newblock {C}o{SQL}: A conversational text-to-{SQL} challenge towards
  cross-domain natural language interfaces to databases.
\newblock In \emph{Proceedings of EMNLP 2019}, pages 1962--1979, Hong Kong,
  China.

\bibitem[{Yu et~al.(2021)Yu, Zhang, Polozov, Meek, and Awadallah}]{yu2021SCoRE}
Tao Yu, Rui Zhang, Oleksandr Polozov, Christopher Meek, and Ahmed~Hassan
  Awadallah. 2021.
\newblock {SCoRE}: Pre-training for context representation in conversational
  semantic parsing.
\newblock In \emph{International Conference on Learning Representations}.

\bibitem[{Yu et~al.(2018)Yu, Zhang, Yang, Yasunaga, Wang, Li, Ma, Li, Yao,
  Roman, Zhang, and Radev}]{Yu18c}
Tao Yu, Rui Zhang, Kai Yang, Michihiro Yasunaga, Dongxu Wang, Zifan Li, James
  Ma, Irene Li, Qingning Yao, Shanelle Roman, Zilin Zhang, and Dragomir Radev.
  2018.
\newblock Spider: A large-scale human-labeled dataset for complex and
  cross-domain semantic parsing and text-to-sql task.
\newblock In \emph{Proceedings of EMNLP 2018}, Brussels, Belgium.

\bibitem[{Yu et~al.(2019{\natexlab{b}})Yu, Zhang, Yasunaga, Tan, Lin, Li,
  Heyang~Er, Pang, Chen, Ji, Dixit, Proctor, Shim, Jonathan~Kraft, Xiong,
  Socher, and Radev}]{Yu19}
Tao Yu, Rui Zhang, Michihiro Yasunaga, Yi~Chern Tan, Xi~Victoria Lin, Suyi Li,
  Irene~Li Heyang~Er, Bo~Pang, Tao Chen, Emily Ji, Shreya Dixit, David Proctor,
  Sungrok Shim, Vincent~Zhang Jonathan~Kraft, Caiming Xiong, Richard Socher,
  and Dragomir Radev. 2019{\natexlab{b}}.
\newblock Sparc: Cross-domain semantic parsing in context.
\newblock In \emph{Proceedings of the 57th Annual Meeting of the Association
  for Computational Linguistics}, Florence, Italy.

\bibitem[{Zaheer et~al.(2020)Zaheer, Guruganesh, Dubey, Ainslie, Alberti,
  Ontanon, Pham, Ravula, Wang, Yang, and Ahmed}]{Zaheer2020BigBT}
Manzil Zaheer, Guru Guruganesh, Kumar~Avinava Dubey, Joshua Ainslie, Chris
  Alberti, Santiago Ontanon, Philip Pham, Anirudh Ravula, Qifan Wang, Li~Yang,
  and Amr Ahmed. 2020.
\newblock Big {Bird}: {Transformers} for {Longer} {Sequences}.
\newblock In \emph{Advances in {Neural} {Information} {Processing} {Systems}},
  volume~33, pages 17283--17297.

\bibitem[{Zelle and Mooney(1996)}]{ZelleM96}
John~M. Zelle and Raymond~J. Mooney. 1996.
\newblock Learning to parse database queries using inductive logic programming.
\newblock In \emph{{AAAI} 1996}, pages 1050--1055.

\bibitem[{Zettlemoyer and Collins(2005)}]{Zettlemoyer05}
Luke~S. Zettlemoyer and Michael Collins. 2005.
\newblock Learning to map sentences to logical form: Structured classification
  with probabilistic categorial grammars.
\newblock \emph{UAI}.

\bibitem[{Zhang et~al.(2020)Zhang, Wang, Wang, Cao, Zhang, and
  Wang}]{zhang2020table}
Hongzhi Zhang, Yingyao Wang, Sirui Wang, Xuezhi Cao, Fuzheng Zhang, and
  Zhongyuan Wang. 2020.
\newblock Table fact verification with structure-aware transformer.
\newblock In \emph{Proceedings of EMNLP 2020}, pages 1624--1629.

\bibitem[{Zhong et~al.(2021)Zhong, Lee, Zhang, and Klein}]{Zhong2021AdaptingLM}
Ruiqi Zhong, Kristy Lee, Zheng Zhang, and Dan Klein. 2021.
\newblock Adapting language models for zero-shot learning by meta-tuning on
  dataset and prompt collections.
\newblock In \emph{Findings of EMNLP}.

\bibitem[{Zhong et~al.(2020)Zhong, Yu, and Klein}]{ruiqi20}
Ruiqi Zhong, Tao Yu, and Dan Klein. 2020.
\newblock Semantic evaluation for text-to-sql with distilled test suite.
\newblock In \emph{EMNLP 2020}.

\bibitem[{Zhong et~al.(2017)Zhong, Xiong, and Socher}]{zhongSeq2SQL2017}
Victor Zhong, Caiming Xiong, and Richard Socher. 2017.
\newblock Seq2sql: Generating structured queries from natural language using
  reinforcement learning.
\newblock \emph{CoRR}, abs/1709.00103.

\end{thebibliography}
\bibliographystyle{acl_natbib}

\clearpage

\appendix

\section{Contributions}
\label{sec:contributions}

\noindent\textbf{Code implementation \ \ \ } Tianbao Xie and Chen Henry Wu implemented the code base of the \uskg framework and experiment pipeline. 
The code of PICARD and advice from Torsten Scholak sped up the implementation. 

\noindent\textbf{Task unification \ \ } Tianbao Xie, Peng Shi, Michihiro Yasunaga, Chen Henry Wu, and Ming Zhong implemented the \ntasks tasks into the text-to-text format, adapted the metrics, and verified the performances. 

\noindent\textbf{Paper writing \ \ } Chen Henry Wu and Tianbao Xie finished most part of the paper. 
Michihiro Yasunaga, Peng Shi, and Chengzu Li added results and analysis for their corresponding parts. 
Peng Shi drafted related work on SKG with PLMs. 
Torsten Scholak, Pengcheng Yin, Rui Zhang, Ruiqi Zhong, Victor Zhong, Michihiro Yasunaga, Connor Boyle, Chien-Sheng Wu, Sida Wang, Bailin Wang, Ansong Ni, Ziyu Yao, Lingpeng Kong, Caiming Xiong, Dragomir Radev, Noah A. Smith, and Luke Zettlemoyer carefully reviewed the paper and gave feedback for multiple rounds. 

\noindent\textbf{Experiments \ \ } Chen Henry Wu, Tianbao Xie, and Chien-Sheng Wu conducted experiments on individual tasks and multi-task learning. Tianbao conducted the zero-shot learning experiments. Chengzu Li and Tianbao Xie conducted the few-shot learning experiments. Tianbao Xie conducted experiments on the ordering of sequence inputs and order-sensitivity. 
Chengzu Li, Connor Boyle, and Peng Shi conducted the experiments on converting structured knowledge into natural language. 

\noindent\textbf{Human evaluation \ \ } Chen Henry Wu organized the human evaluation. Torsten Scholak, Rui Zhang, Chengzu Li, Connor Boyle, Tianbao Xie, Peng Shi, Tao Yu, and Chen Henry Wu were the human participants. 

\noindent\textbf{Error analysis and case study \ \ } Tianbao Xie, Chen Henry Wu, and Michihiro Yasunaga designed and conducted the error analysis for semantic parsing and generation tasks. Authors who participated in the human annotation selected the cases for case study. 

\noindent\textbf{Discussion \ \ } 
We had three separate weekly meetings, and everyone in the project attended one of them. 
Torsten Scholak, Ruiqi Zhong, Pengcheng Yin, Victor Zhong, Peng Shi, Rui Zhang, Sida Wang, and Lingpeng Kong actively provided advice.
Torsten Scholak provided signals that prefix-tuning would be comparable to fine-tuning.
Ruiqi Zhong gave advice on analyzing the effect of model size,
Pengcheng Yin and Peng Shi gave advice on analysis on converting structured knowledge into natural language.
Pengcheng Yin helped interpret experimental results.
Ziyu Yao suggested that we report both sota (w/ extra) and sota (w/o extra) for a fair comparison. 
Victor Zhong and Bailin Wang gave valuable suggestions on multi-task learning and task transfer analysis. 
Luke Zettlemoyer, Noah A. Smith, Caiming Xiong, and Dragomir Radev gave valuable comments on research questions and experimental design.

\noindent\textbf{Computing resources \ \ } We thank Salesforce Research, an Amazon Research Award, ServiceNow Research, and Yale NLP for providing computing resources generously. 

\noindent Tao Yu designed and led the research. 

\section*{Acknowledgments}
We thank Yifei Min and Libo Qin for their early-stage discussion. We thank Panupong Pasupat and William W. Cohen for their valuable feedback on our initial draft. We thank Qian Liu for his \textsc{TaPeX} code and advice on question answering tasks. 
We thank wandb for free logging and OpenAI for free Codex usage. 

\clearpage

\section{Results with Full Metrics}
\label{app:all-results}
\begin{table}[ht]
	\centering
	\begin{adjustbox}{width=\columnwidth}
		\begin{tabular}{@{}lccccc@{}}
			\toprule
			& Metric & T5-base
			& T5-large
			& T5-3B
			\\ 
			\midrule
	        \multirow{3}*{Spider} 
            & Match & 58.12$_{1.46}$ & 66.63$_{2.31}$ & 71.76 \\
            & Exec & 60.06$_{0.54}$ & 68.28$_{1.61}$ & 74.37 \\
            & Test suite & 56.22$_{0.73}$ & 64.12$_{1.28}$ & 68.38 \\
            \midrule
            \multirow{1}*{GrailQA}
            & Match & 60.00 & 67.00 & 69.00 \\
            \midrule
            \multirow{1}*{WebQSP}
            & F1 & 72.50 & 73.96 & 75.97 \\
            \midrule
            \multirow{2}*{MTOP}
            & Match & 83.89 & 84.70 & 84.88 \\
            & Template & 88.85 & 88.32 & 88.86 \\
			\midrule
            \multirow{1}*{WikiTQ}
            & Acc & 36.94$_{0.19}$ & 43.30$_{0.25}$ & 50.65 \\
            \midrule
            \multirow{1}*{WikiSQL}
            & Acc & 84.50 & 86.27 & 87.34 \\
            \midrule
            \multirow{2}*{CompWebQ}
            & Acc & 66.71 & 68.85 & 70.27 \\
            & F1 & 80.02 & 81.05 & 81.43 \\
            & Hits@1 & 83.64 & 85.49 & 86.20 \\
            \midrule
            \multirow{2}*{HybridQA}
            & Acc & 54.07 & 56.95 & 59.41 \\
            & F1 & 61.85 & 64.62 & 66.76 \\
            \midrule
            \multirow{2}*{MMQA}
            & Acc & 67.29 & 74.08 & 78.48 \\
            & F1 & 75.51 & 81.84 & 82.28 \\
            \midrule
            \multirow{1}*{FeTaQA}
            & BLEU & 29.00 & 30.94 & 31.73 \\
			\midrule 
			\multirow{1}*{DART} 
			& BLEU & 50.62$_{0.72}$ & 51.72$_{0.15}$ & 50.38 \\
            \midrule
			\multirow{1}*{ToTTo}
			& BLEU & 48.29 & 48.95 & 48.95 \\
            \midrule
			\multirow{1}*{MultiWoZ2.1}
			& Joint Acc & 57.52$_{0.96}$ & 58.23$_{0.39}$ & 58.46 \\
            \midrule
			\multirow{1}*{KVRET}
			& BLEU & 20.04 & 18.84 & 17.75 \\
            \midrule
			\multirow{4}*{SParC}
			& Match & 50.54 & 56.69 & 61.51 \\
			& Exec & 53.95 & 60.60 & 67.33 \\
			& Match (interact) & 31.28 & 37.44 & 41.94 \\
			& Exec (interact) & 34.36 & 41.23 & 46.45 \\
			\midrule
			\multirow{4}*{CoSQL}
			& Match & 42.30 & 48.26 & 54.08 \\
			& Exec & 49.26 & 56.01 & 62.23 \\
			& Match (interact) & 12.63 & 16.72 & 22.78 \\
			& Exec (interact) & 16.04 & 20.14 & 26.16 \\
			\midrule
			\multirow{1}*{SQA}
			& Overall Acc  & 49.49 & 59.12 & 60.93 \\
			\midrule
			\multirow{1}*{TabFact}
			& Acc & 76.34$_{0.36}$ & 81.40$_{0.16}$ & 83.97 \\
			\midrule
            \multirow{1}*{FEVEROUS}
            & Acc & 75.05 & 79.81 & 82.40 \\
			\midrule
			\multirow{1}*{SQL2Text}
			& BLEC & 93.69$_{0.29}$ & 93.35$_{0.29}$ & 92.71 \\
            \midrule
			\multirow{1}*{Logic2Text}
			& BLEC & 92.15 & 92.88 & 91.69 \\
			\bottomrule
		\end{tabular}
		\end{adjustbox}
	\caption{Development set performance with full metrics. We do three experiments with different random seeds on representative task of each family and report their averages and standard variances format as $avr_{var}$.}
	\label{tab:baseline-full-dev}
\end{table}

\begin{table}[ht]
	\centering
	\begin{adjustbox}{width=\columnwidth}
		\begin{tabular}{@{}lccccc@{}}
			\toprule
			& Metric & T5-base
			& T5-large
			& T5-3B
			\\ 
			\midrule
            \multirow{1}*{GrailQA}
            & Match & 62.39 & 67.30 & 70.11 \\
            \midrule
            \multirow{1}*{WebQSP}
            & F1 & 78.83 & 79.45 & 80.70 \\
			\midrule
			\multirow{2}*{MTOP}
            & Match & 85.49 & 86.17 & 86.78 \\
            & Template & 87.52 & 89.53 & 90.20 \\
			\midrule
            \multirow{1}*{WikiTQ}
            & Acc & 35.76$_{0.66}$ & 43.22$_{0.65}$ & 49.29  \\
            \midrule
            \multirow{1}*{WikiSQL}
            & Acc & 82.63 & 84.80 & 85.96 \\
            \midrule
            \multirow{2}*{CompWebQ}
            & Acc & 68.43 & 71.38 & 73.26 \\
            & F1 & 80.20 & 81.76 & 82.58 \\
            & Hits@1 & 83.70 & 85.40 & 86.08 \\
            \midrule
            \multirow{7}*{FeTaQA}
            & BLEU & 29.91 & 32.45 & 33.44 \\
            & ROUGE-1-Fmeasure & 61.77 & 64.01 & 65.21 \\
            & ROUGE-2-Fmeasure & 39.44 & 42.26 & 43.09 \\
            & ROUGE-L-Fmeasure & 51.93 & 54.29 & 55.31 \\
            & METEOR & 48.53 & 50.80 & 51.23 \\
            & BertScore-F1 & 0.92 & 0.93 & 0.93 \\
            & BLEURT & -0.01 & 0.06 & 0.09 \\
			\midrule 
			\multirow{5}*{DART} 
            & BLEU & 46.22$_{0.66}$ & 46.89$_{0.53}$ & 46.66 \\
            & TER & 61.80$_{0.20}$ & 60.97$_{0.31}$ & 60.70 \\
            & METEOR & 55.09$_{0.35}$ & 55.76$_{0.25}$ & 55.67 \\
            & BertScore-F1 & 0.95$_{0.00}$ & 0.95$_{0.00}$ & 0.95 \\
            & BLEURT & 0.2833$_{0.0057}$ & 0.30$_{0.00}$ & 0.30 \\
            \midrule
			\multirow{1}*{MultiWoZ2.1}
            & Joint Acc & 54.64$_{0.22}$ & 54.45$_{0.20}$ & 55.42 \\
            \midrule
			\multirow{5}*{KVRET}
            & BLEU & 17.41 & 17.27 & 15.45 \\
            & F1 micro all & 66.45 & 65.85 & 67.88 \\
            & F1 micro schedule & 73.48 & 75.90 & 77.99 \\
            & F1 micro navigate & 64.89 & 62.72 & 65.47 \\
            & F1 micro weather & 63.78 & 62.80 & 64.01 \\
            \midrule
			\multirow{6}*{SQA}
            & Overall Acc  & 52.91 & 61.28 & 62.37 \\
            & Pos 0 Acc  & 62.93 & 67.80 & 59.51 \\
            & Pos 1 Acc  & 44.43 & 55.08 & 60.25 \\
            & Pos 2 Acc  & 50.44 & 61.88 & 68.77 \\
            & Pos 3 Acc  & 53.71 & 58.08 & 65.07 \\
            & Interaction Acc  & 22.24 & 32.59 & 33.17 \\
			\midrule
			\multirow{4}*{TabFact}
            & All Acc & 76.13$_{0.39}$ & 80.85$_{0.24}$ & 83.68 \\
            & Simple Acc & - & 91.38$_{0.32}$ & 93.10 \\
			& Complex Acc & - & 75.76$_{0.19}$ & 79.12 \\
			& Small Acc & - & 82.61$_{0.32}$ & 85.39 \\
            \midrule
			\multirow{1}*{SQL2Text}
            & BLEC & 93.52$_{1.00}$ & 93.68$_{1.12}$  & 94.78  \\
            \midrule
			\multirow{1}*{Logic2Text}
            & BLEC & 90.66 & 90.57 & 91.39 \\
			\bottomrule
		\end{tabular}
	\end{adjustbox}
	\caption{Test set performance with full metrics (for tasks with a publicly available test set). We do three experiments with different random seeds on representative task of each family and report their averages and standard variances format as $avr_{var}$.}
	\label{tab:baseline-full-test}
\end{table}
\begin{table}[ht]
	\centering
	\begin{adjustbox}{width=\columnwidth}
		\begin{tabular}{@{}lccccc@{}}
			\toprule
			& Metric & T5-base
			& T5-large
			& T5-3B
			\\ 
			\midrule
            & BLEU(dev) & 22.80 & 23.07 & 22.71 \\
            & BLEU(test) & 21.21 & 22.36 & 20.40 \\
            & F1 micro all(test) & 67.49 & 68.03 & 70.07 \\
            & F1 micro schedule(test) & 79.39 & 79.47 & 78.54 \\
            & F1 micro navigate(test) & 62.87 & 63.59 & 65.34\\
            & F1 micro weather(test) & 61.43 & 62.61 & 66.74 \\
            & F1 macro all(test) & 65.91 & 64.87 & 66.07 \\
            & F1 macro schedule(test) & 78.73 & 77.23 & 76.02 \\
            & F1 macro navigate(test) & 59.53 & 58.99 & 60.47 \\
            & F1 macro weather(test) & 64.05 & 62.58 & 65.78 \\
			\bottomrule
		\end{tabular}
	\end{adjustbox}
	\caption{Baselines results are higher in pre-processed KVRET dataset. It doesn't change our conclusion on T5 with simple modification when necessary achieves sota on almost all tasks.}
	\label{tab:kvret-preprocess}
\end{table}
For the KVRET dataset, instead of the version used in our main tables, we re-run another more widely used pre-processed version \cite{P18-1136, wu2019global, qin-etal-2020-dynamic} on T5-base, T5-large and T5-3b. Results are shown in Table \ref{tab:kvret-preprocess}. 
\begin{figure*}[t]
\centering
    \subfigure[Spider]{
        \includegraphics[width=0.23\linewidth]{figures/human_and_error/Spider-error.pdf}
    }
    \subfigure[CoSQL]{
        \includegraphics[width=0.23\linewidth]{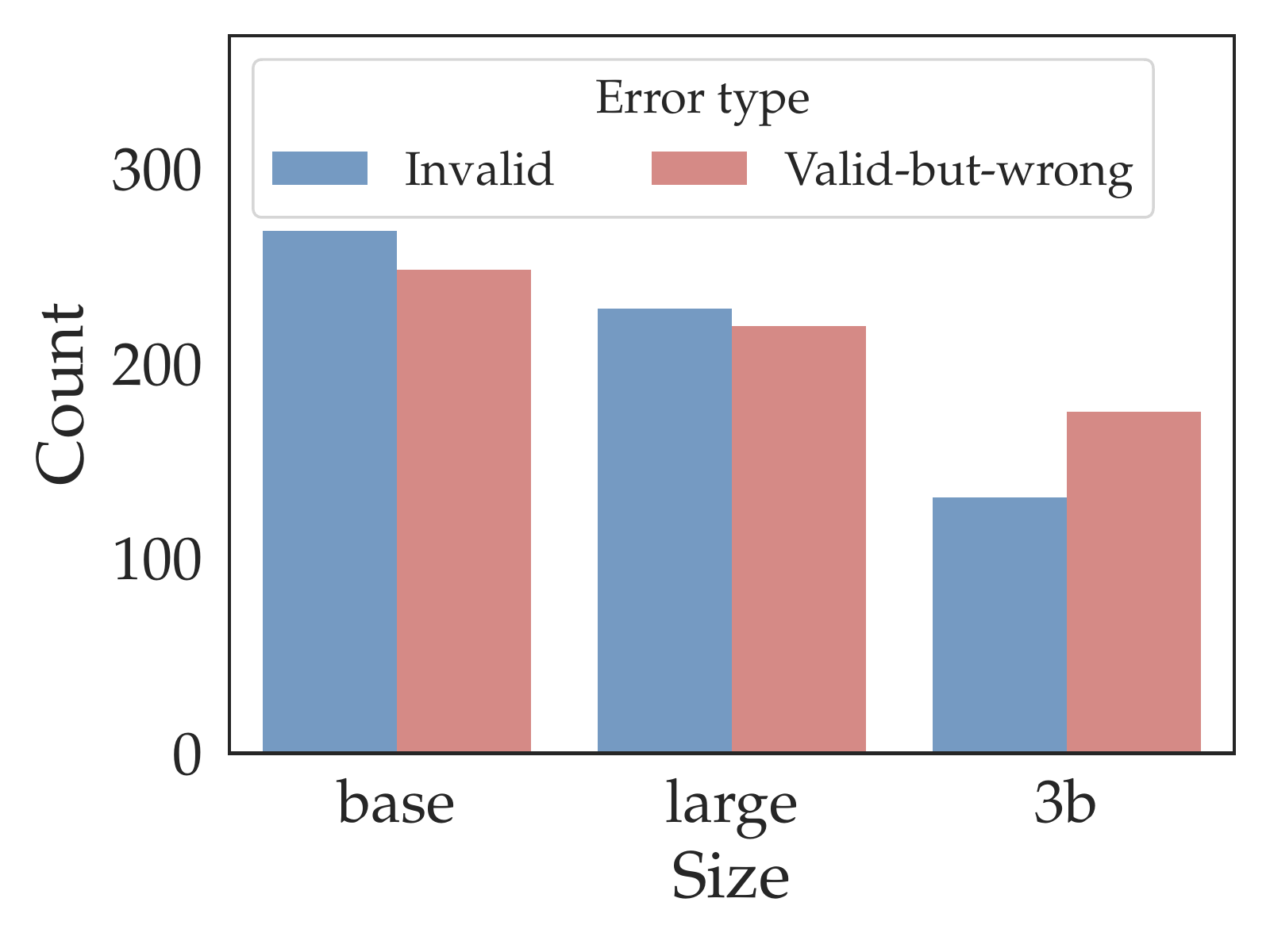}
    }
    \subfigure[SParC]{
        \includegraphics[width=0.23\linewidth]{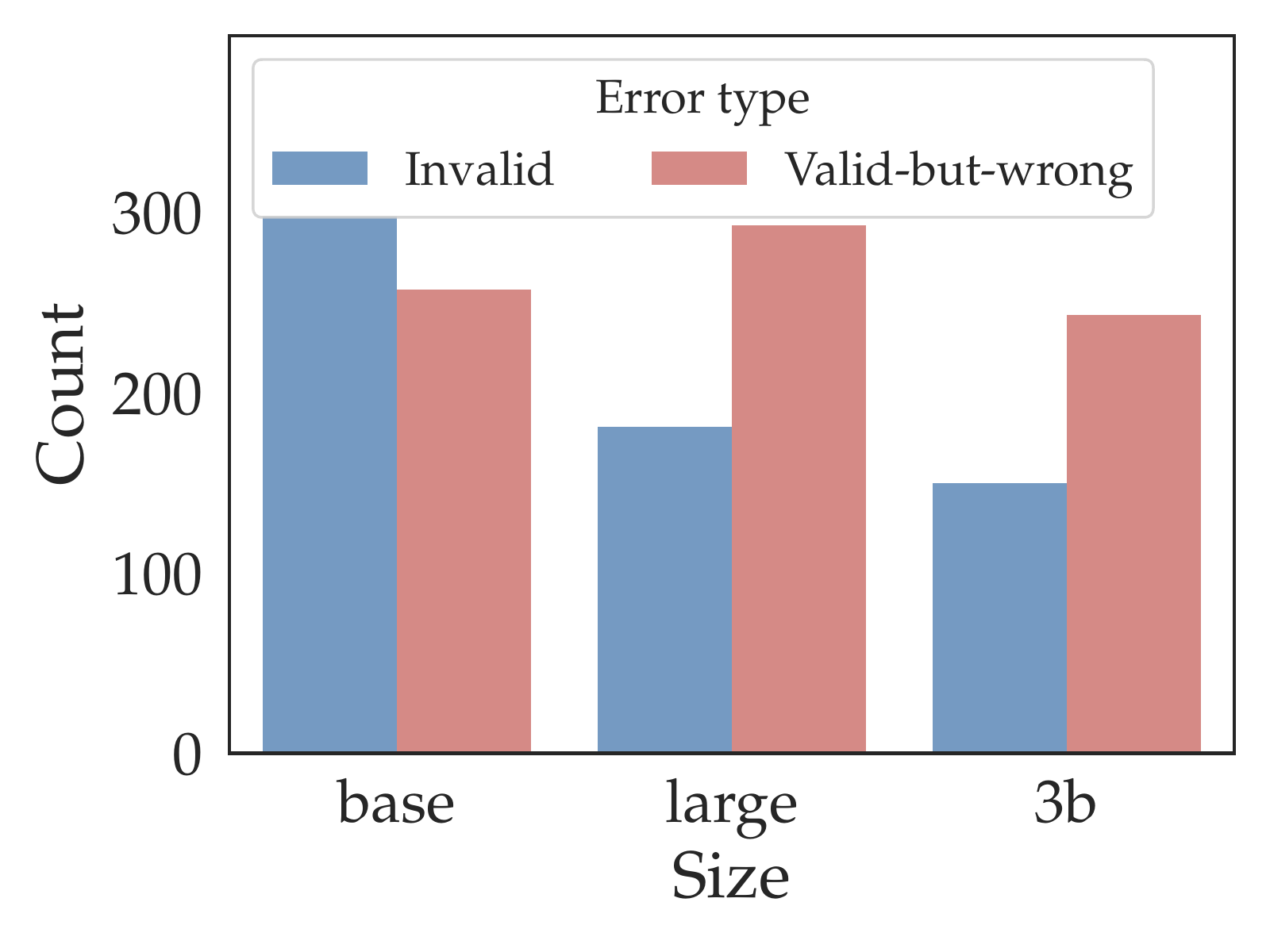}
    }
    \subfigure[GrailQA]{
        \includegraphics[width=0.23\linewidth]{figures/human_and_error/GrailQA-error.pdf}
    }
    \subfigure[WebQSP]{
        \includegraphics[width=0.23\linewidth]{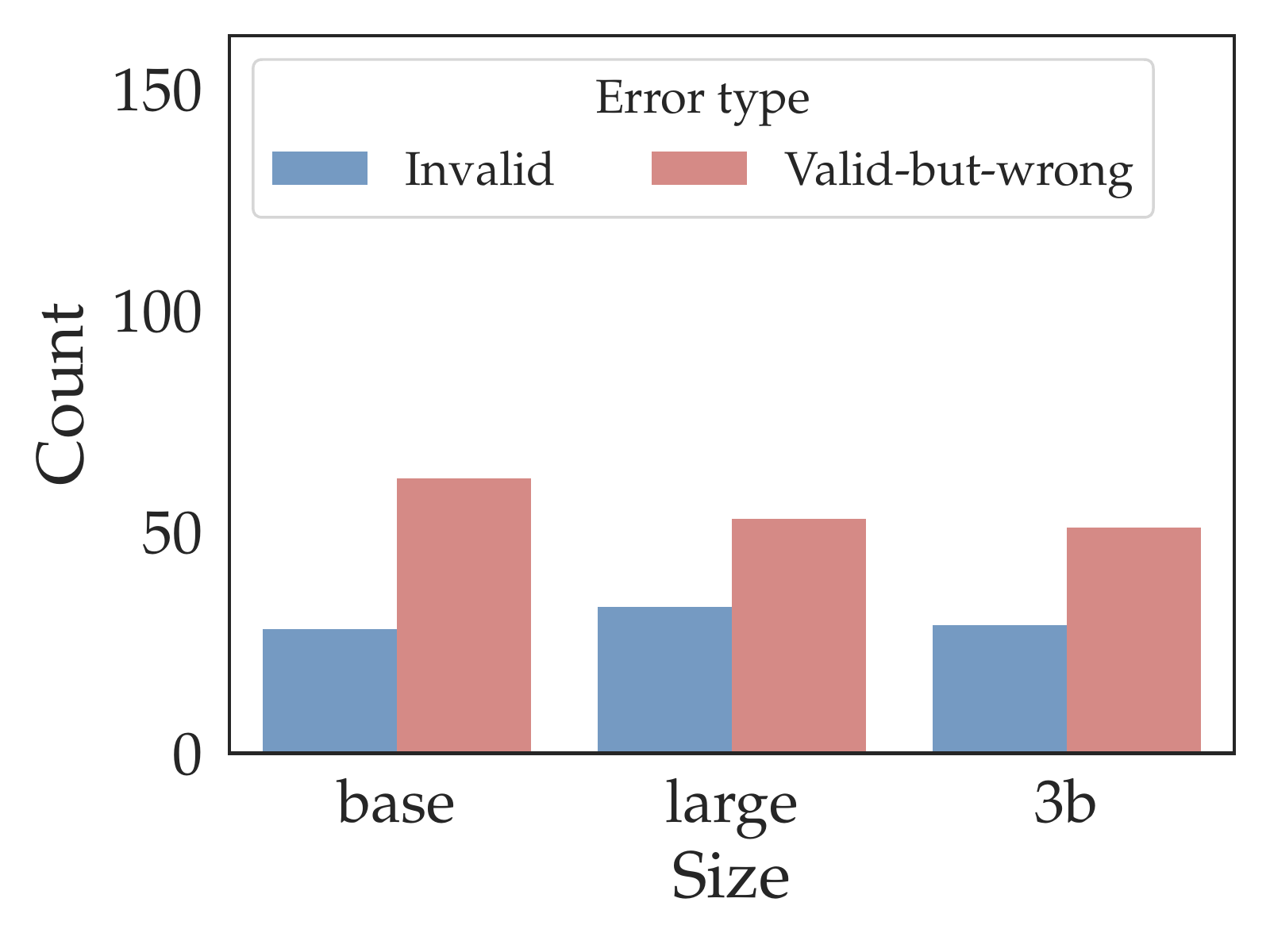}
    }
    \subfigure[MTOP]{
        \includegraphics[width=0.23\linewidth]{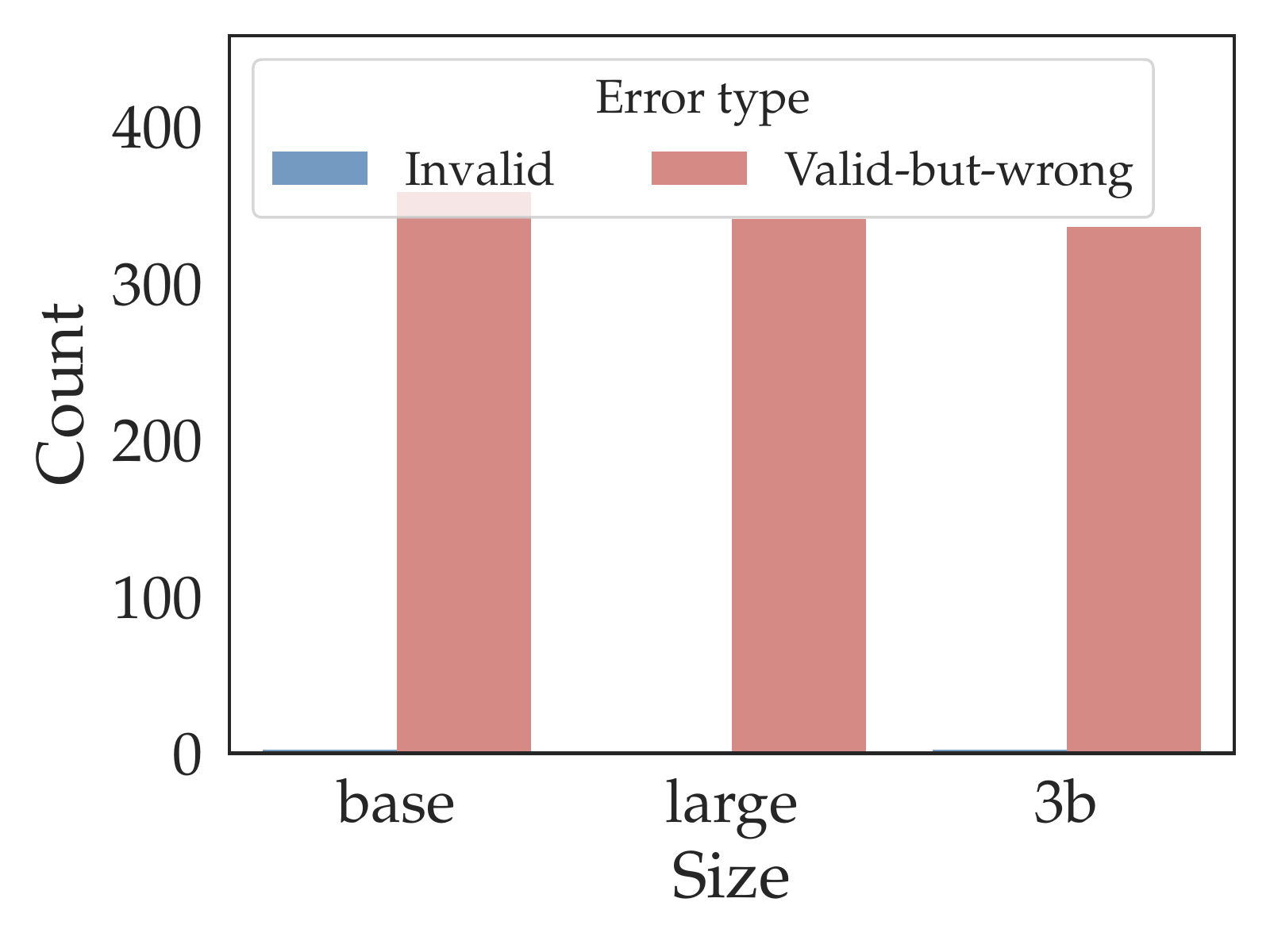}
    }
    \subfigure[FeTaQA]{
        \includegraphics[width=0.23\linewidth]{figures/human_and_error/FeTaQA-error.pdf}
    }
    \subfigure[DART]{
        \includegraphics[width=0.23\linewidth]{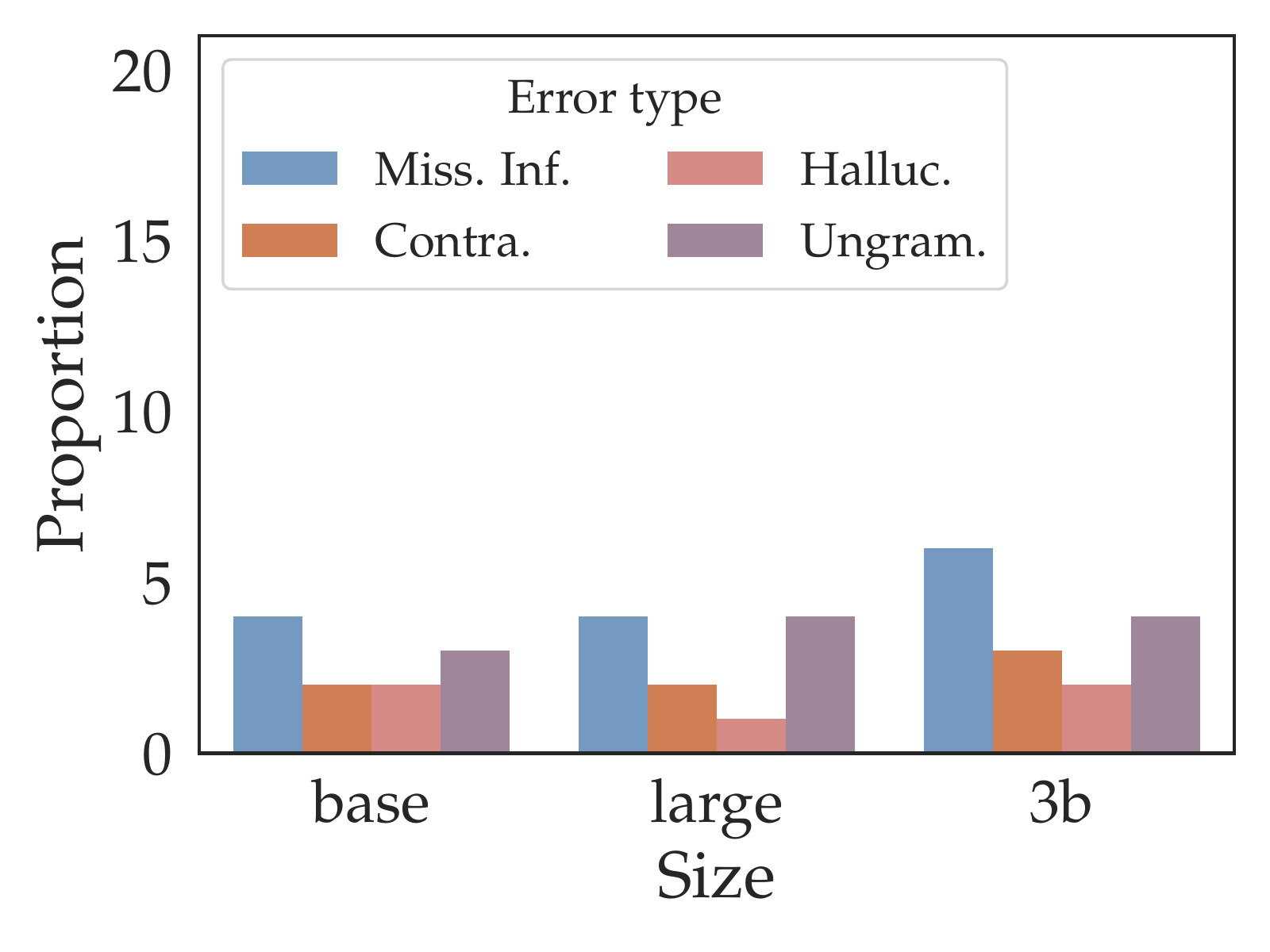}
    }
    \subfigure[ToTTo]{
        \includegraphics[width=0.23\linewidth]{figures/human_and_error/ToTTo-error.pdf}
    }
    \subfigure[KVRET]{
        \includegraphics[width=0.23\linewidth]{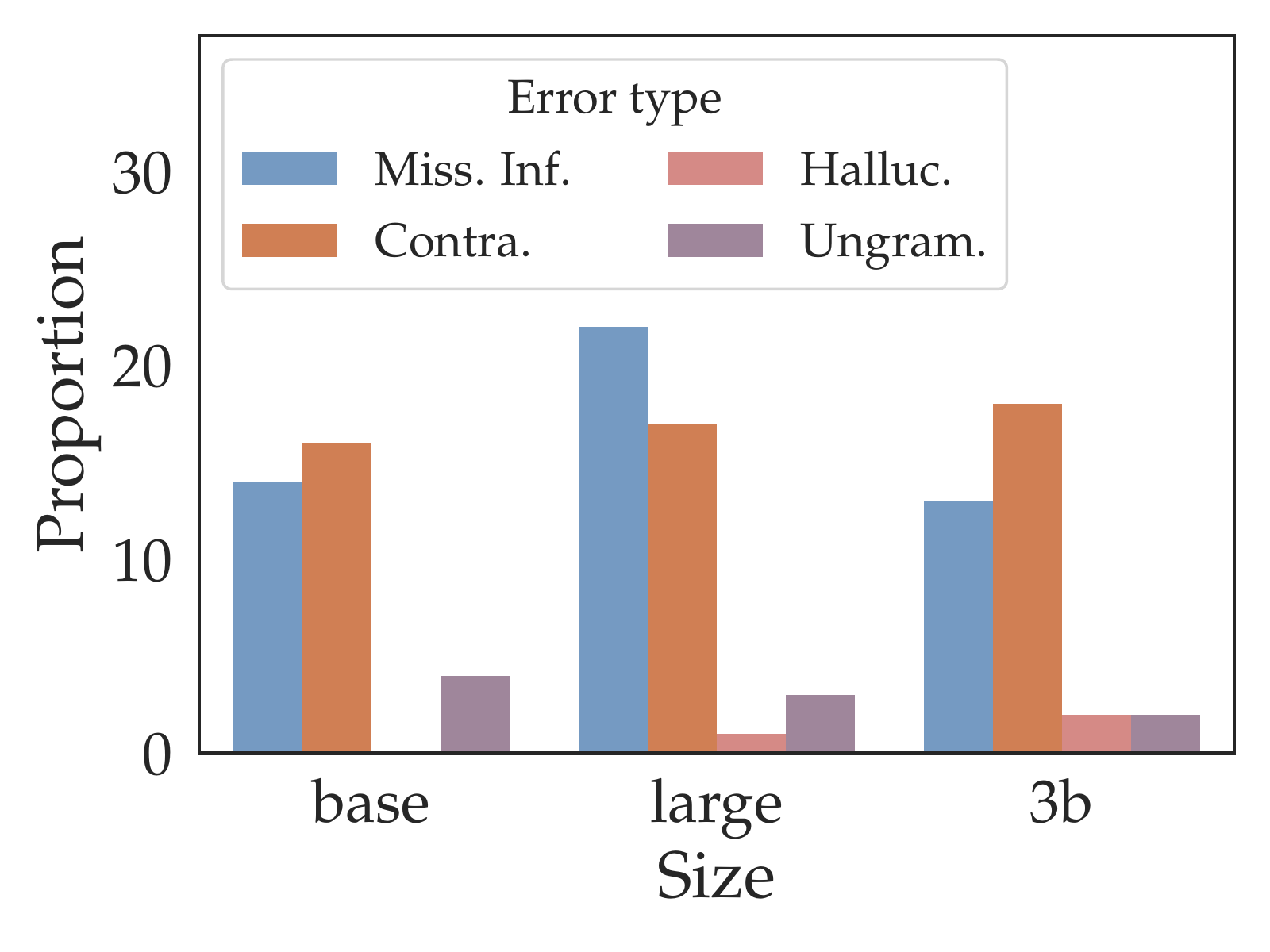}
    }
    \subfigure[SQL2Text]{
        \includegraphics[width=0.23\linewidth]{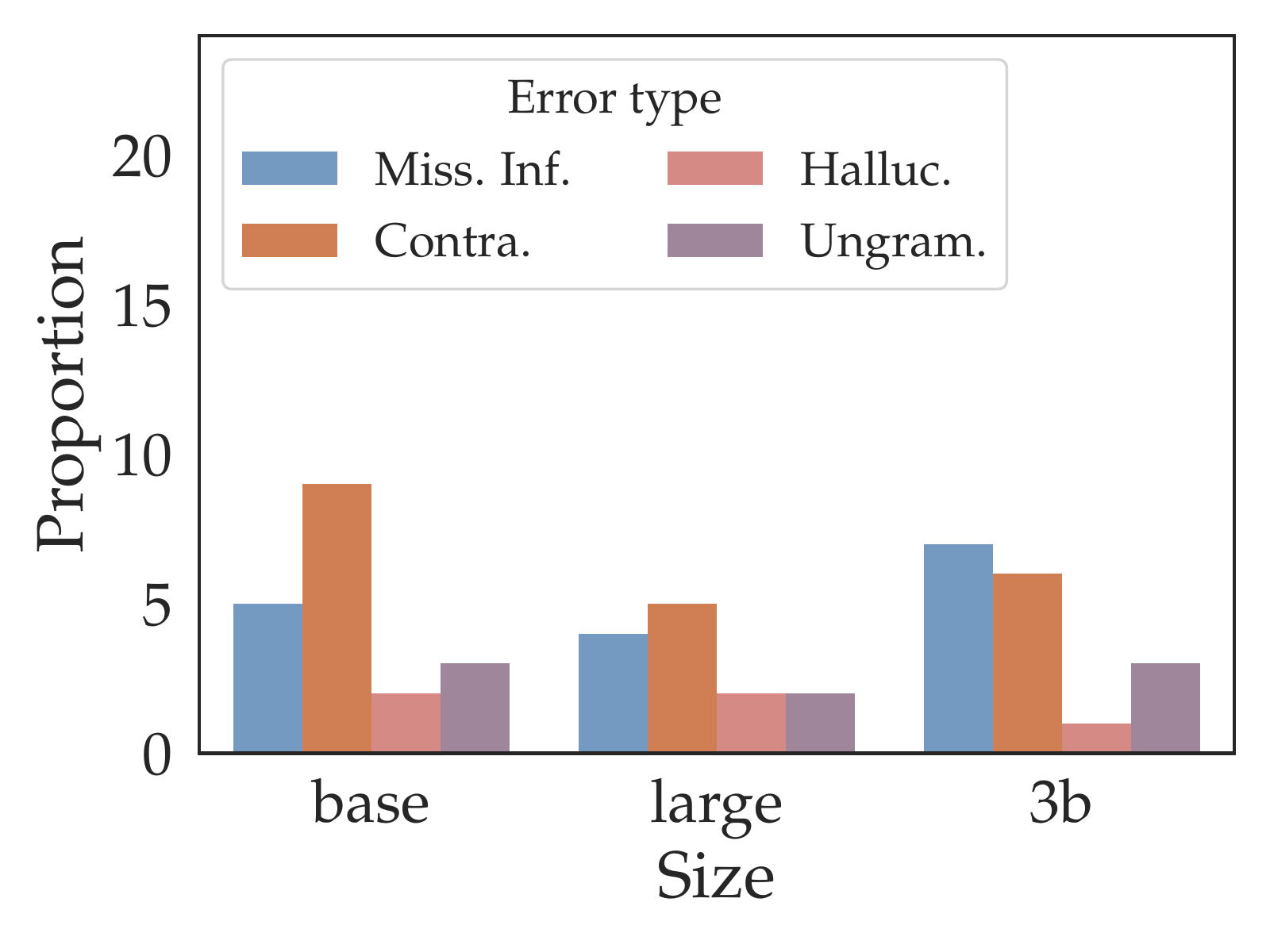}
    }
    \subfigure[Logic2Text]{
        \includegraphics[width=0.23\linewidth]{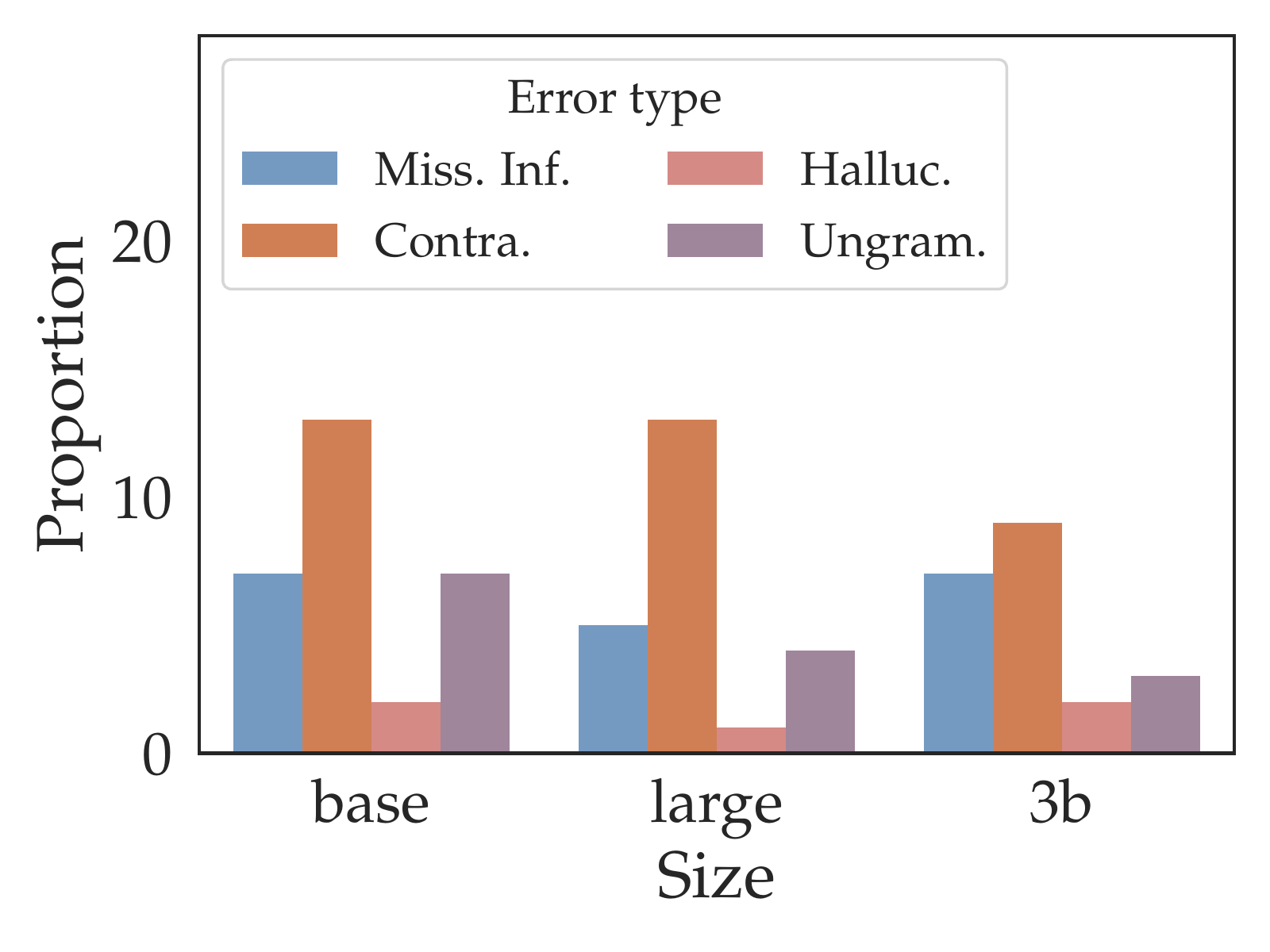}
    }
\caption{\label{fig:error-analysis-full} Error analysis. For semantic parsing, we show the number of invalid/valid-but-wrong predictions. For generation tasks, we show the proportion of missing-information/contradiction/hallucination/ungrammatical predictions among all predictions (one prediction may have multiple errors). }
\end{figure*}

\section{Input and Output Length Analysis}
\label{length_analysis}
\def\infinity{\rotatebox{90}{8}}

\begin{figure}[ht]
    \centering
	\includegraphics[scale=0.4]{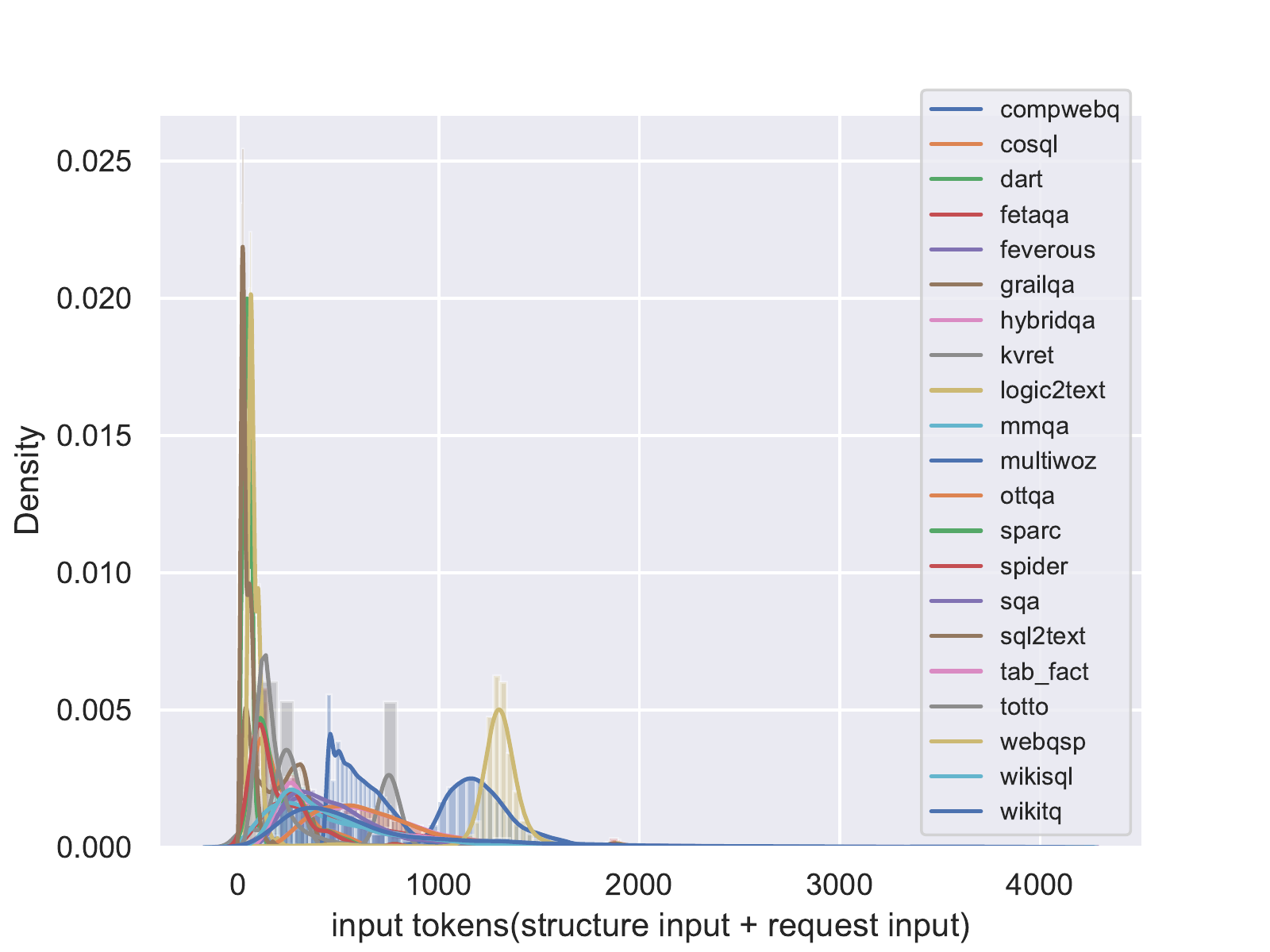}
	\caption{Input token distribution(<4096) in train set from different tasks. We exclude MTOP since it concentrates on a relatively small field which would make this figure unreadable. In general, 1024 is a good length for practice, and for most tasks, 2048 can hold all its inputs. 
	}
	\label{fig:length-distribution}
\end{figure}

\begin{figure}[ht]
    \centering
	\includegraphics[scale=0.4]{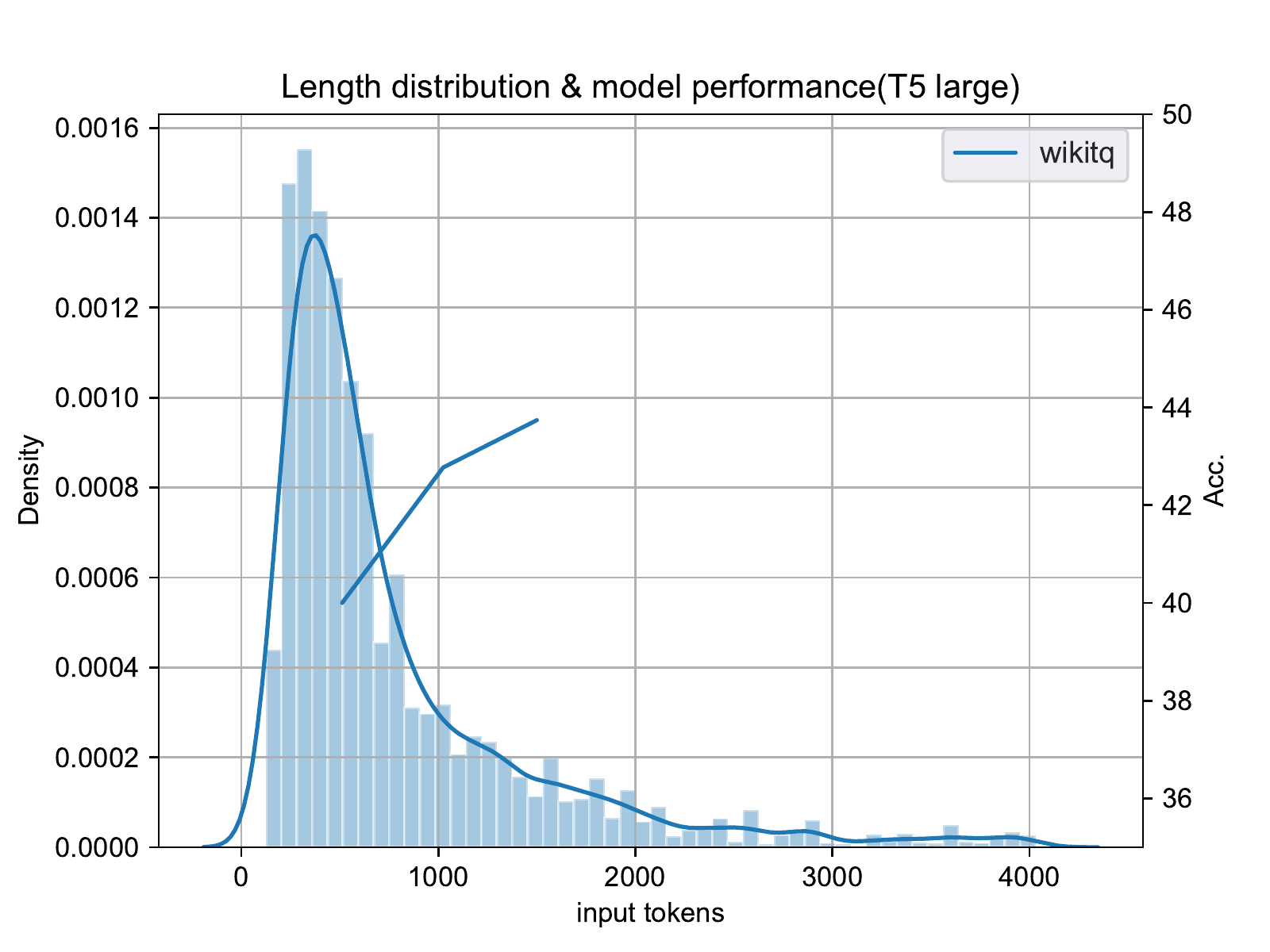}
	\caption{Length effect on WikiTableQuestion.
	}
	\label{fig:length-effect-wikitq}
\end{figure}

Linearization of large structured knowledge input (e.g., large tables and KGs) can be arbitrarily long, which needs to be truncated to fit in GPUs with a limited size. The input and output are tokenized by T5Tokenizer in Huggingface's Transformers.\footnote{\url{https://huggingface.co/t5-base/tree/main}} We visualize the length distribution in Figure~\ref{fig:length-distribution}, and details are presented in Table~\ref{tab:length_distribution}. 
Among the datasets with very long inputs,
we choose WikiTableQuestion to study the impact of input length.
We visualize the table length distribution and performances with different input truncation lengths in Figure~\ref{fig:length-effect-wikitq}.
We observe that the accuracy increases as the input becomes longer, 
motivating future work to study how to effectively encode large structured input, e.g., leveraging sparse attention \cite{Zaheer2020BigBT}. 

\begin{table*}[ht]
	\centering
	\begin{adjustbox}{width=\textwidth}
		\begin{tabular}{@{}lcccccccccccc@{}}
		\toprule
        & \multicolumn{3}{c}{Structure Input Tokens} & \multicolumn{3}{c}{Text Input Tokens} & \multicolumn{3}{c}{Structure Input + Text Input Tokens} & \multicolumn{3}{c}{Sequence Output Tokens} \\
        \cmidrule(lr){2-4}\cmidrule(lr){5-7}\cmidrule(lr){8-10}\cmidrule(lr){11-13}
        Distribution(\%) & [0, 512) & [512, 1024) & [1024, \infinity) & [0, 512) & [512, 1024) & [1024, \infinity) & [0, 512) & [512, 1024) & [1024, \infinity) & [0, 128) & [128, 256) & [256, \infinity)\\
        \midrule
    Spider & 97.01 & 1.81 & 1.17 & 100.00 & 0.00 & 0.00 & 95.47 & 3.35 & 1.17 & 98.81 & 1.18 & 0.0\\
    GRAILQA  & 100.00 & 0.00 & 0.00 & 100.00 & 0.00 & 0.00 & 99.96 & 0.04 & 0.00 & 99.97 & 0.03 & 0.00 \\
    WebQsp & 3.40 & 2.32 & 94.28 & 100.00 & 0.00 & 0.00 & 3.18 & 2.47 & 94.35 & 99.81 & 0.19 & 0.00 \\
    MTOP & 0.00 & 100.00 & 0.00 & 100.00 & 0.00 & 0.00 & 0.00 & 100.00 & 0.00 & 99.97 & 0.03 & 0.00 \\
    WikiTableQuestions  & 48.32 & 27.48 & 24.18 & 100.00 & 0.00 & 0.00 & 46.03 & 29.43 & 24.52 & 99.98 & 0.01 & 0.01\\
    WikiSQL  & 63.38 & 25.33 & 11.29 & 100.00 & 0.00 & 0.00 & 61.50 & 26.79 & 11.70 & 99.97 & 0.02 & 0.01 \\
    ComWebQ  & 1.18 & 14.52 & 84.30 & 100.00 & 0.00 & 0.00 & 1.09 & 11.28 & 87.63 & 99.59 & 0.39 & 0.01 \\
    HybridQA & 35.53 & 50.63 & 13.8 & 100.00 & 0.00 & 0.00 & 31.77 & 53.35 & 14.86 & 100.00 & 0.00 & 0.0\\
    MultiModalQA  & 63.02 & 25.67 & 11.30 & 100.00 & 0.00 & 0.00 & 60.54 & 27.26 & 12.18 & 99.99 & 0.01 & 0.00 \\
    FeTaQA  & 60.36 & 28.62 & 11.01 & 100.00 & 0.00 & 0.00 & 58.46 & 29.85 & 11.68 & 100.00 & 0.00 & 0.0\\
    DART  & 100.00 & 0.00 & 0.00 & 100.00 & 0.00 & 0.00 & 100.00 & 0.00 & 0.00 & 99.99 & 0.01 & 0.0\\
    ToTTo  & 95.80 & 2.87 & 1.31 & 100.00 & 0.00 & 0.00 & 95.80 & 2.87 & 1.31 & 99.99 & 0.01 & 0.0\\
    MultiWoZ  & 100.00 & 0.00 & 0.00 & 98.77 & 1.21 & 0.01 & 54.76 & 45.09 & 0.13 & 0.00 & 100.00 & 0.0\\
    KVRET  & 65.08 & 34.91 & 0.00 & 100.00 & 0.00 & 0.00 & 65.08 & 34.91 & 0.00 & 99.97 & 0.03 & 0.0\\
    SParC & 96.70 & 2.02 & 1.28 & 100.00 & 0.00 & 0.00 & 95.10 & 3.62 & 1.28 & 99.34 & 0.66 & 0.00 \\
    CoSQL  & 96.03 & 2.23 & 1.73 & 100.00 & 0.00 & 0.00 & 93.98 & 4.28 & 1.73 & 99.06 & 0.93 & 0.0\\
    SQA  & 64.54 & 29.74 & 5.71 & 100.00 & 0.00 & 0.00 & 60.96 & 33.11 & 5.92& 95.12 & 4.19 & 0.67\\
    TabFact  & 63.22 & 28.19 & 8.58 & 100.00 & 0.00 & 0.00 & 60.68 & 30.20 & 9.10 & 100.00 & 0.00 & 0.0\\
    FEVEROUS  & 61.37 & 22.24 & 16.39 & 100.00 & 0.00 & 0.00 & 57.53 & 25.07 & 17.40 & 100.00 & 0.00 & 0.00 \\
    SQL2Text & 100.00 & 0.00 & 0.00 & 100.00 & 0.00 & 0.00 & 100.00 & 0.00 & 0.0& 100.00 & 0.00 & 0.0\\
    Logic2Text & 100.00 & 0.00 & 0.00 & 100.00 & 0.00 & 0.00 & 100.00 & 0.00 & 0.0& 100.00 & 0.00 & 0.0\\
    \bottomrule
    \end{tabular}
    \end{adjustbox}
	\caption{Input and output length for each task's train set.}
	\label{tab:length_distribution}
\end{table*}

\begin{table*}[ht]
	\centering
	\begin{adjustbox}{width=\textwidth}
		\begin{tabular}{@{}lcccccccccccc@{}}
		\toprule
        & \multicolumn{3}{c}{Structure Input Tokens} & \multicolumn{3}{c}{Text Input Tokens} & \multicolumn{3}{c}{Structure Input + Text Input Tokens} & \multicolumn{3}{c}{Sequence Output Tokens} \\
        \cmidrule(lr){2-4}\cmidrule(lr){5-7}\cmidrule(lr){8-10}\cmidrule(lr){11-13}
        Distribution(\%) & [0, 512) & [512, 1024) & [1024, \infinity) & [0, 512) & [512, 1024) & [1024, \infinity) & [0, 512) & [512, 1024) & [1024, \infinity) & [0, 128) & [128, 256) & [256, \infinity)\\
        \midrule
    Spider & 100.00 & 0.00 & 0.00 & 100.00 & 0.00 & 0.00 & 100.00 & 0.00 & 0.00 & 99.23 & 0.77 & 0.00 \\
    GRAILQA & 100.00 & 0.00 & 0.00 & 100.00 & 0.00 & 0.00 & 100.00 & 0.00 & 0.00 & 100.00 & 0.00 & 0.00   \\
    WebQsp  & 3.56 & 1.29 & 95.15 & 100.00 & 0.00 & 0.00 & 3.56 & 1.29 & 95.15 & 99.68 & 0.32 & 0.00 \\
    Russ & 100.00 & 0.00 & 0.00 & 100.00 & 0.00 & 0.00 & 100.00 & 0.00 & 0.00 & 100.00 & 0.00 & 0.00\\
    MTOP & 0.00 & 100.00 & 0.00 & 100.00 & 0.00 & 0.00 & 0.00 & 100.00 & 0.00 & 100.00 & 0.00 & 0.00   \\
    WikiTableQuestions  & 49.56 & 28.65 & 21.79 & 100.00 & 0.00 & 0.00 & 48.60 & 29.11 & 22.29 & 99.93 & 0.07 & 0.00 \\
    WikiSQL  & 63.90 & 25.88 & 10.22 & 100.00 & 0.00 & 0.00 & 62.06 & 26.99 & 10.95 & 100.00 & 0.00 & 0.00 \\
    ComWebQ  & 0.28 & 15.79 & 83.93 & 100.00 & 0.00 & 0.00 & 0.28 & 12.66 & 87.06 & 99.00 & 1.00 & 0.00 \\
    HybridQA & 38.37 & 52.63 & 9.00 & 100.00 & 0.00 & 0.00 & 34.16 & 56.00 & 9.84 & 100.00 & 0.00 & 0.00\\
    MultiModalQA & 66.22 & 25.72 & 8.06 & 100.00 & 0.00 & 0.00 & 64.02 & 27.38 & 8.59 & 100.00 & 0.00 & 0.00  \\
    FeTaQA  & 67.03 & 27.47 & 5.49 & 100.00 & 0.00 & 0.00 & 64.84 & 29.57 & 5.59 & 100.00 & 0.00 & 0.00\\
    DART  & 100.00 & 0.00 & 0.00 & 100.00 & 0.00 & 0.00 & 100.00 & 0.00 & 0.00 & 100.00 & 0.00 & 0.00\\
    ToTTo  & 95.82 & 2.92 & 1.26 & 100.00 & 0.00 & 0.00 & 95.82 & 2.92 & 1.26 & 100.00 & 0.00 & 0.00 \\
    MultiWoZ  & 100.00 & 0.00 & 0.00 & 99.16 & 0.84 & 0.00 & 25.07 & 74.68 & 0.24 & 0.00 & 100.00 & 0.00 \\
    KVRET  & 65.76 & 34.24 & 0.00 & 100.00 & 0.00 & 0.00 & 65.76 & 34.24 & 0.00 & 99.79 & 0.21 & 0.00 \\
    SParC  & 100.00 & 0.00 & 0.00 & 100.00 & 0.00 & 0.00 & 100.00 & 0.00 & 0.00 & 99.26 & 0.74 & 0.00 \\
    CoSQL  & 100.00 & 0.00 & 0.00 & 100.00 & 0.00 & 0.00 & 99.62 & 0.38 & 0.00 & 99.23 & 0.77 & 0.00 \\
    SQA  & 60.09 & 33.38 & 6.53 & 100.00 & 0.00 & 0.00 & 56.91 & 36.42 & 6.67 & 94.17 & 5.39 & 0.44 \\
    TabFact  & 62.17 & 29.31 & 8.52 & 100.00 & 0.00 & 0.00 & 59.95 & 30.91 & 9.14 & 100.00 & 0.00 & 0.00\\
    FEVEROUS & 61.56 & 23.71 & 14.73 & 100.00 & 0.00 & 0.00 & 57.57 & 26.58 & 15.85 & 100.00 & 0.00 & 0.00 \\
    SQL2Text & 100.00 & 0.00 & 0.00 & 100.00 & 0.00 & 0.00 & 100.00 & 0.00 & 0.00 & 100.00 & 0.00 & 0.00\\
    Logic2Text & 100.00 & 0.00 & 0.00 & 100.00 & 0.00 & 0.00 & 100.00 & 0.00 & 0.00 & 100.00 & 0.00 & 0.00 \\
    \bottomrule
    \end{tabular}
    \end{adjustbox}
	\caption{Input and output length for each task's development set.}
	\label{tab:length_distribution_dev}
\end{table*}

\begin{table*}[ht]
	\centering
	\begin{adjustbox}{width=\textwidth}
		\begin{tabular}{@{}lcccccccccccc@{}}
		\toprule
        & \multicolumn{3}{c}{Structure Input Tokens} & \multicolumn{3}{c}{Text Input Tokens} & \multicolumn{3}{c}{Structure Input + Text Input Tokens} & \multicolumn{3}{c}{Sequence Output Tokens} \\
        \cmidrule(lr){2-4}\cmidrule(lr){5-7}\cmidrule(lr){8-10}\cmidrule(lr){11-13}
        Distribution(\%) & [0, 512) & [512, 1024) & [1024, \infinity) & [0, 512) & [512, 1024) & [1024, \infinity) & [0, 512) & [512, 1024) & [1024, \infinity) & [0, 128) & [128, 256) & [256, \infinity)\\
        \midrule
    Spider & - & - & - & - & - & - & - & - & - & - & - & - \\
    GRAILQA & 100.00 & 0.00 & 0.00 & 100.00 & 0.00 & 0.00 & 100.00 & 0.00 & 0.00 & 99.98 & 0.02 & 0.00  \\
    WebQsp & 3.48 & 1.95 & 94.57 & 100.00 & 0.00 & 0.00 & 3.36 & 2.07 & 94.57 & 100.00 & 0.00 & 0.00 \\
    Russ & 100.00 & 0.00 & 0.00 & 100.00 & 0.00 & 0.00 & 100.00 & 0.00 & 0.00 & 100.00 & 0.00 & 0.00  \\
    MTOP & 0.00 & 100.00 & 0.00 & 100.00 & 0.00 & 0.00 & 0.00 & 100.00 & 0.00 & 100.00 & 0.00 & 0.00  \\
    WikiTableQuestions  & 48.00 & 31.15 & 20.86 & 100.00 & 0.00 & 0.00 & 47.08 & 31.70 & 21.22 & 99.98 & 0.02 & 0.00\\
    WikiSQL & 61.49 & 26.00 & 12.51 & 100.00 & 0.00 & 0.00 & 59.57 & 27.43 & 13.00 & 99.96 & 0.03 & 0.01 \\
    ComWebQ  & 0.85 & 16.02 & 83.13 & 100.00 & 0.00 & 0.00 & 0.85 & 13.07 & 86.08 & 99.43 & 0.57 & 0.00  \\
    HybridQA & - & - & - & - & - & - & - & - & - & - & - & - \\
    FeTaQA  & 65.40 & 28.01 & 6.59 & 100.00 & 0.00 & 0.00 & 63.26 & 29.51 & 7.24 & 100.00 & 0.00 & 0.00 \\
    DART  & 100.00 & 0.00 & 0.00 & 100.00 & 0.00 & 0.00 & 100.00 & 0.00 & 0.00 & 100.00 & 0.00 & 0.00 \\
    ToTTo  & - & - & - & - & - & - & - & - & - & - & - & - \\
    MultiWoZ & 100.00 & 0.00 & 0.00 & 98.71 & 1.29 & 0.00 & 24.82 & 74.93 & 0.24 & 0.00 & 100.00 & 0.00 \\
    KVRET  & 66.14 & 33.86 & 0.00 & 100.00 & 0.00 & 0.00 & 66.14 & 33.86 & 0.00 & 100.00 & 0.00 & 0.00 \\
    SParC & - & - & - & - & - & - & - & - & - & - & - & -  \\
    CoSQL  & - & - & - & - & - & - & - & - & - & - & - & - \\
    SQA & 62.54 & 30.92 & 6.54 & 100.00 & 0.00 & 0.00 & 61.37 & 32.05 & 6.58 & 93.69 & 5.68 & 0.63 \\
    TabFact  & 64.59 & 28.01 & 7.40 & 100.00 & 0.00 & 0.00 & 62.55 & 29.35 & 8.10 & 100.00 & 0.00 & 0.00\\
    FEVEROUS  & - & - & - & - & - & - & - & - & - & - & - & - \\
    SQL2Text& 100.00 & 0.00 & 0.00 & 100.00 & 0.00 & 0.00 & 100.00 & 0.00 & 0.00 & 100.00 & 0.00 & 0.00 \\
    Logic2Text & 100.00 & 0.00 & 0.00 & 100.00 & 0.00 & 0.00 & 100.00 & 0.00 & 0.00 & 100.00 & 0.00 & 0.00 \\
    \bottomrule
    \end{tabular}
    \end{adjustbox}
	\caption{Input and output length for each task's test set.}
	\label{tab:length_distribution_test}
\end{table*}

\section{Experimental Setup}
\label{app:experimental-setup}

\subsection{Implementation Details}
\label{subapp:implementation-details}
We use T5 \cite{2020t5} as our backbone language model. Each experiment For T5-3B experiments, we use Deepspeed\footnote{\url{https://github.com/microsoft/DeepSpeed}} to save memory. We use batch size 32 as default, except WikiTQ, WikiSQL, and TabFact, for which for use batch size 128 because we found it to work significantly better. We use the Adafactor optimizer for T5-base and T5-large, and AdamW for T5-3b. We evaluate on the development set for each 500 steps and use the average development set metric for best checkpoint selection. For all tasks, we set learning rate to 5e-5 and used linear learning rate decay. All experiments are done on NVIDIA Tesla V100 and NVIDIA Tesla A100.

\subsection{Metric Details}
For most semantic parsing tasks, we report the exact match accuracy of logical forms, and for task has test suite \cite{ruiqi20}, we add test suite metric to represent model's performance; an exception is WebQSP, for which we follow previous work to execute the parses and report the F1 score. 
For QA, we report the exact match accuracy of answer sets. For data-to-text generation, we report sacre-BLEU \cite{post-2018-call}.\footnote{Signature: BLEU + case.lc + numrefs.1 + smooth.exp + tok.13a + version.1.4.0
} 
We use each task's representative metric used by previous works. 
For fact verification, we report the accuracy. For high-fidelity NLG, we report BLEC \cite{shu-etal-2021-logic}, which is the exact match between keywords in the formal language and the natural language. Unless specified, we use T5-large and report the development set performance. 

\subsection{T0 Zero-shot Experimental Details}
\label{subapp:t0_zero}
For each task in \uskg we search \citet{sanh2021multitask} for the most similar instructions(if there is no one for use, we create one follow their writing style), make our input in that format and directly test on T0 3B. The specific instructions are shown below.
\begin{lstlisting}
Spider
Given database schema "[linearized database schema]". Can you tell me the SQL for "[request]"?

WikiTQ
I know that the answer to "[request]" is in "[linearized table]". Can you tell me what it is?

DART
Put the triples together to form a sentence: [relation triples]

MultiWoZ
Known ontology "[ontology]", the dialogue state when "[dialogue history and current request]" is given

TabFact
Suppose "[linearized table]" Can we infer that "[statement]"?

SQL2Text
Paraphrase "[SQL]" to natural language:

\end{lstlisting}

\subsection{GPT3 and Codex Details}
\label{subapp:gpt3_codex}
\subsubsection{Hyperparameter Settings}
\paragraph{Temperature} For GPT3 and Codex, we set the decoding temperature to 0 (i.e., greedy decoding without sampling) for Spider, WikiTQ, MultiWoZ and TabFact. We observe a drop of 10\% in the exact match metric when set the temperature to 1 by default in OpenAI. For Codex, we tune the temperature from 0 to 1 in a step of 0.1 for DART, SQL2Text, and no significant difference is observed. For GPT3, we do not tune on that to stay within our budget.
\paragraph{Max output length}
We set max output length to 256 for Spider, WikiTQ, MultiWoZ and SQL2Text, while 4 for TabFact to contain more length in the input side(the concept of max length in GPT3 and Codex is the sum of input tokens length and output tokens length). We set ``\textbackslash n'' as the stop token.
\subsubsection{Prompts}
We use simple prompt words for each task to concatenate the request, linearized structured knowledge, and context together.
For example, for each example in WikiTQ, we format it as ``\textit{examples}\textbackslash n\textbackslash n[linearized table] || Write a answer for [request] \textbackslash nThe answer is:'', and make GPT3 and Codex make the completion as prediction. 
We do experiments on Spider with different format of forming structured knowledge (e.g., linearization, description), but get a similar result.
Better usage of GPT3 and Codex under the \uskg framework is an interesting direction.

\subsection{Human Evaluation}
\label{subapp:human-evaluation}
Participants of our human evaluation are eight of the authors of this paper. They are familiar with the tasks being evaluated. The human evaluation guideline is shown below. 
\begin{lstlisting}
## General Guideline
1. Each line is a dev set sample, with some inputs (detailed below), a human reference (seq_out) shown in blue, and three model outputs named model1, model2, and model3.
2. Each model output receives a 0-1 score (0 stands for incorrect, and 1 stands for correct). By "correct" we mean "responding to the user request properly and correctly, without grammar or wording mistakes". 
3. When an output is incorrect, you specify the type(s) of error, e.g., 1) missing information, 2) contradiction, 3) hallucination, and 4) ungrammatical. 

## Task-Specific Details
### DART
1. Task: triples-to-text generation.
2. struct_in: a set of relation-triples joined by ``|``. Each relation-triple is of form ``entityA : relation : entityB``.

### FeTaQA
1. Task: free-form QA
2. question: a question about the table.
3. table: a table represented as a dictionary: {"header": [header item, ...], "rows": [[cell value, ...], ...]}.
4. meta: table_page_title | table_section_title

### KVRET
1. Task: dialogue system
2. dialogue: a dialogue represented as a dictionary: {"driver": [request1, ...], "assistant": [response1, ...]}, the last response of the assistant is the human reference. 
3. kb: a knowledge base represented as a dictionary: {"header": [header item, ...], "rows": [[cell value, ...], ...]}.

### Logic2Text
1. Task: logic expression to text translation
2. table: a table represented as a dictionary: {"caption": table caption, "header": [header item, ...], "rows": [[cell value, ...], ...]}.
3. logic_str: logic expression of a statement.

### SQL2Text
1. Task: SQL to text translation
2. query: SQL.

### ToTTo
1. Task: highlighted-table-to-text generation.
2. table_page_title and section: table meta information.
3. Visualization of highlighted tables is provided in ``totto_vis/``.
\end{lstlisting}

\subsection{Hyperparameters}
\label{subapp:hyperparameters}

\begin{table*}[ht]
    \small
	\centering
		\begin{tabular}{@{}ccccc@{}}
			\toprule
			Task type & Task & Input length & Batch size & Beam size \\
			\midrule
	        \multirow{4}*{\textit{Semantic Parsing}} & Spider \cite{Yu18c} & 512 & 32 & 1 \\
            & GrailQA \cite{gu2021beyond} & 512 & 32 & 4 \\
            & WebQSP \cite{yih-etal-2016-value} & 1024 & 32 & 4 \\
            & MTOP \cite{li-etal-2021-mtop} & 1024 & 32 & 4 \\
			\midrule
			\multirow{6}*{\textit{Question Answering}}
			& WikiSQL \cite{zhongSeq2SQL2017} & 1024 & 128 & 4 \\
            & WikiTQ \cite{pasupat-liang-2015-compositional} & 1024 & 128 & 4 \\
            & CompWebQ \cite{talmor18compwebq} & 1024 & 32 & 4 \\
            & HybridQA \cite{chen2020hybridqa} & 1024 & 32 & 4 \\
            & MultiModalQA \cite{talmor2021multimodalqa} & 1024 & 32 & 4 \\
            & FeTaQA \cite{nan2021feta} & 512 & 32 & 4  \\
			\midrule
			\multirow{2}*{\textit{Data-to-Text}} & DART \cite{nan2021dart} & 512 & 32 & 4 \\
			& ToTTo \cite{parikh2020totto} & 512 & 32 & 4 \\
			\midrule
			& MultiWoZ2.1 \cite{eric2019multiwoz} & 1024 & 32 & 4 \\
			& KVRET \cite{Eric2017KeyValueRN} & 1024 & 32 & 4 \\
			\textit{Conversational} 
			& SParC \cite{Yu19} & 512 & 32 & 1 \\
			& CoSQL \cite{yu-etal-2019-cosql} & 512 & 32 & 1 \\
			& SQA \cite{iyyer-etal-2017-search} & 1024 & 128 & 4 \\
			\midrule
			\multirow{2}*{\textit{Fact Verification}} & TabFact \cite{2019TabFactA} & 1024 & 128 & 4 \\
			& FEVEROUS
			\cite{aly2021fact} & 1024 & 32 & 4 \\
			\midrule
			\multirow{2}*{\textit{High-fidelity NLG}} & SQL2Text \cite{shu-etal-2021-logic} & 512 & 32 & 4 \\
			& Logic2Text \cite{chen-etal-2020-logic2text} & 512 & 32 & 4 \\
			\bottomrule
		\end{tabular}
	\caption{Hyperparameters for each SKG task. 
	}
	\label{tab:hyperparameters}
\end{table*}
Shown in Table \ref{tab:hyperparameters}. For semantic parsing tasks, the decoding was done under the greedy search, where we set the beam size to 1 specially. For tasks with a long linearized sequence, we used 1024 as input length to hold the maximum of input; reasons are explained in App.~\ref{length_analysis}.

\section{Training Details}
\label{app:step}
Here we show comparisons of finetuning and prefix-tuning on aspect of training.
For prefix-tuning, we use random initialization as done by \citet{li2021prefixtuning}.
In general, prefix-tuning needs more steps than finetuning but has the ability to reach comparable results with continued training.
\begin{table}[t]
	\centering
	\small
	\begin{adjustbox}{width=0.7\linewidth}
		\begin{tabular}{@{}lcc@{}}
		    \toprule
			Task & Finetune & Prefix-tuning  \\
			\midrule
			Spider & 16500 & 100000 \\
            GrailQA  & 17000 & 78000 \\
            WebQSP  & 1500 & 8000 \\
            MTOP  & 30000 & 60000\\
			WikiSQL  & 8500 & 80000 \\
            WikiTQ  & 1500 & 16000\\
            CompWebQ  & 3500 & 27000 \\
            HybridQA  & 7000 & 30000 \\
            MultiModalQA & 6000 & 40000 \\
            FeTaQA & 11000 & 20000 \\
			DART  & 7000 & 250000 \\
			ToTTo & 12000 & >250000 \\
			MultiWoZ2.1 & 6000 & 40000 \\
			KVRET  & 4000 & 40000 \\
			SParC  & 2000 & 6400 \\
			CoSQL & 38000 & 100000 \\
			SQA & 27000 & >250000\\
			TabFact & 8000 & 210000 \\
			FEVEROUS  & 1200 & 40000 \\
			SQL2Text & 3000 & 10000 \\
			Logic2Text & 3500 & 10000\\
			\bottomrule
		\end{tabular}
	\end{adjustbox}
	\caption{The comparison of approximate training steps finetuning and prefix-tuning used to reach the decent performance on T5 base. >250000 means we stop the training due to time limitation. Prefix-tuning needs more steps to converge and converges to comparable performances.}
	\label{tab:steps_comparison}
\end{table}

\section{Task Unification}
\label{app:task-unification}
\subsection{Term Definition}
\label{subapp:term-definition}
\paragraph{Highlighted tables}
A highlighted table contains a table, table metadata (such as the title),
and a set of highlighted cells which entails the text description \cite{parikh2020totto}. 
\paragraph{Relation-triples}
Relation triples are a set of subject-predicate-object triples to capture rich relationships in the data. Many data-to-text tasks such as DART \cite{nan2021dart} take these relation triples as inputs and generate natural language from them.
\paragraph{Knowledge Graph}
A knowledge graph is a multi-relational graph composed of entities (nodes) and relations (different types of edges). Each edge is represented as a triple of the form (head entity, relation, tail entity), also called a fact, indicating that two entities are connected by a specific relation \cite{wang2017survey}.
\paragraph{Dialogue State and Ontology}
A dialogue state $s_t$ at any turn ${t}$ in a dialogue comprises the summary of the dialogue history until turn $t$, such that $s_t$ contains all sufficient information for the system to choose the next action. \cite{williams2016dialog}
Specifically, it captures the user goals in the conversation in the form of (slot, value) pairs. The set of possible slots is predefined in the ontology ${O}$, typically domain-dependent, while the values assumed by each slots are provided by the user as a dialogue goal. 

\subsection{Linearization}
\begin{itemize}[leftmargin=*]
    \item \textbf{Tables.} Following \citet{liu2021tapex},  we linearize the table into a sequence. By inserting several special tokens to indicate the table boundaries, a  linearized table can be represented as ``col: $c_1$, ..., $c_N$ row 1 : $r_1$ row 2 : $r_2$... $r_M$ '', $N$ and $M$ are the number of columns and rows.
    \item \textbf{Highlighted tables.} Following \citet{parikh2020totto}, we represent each highlighted cell by concatenating its value, column headers, and row headers. The table is represented as the concatenation of the page title, section title, and representations of all highlighted cells. 
    \item \textbf{Relation-triples and knowledge graphs.} Following \citet{nan2021dart}, each relation-triple is linearized as ``\textit{sub} : \textit{rela} : \textit{obj}'', and different triples are joined by `` | ''. 
    The subgraph retrieved from the knowledge graph is treated as a list of relation-triples and we use the same formulation. 
    \item \textbf{Ontology.} Following \citet{hosseini2020simple} and \citet{lin2021leveraging}, for each slot in ontology, each slot along with its all possible values is formatted as ``\textit{slot} : \textit{value}$_1$, ... \textit{value}$_{\textit{slot}_n}$ '', different slot-values are joined by `` | '' 
\end{itemize}

\subsection{Output Format}
When the output is \textit{natural language} or \textit{formal language} we do not modify it because it is already in sequence format; 
a \textit{set of answers}, we use a comma followed by a space to join the answers; a \textit{Boolean value}, we map True to ``entailed'' and False to ``refuted''; 
a \textit{dialogue state}, we follow \citet{hosseini2020simple} to place its slot-value pairs sequentially.

\section{Input and Output Examples for Each Task}
\label{app:examples}
\subsection{Spider}

\textbf{Structured Input:} 

\begin{lstlisting}
| concert_singer | stadium : stadium_id , location , name , capacity , highest , lowest , average | singer : singer_id , name , country , song_name , song_release_year , age , is_male | concert : concert_id , concert_name , theme , stadium_id , year | singer_in_concert : concert_id , singer_id
\end{lstlisting}
\textbf{Request Input:} 
\begin{lstlisting}
How many singers do we have?
\end{lstlisting}
\textbf{Sequence Output:} 
\begin{lstlisting}
select count(*) from singer
\end{lstlisting}

\subsection{GRAILQA}

\textbf{Structured Input:} 

\begin{lstlisting}
soviet red army: m.06dr9 | organization.organization.founders government.governmental_body.jurisdiction organization.organization_founder.organizations_founded military.military_service.military_person government.political_party_tenure government.national_anthem_of_a_country visual_art.art_subject.artwork_on_the_subject government.government_agency government.governmental_jurisdiction.government people.deceased_person.place_of_burial people.deceased_person.date_of_death people.person.children people.person.parents people.person.height_meters government.government_position_held.office_holder government.government people.person people.person.sibling_s people.person.quotations people.person.gender
\end{lstlisting}
\textbf{Request Input:} 
\begin{lstlisting}
the person who founded the soviet red army also founded what government agency?
\end{lstlisting}
\textbf{Sequence Output:} 
\begin{lstlisting}
(AND government.government_agency (JOIN organization.organization.founders (JOIN (R organization.organization.founders) m.06dr9)))
\end{lstlisting}

\subsection{CompWebQ}
\textbf{Structured Input:} 
\begin{lstlisting}
Liam Hemsworth celebrities.celebrity.sexual_relationships..celebrities.romantic_relationship.celebrity Liam Hemsworth | Liam Hemsworth type.object.type tv.tv_actor | Liam Hemsworth film.actor.film..film.performance.character Billy The Kid | Liam Hemsworth film.actor.film..film.performance.film The Hunger Games | 
(omitted to save space)
| Liam Hemsworth type.object.type celebrities.celebrity | Cut Bank film.film.cinematography Ben Richardson | Cut Bank film.film.language English Language | Cut Bank film.film.edited_by Carol Littleton | Liam Hemsworth film.actor.film..film.performance.film The Hunger Games: Mockingjay, Part 1 | Liam Hemsworth film.actor.film..film.performance.film Timeless
\end{lstlisting}
\textbf{Request Input:} 
\begin{lstlisting}
What movie was produced by Dan Cohen and features Liam Hemsworth as an actor?
\end{lstlisting}
\textbf{Sequence Output:} 
\begin{lstlisting}
Cut Bank
\end{lstlisting}

\subsection{WebQsp}
\textbf{Structured Input:} 
\begin{lstlisting}
Spain: m.06mkj | m.06mkj location.location.contains m.0g3qgy | m.06mkj location.location.contains m.02qf5mh | m.0j5_3sv government.government_position_held.office_position_or_title m.0j5_3sz | m.06mkj location.location.contains m.02zb43k | 
(omitted to save space)
| m.06mkj government.governmental_jurisdiction.governing_officials m.010wswjc | m.06mkj location.location.contains m.09k5hy | m.010wswjc government.government_position_held.office_position_or_title m.0j5_3sz | m.06mkj location.location.contains m.02z98t5 | m.06mkj location.location.contains m.03qcr60
\end{lstlisting}
\textbf{Request Input:} 
\begin{lstlisting}
what is the king of spain's name?
\end{lstlisting}
\textbf{Sequence Output:} 
\begin{lstlisting}
(JOIN (R government.government_position_held.office_holder) (AND (JOIN government.government_position_held.time_macro 2015^^http://www.w3.org/2001/XMLSchema#date) (AND (JOIN government.government_position_held.office_position_or_title m.0j5_3sz) (JOIN (R government.governmental_jurisdiction.governing_officials) m.06mkj))))
\end{lstlisting}


\subsection{MTOP}
\textbf{Structured Input:} 
\begin{lstlisting}
 IN:GET: MESSAGE, WEATHER, ALARM, INFO_RECIPES, STORIES_NEWS, REMINDER, RECIPES, EVENT, CALL_TIME, LIFE_EVENT, INFO_CONTACT, CONTACT, TIMER, REMINDER_DATE_TIME, AGE, SUNRISE, EMPLOYER, EDUCATION_TIME, JOB, AVAILABILITY, 
 (omitted to save space)
 IN:PREVIOUS: TRACK_MUSIC | IN:HOLD: CALL | IN:SKIP: TRACK_MUSIC | IN:LIKE: MUSIC | IN:RESTART: TIMER | IN:RESUME: TIMER, CALL, MUSIC | IN:MERGE: CALL | IN:REPLAY: MUSIC | IN:LOOP: MUSIC | IN:STOP: MUSIC, SHUFFLE_MUSIC | IN:UNLOOP: MUSIC | IN:CANCEL: MESSAGE, CALL | IN:REWIND: MUSIC | IN:REPEAT: ALL_MUSIC, ALL_OFF_MUSIC | IN:FAST: FORWARD_MUSIC | IN:DISLIKE: MUSIC | IN:DISPREFER:  | IN:HELP: REMINDER | IN:FOLLOW: MUSIC
\end{lstlisting}
\textbf{Request Input:} 
\begin{lstlisting}
call Nicholas and Natasha
\end{lstlisting}
\textbf{Sequence Output:} 
\begin{lstlisting}
[IN:CREATE_CALL [SL:CONTACT Nicholas ] [SL:CONTACT Natasha ] ]
\end{lstlisting}

\subsection{WikiSQL}
\textbf{Structured Input:} 
\begin{lstlisting}
col : player | no. | nationality | position | years in toronto | school/club team row 1 : antonio lang | 21 | united states | guard-forward | 1999-2000 | duke row 2 : voshon lenard | 2 | united states | guard | 2002-03 | minnesota row 3 : martin lewis | 32, 44 | united states | guard-forward | 1996-97 | butler cc (ks) row 4 : brad lohaus | 33 | united states | forward-center | 1996 | iowa row 5 : art long | 42 | united states | forward-center | 2002-03 | cincinnati row 6 : john long | 25 | united states | guard | 1996-97 | detroit row 7 : kyle lowry | 3 | united states | guard | 2012-present | villanova
\end{lstlisting}
\textbf{Request Input:} 
\begin{lstlisting}
what position does the player who played for butler cc (ks) play?
\end{lstlisting}
\textbf{Sequence Output:} 
\begin{lstlisting}
guard-forward
\end{lstlisting}

\subsection{WikiTableQuestions}
\textbf{Structured Input:} 
\begin{lstlisting}
col : team | county | wins | years won row 1 : greystones | wicklow | 1 | 2011 row 2 : ballymore eustace | kildare | 1 | 2010 row 3 : maynooth | kildare | 1 | 2009 row 4 : ballyroan abbey | laois | 1 | 2008 row 5 : fingal ravens | dublin | 1 | 2007 row 6 : confey | kildare | 1 | 2006 row 7 : crettyard | laois | 1 | 2005 row 8 : wolfe tones | meath | 1 | 2004 row 9 : dundalk gaels | louth | 1 | 2003
\end{lstlisting}
\textbf{Request Input:} 
\begin{lstlisting}
which team won previous to crettyard?
\end{lstlisting}
\textbf{Sequence Output:} 
\begin{lstlisting}
wolfe tones
\end{lstlisting}

\subsection{HybridQA}
\textbf{Structured Input:} 
\begin{lstlisting}
col : position | athlete | nationality | time row 1 : 1 | patrick makau musyoki | kenya | 2:03.38 row 2 : 2 | stephen kwelio chemlany | kenya | 2:07.55 row 3 : 3 | edwin kimaiyo | kenya | 2:09.50 row 4 : 4 | felix limo | kenya | 2:10.38 row 5 : 5 | scott overall | united kingdom | 2:10.55 row 6 : 6 | ricardo serrano | spain | 2:13.32 row 7 : 7 | pedro nimo | spain | 2:13.34 row 8 : 8 | simon munyutu | france | 2:14.20 row 9 : 9 | driss el himer | france | 2:14.46 row 10 : 10 | hendrick ramaala | south africa | 2:16.00passages: ricardo serrano (athlete): at the 2011 iaaf world cross country championships he was 89th overall . his marathon debut followed later that year and he was sixth at the 2011 berlin marathon with a time of 2:13.32 hours . | spain: with an area of 505,990 km2 ( 195,360 sq mi ) , spain is the largest country in southern europe , the second largest country in western europe and the european union , and the fourth largest country in the european continent . by population ( about 47 million ) , spain is the sixth largest in europe and the fifth in the european union . | 
\end{lstlisting}
\textbf{Request Input:} 
\begin{lstlisting}
what place was achieved by the person who finished the berlin marathon in 2:13.32 in 2011 the first time he competed in a marathon ?
\end{lstlisting}
\textbf{Sequence Output:} 
\begin{lstlisting}
sixth
\end{lstlisting}

\subsection{MultiModalQA}
\textbf{Structured Input:} 
\begin{lstlisting}
ben piazza | filmography col : year | title | role | notes row 1 : 1957 | a dangerous age | david |  row 2 : 1959 | the hanging tree | rune |  row 3 : 1962 | no exit | camarero |  row 4 : 1970 | tell me that you love me, junie moon | jesse |  row 5 : 1972 | the outside man | desk clerk |  row 6 : 1973 | the candy snatchers | avery |  row 7 : 1976 | the bad news bears | bob whitewood |  row 8 : 1977 | i never promised you a rose garden | jay blake |  row 9 : 1979 | nightwing | roger piggott |  row 10 : 1979 | the concorde ... airport '79 | associate | tv version, uncredited row 11 : 1980 | the blues brothers | father |  row 12 : 1982 | waltz across texas | bill wrather |  row 13 : 1985 | mask | mr. simms |  row 14 : 1988 | clean and sober | kramer |  row 15 : 1990 | rocky v | doctor | uncredited row 16 : 1991 | guilty by suspicion | darryl zanuck |
\end{lstlisting}
\textbf{Request Input:} 
\begin{lstlisting}
for which film did ben piazza play the role of mr. simms?
\end{lstlisting}
\textbf{Sequence Output:} 
\begin{lstlisting}
mask
\end{lstlisting}

\subsection{FeTaQA}
\textbf{Structured Input:} 
\begin{lstlisting}
andy karl | awards and nominations col : year | award | category | work | result row 1 : 2013 | drama desk award | outstanding featured actor in a musical | the mystery of edwin drood | nominated row 2 : 2013 | broadway.com audience awards | favorite onstage pair (with jessie mueller) | the mystery of edwin drood | nominated row 3 : 2014 |
(omitted to save space)
| groundhog day | nominated row 15 : 2017 | tony award | best actor in a musical | groundhog day | nominated row 16 : 2017 | drama desk award | outstanding actor in a musical | groundhog day | won row 17 : 2017 | drama league award | distinguished performance | groundhog day | nominated row 18 : 2017 | outer critics circle award | outstanding actor in a musical | groundhog day | won
\end{lstlisting}
\textbf{Request Input:} 
\begin{lstlisting}
when did andy karl win the olivier award and for which of his work?
\end{lstlisting}
\textbf{Sequence Output:} 
\begin{lstlisting}
for his performance in groundhog day, andy karl received the 2017 olivier award for best actor in a musical.
\end{lstlisting}

\subsection{DART}
\textbf{Structured Input:} 
\begin{lstlisting}
Mars Hill College : joined : 1973 | Mars Hill College : location : Mars Hill, North Carolina
\end{lstlisting}
\textbf{Request Input:} 
\begin{lstlisting}
None
\end{lstlisting}
\textbf{Sequence Output:} 
\begin{lstlisting}
A school from Mars Hill, North Carolina, joined in 1973.
\end{lstlisting}

\subsection{ToTTo}
\textbf{Structured Input:} 
\begin{lstlisting}
<page_title> List of Governors of South Carolina </page_title> <section_title> Governors under the Constitution of 1868 </section_title> <table> <cell> 76 <col_header> # </col_header> <col_header> 74 </col_header> <col_header> 75 </col_header> </cell> <cell> Daniel Henry Chamberlain <col_header> Governor </col_header> <row_header> 76 </row_header> </cell> <cell> December 1, 1874 <col_header> Took Office </col_header> <row_header> 76 </row_header> </cell> </table>
\end{lstlisting}
\textbf{Request Input:} 
\begin{lstlisting}
None
\end{lstlisting}
\textbf{Sequence Output:} 
\begin{lstlisting}
Daniel Henry Chamberlain was the 76th Governor of South Carolina from 1874.
\end{lstlisting}

\subsection{MultiWoZ2.1}
\textbf{Structured Input:} 
\begin{lstlisting}
hotel-pricerange: cheap, dontcare, expensive, moderate; hotel-type: guesthouse, hotel; hotel-parking: dontcare, free, no, yes; hotel-book day: friday, monday, saturday, sunday, thursday, tuesday, wednesday; hotel-book people: 1, 2, 3, 4, 5, 6, 7, 8; hotel-book stay: 1, 2, 3, 4, 5, 6, 7, 8; hotel-area: centre, dontcare, east, north, south, west; hotel-stars: 0, 1, 2, 3, 4, 5, dontcare; hotel-internet: dontcare, no, yes; hotel-name: none; train-destination: none; train-day: dontcare, friday, monday, saturday, sunday, thursday, tuesday, wednesday; train-departure: none; train-arriveby: none; train-book people: 0, 1, 10, 15, 2, 3, 4, 5, 6, 7, 8, 9; taxi-destination: none; taxi-departure: none; taxi-leaveat: none; train-leaveat: none; attraction-area: cambridge, centre, dontcare, east, north, south, west; restaurant-pricerange: cheap, dontcare, expensive, moderate; restaurant-area: centre, east, north, south, west; restaurant-food: none; attraction-name: none; restaurant-name: none; attraction-type: architecture, boat, church, cinema, college, concerthall, entertainment, hotspot, multiple sports, museum, nightclub, park, special, swimmingpool, theatre; restaurant-book day: friday, monday, saturday, sunday, thursday, tuesday, wednesday; restaurant-book people: 1, 2, 3, 4, 5, 6, 7, 8; restaurant-book time: none; taxi-arriveby: none; 
\end{lstlisting}
\textbf{Request Input:} 
\begin{lstlisting}
i am looking to book a train that is leaving from cambridge to bishops stortford on friday .
\end{lstlisting}
\textbf{Context:} \begin{lstlisting}
booking was successful . reference number is : bmukptg6 . can i help you with anything else today ? | friday and can you book it for me and get a reference number ? | what day will you be staying ? | could you book the wartworth for 1 night , 1 person ? | if you would like something cheap , i recommend the allenbell . for something moderate -ly priced , i would recommend the warkworth house . | that does not matter as long as it has free wifi and parking . | i can help you with that . what is your price range ? | i need to book a hotel in the east that has 4 stars . | none
\end{lstlisting}
\textbf{Sequence Output:} 
\begin{lstlisting}
hotel pricerange none, hotel type none, hotel parking yes, hotel book day friday, hotel book people 1, hotel book stay 1, hotel area east, hotel stars 4, hotel internet yes, hotel name wartworth, train destination bishops stortford, train day friday, train departure cambridge, train arriveby none, train book people none, taxi destination none, taxi departure none, taxi leaveat none, train leaveat none, attraction area none, restaurant pricerange none, restaurant area none, restaurant food none, attraction name none, restaurant name none, attraction type none, restaurant book day none, restaurant book people none, restaurant book time none, taxi arriveby none
\end{lstlisting}

\subsection{KVRET}
\textbf{Structured Input:} 
\begin{lstlisting}
col : poi | poi_type | address | distance | traffic_info row 1 : sigona farmers market | grocery store | 638 amherst st | 3 miles | car collision nearby row 2 : cafe venetia | coffee or tea place | 269 alger dr | 1 miles | car collision nearby row 3 : 5672 barringer street | certain address | 5672 barringer street | 5 miles | no traffic row 4 : valero | gas station | 200 alester ave | 2 miles | road block nearby row 5 : stanford childrens health | hospital | 899 ames ct | 5 miles | moderate traffic row 6 : palo alto garage r | parking garage | 481 amaranta ave | 1 miles | moderate traffic row 7 : teavana | coffee or tea place | 145 amherst st | 1 miles | road block nearby row 8 : willows market | grocery store | 409 bollard st | 5 miles | no traffic
\end{lstlisting}
\textbf{Request Input:} 
\begin{lstlisting}
ok, please give me directions via a route that avoids all heavy_traffic.
\end{lstlisting}
\textbf{Context:} \begin{lstlisting}
valero is located at 200_alester_ave. | address to the gas_station
\end{lstlisting}
\textbf{Sequence Output:} 
\begin{lstlisting}
since there is a road_block_nearby, i found another route for you and i sent it on your screen
\end{lstlisting}

\subsection{SParC}
\textbf{Structured Input:} 
\begin{lstlisting}
| flight_2 | airlines : uid , airline , abbreviation , country | airports : city , airportcode , airportname , country , countryabbrev | flights : airline , flightno , sourceairport , destairport
\end{lstlisting}
\textbf{Request Input:} 
\begin{lstlisting}
What is the country corresponding it?
\end{lstlisting}
\textbf{Context:} \begin{lstlisting}
Of these, which is Jetblue Airways? | What are all the airlines?
\end{lstlisting}
\textbf{Sequence Output:} 
\begin{lstlisting}
select country from airlines where airline = "JetBlue Airways"
\end{lstlisting}

\subsection{CoSQL}
\textbf{Structured Input:} 
\begin{lstlisting}
| car_1 | continents : contid , continent | countries : countryid , countryname , continent | car_makers : id , maker , fullname , country | model_list : modelid , maker , model | car_names : makeid , model , make | cars_data : id , mpg , cylinders , edispl , horsepower , weight , accelerate , year
\end{lstlisting}
\textbf{Request Input:} 
\begin{lstlisting}
Which models are the most popular?
\end{lstlisting}
\textbf{Context:} \begin{lstlisting}
How many car models are produced by each maker? List the count and the maker full name.
\end{lstlisting}
\textbf{Sequence Output:} 
\begin{lstlisting}
select t1.model from model_list as t1 join car_makers as t2 on t1.maker = t2.id group by t2.id order by count ( * ) desc limit 1
\end{lstlisting}

\subsection{SQA}
\textbf{Structured Input:} 
\begin{lstlisting}
col : date | number | type | captain | casualties | sunk by row 1 : 4 may 1943 | u-209 | viic | heinrich brodda | 46 | catalina flying-boat of no. 5 squadron row 2 : 5 may 1943 | u-638 | viic | oskar staudinger | 44 | hms sunflower row 3 : 5 may 1943 | u-531 | ixc/40 | herbert neckel | 54 | hms vidette row 4 : 6 may 1943 | u-192 | ixc/40 | werner happe | 55 | hms loosestrife row 5 : 6 may 1943 | u-125 | ixc | ulrich folkers | 54 | "hms oribi |  hms snowflake" row 6 : 6 may 1943 | u-630 | viic | werner winkler | 47 | hms vidette row 7 : 6 may 1943 | u-438 | viic | heinrich hensohn | 48 | hms pelican
\end{lstlisting}
\textbf{Request Input:} 
\begin{lstlisting}
which captain was not oskar staudinger? 
\end{lstlisting}
\textbf{Context:} \begin{lstlisting}
who were the captains of those boats? | what boats were lost on may 5?
\end{lstlisting}
\textbf{Sequence Output:} 
\begin{lstlisting}
herbert neckel
\end{lstlisting}

\subsection{TabFact}
\textbf{Structured Input:} 
\begin{lstlisting}
col : round | clubs remaining | clubs involved | winners from previous round | new entries this round | leagues entering at this round row 1 : first round | 156 | 86 | none | 86 | tff third league & turkish regional amateur league row 2 : second round | 113 | 108 | 43 | 65 | s\u00fcper lig & tff first league & tff second league row 3 : third round | 59 | 54 | 54 | none | none row 4 : fourth round | 32 | 32 | 27 | 5 | s\u00fcper lig row 5 : fifth round | 16 | 16 | 16 | none | none row 6 : group stage | 8 | 8 | 8 | none | none row 7 : semi - finals | 4 | 4 | 4 | none | none row 8 : final | 2 | 2 | 2 | none | none
\end{lstlisting}
\textbf{Request Input:} 
\begin{lstlisting}
during the third round of the turkish cup , there be no new entry during that stage
\end{lstlisting}
\textbf{Sequence Output:} 
\begin{lstlisting}
entailed
\end{lstlisting}

\subsection{FEVEROUS}
\textbf{Structured Input:} 
\begin{lstlisting}
col : no. | title | narrator | aired between | original air date | us viewers row 1 : 1 | "magic is coming" | giancarlo esposito | "a land without magic" "broken" | september 30, 2012 (2012-09-30) | 6.04 row 2 : 2 | "the price of magic" | alan dale | "selfless, brave and true" "lacey" | april 14, 2013 (2013-04-14) | 5.17 row 3 : 3 | "journey to neverland" | alfred molina | "and straight on 'til morning" "the heart of the | 
(omitted to save space)
| "dark swan rises: a once upon a time fan | howard parker | "operation mongoose, part 2" "the | september 27, 2015 (2015-09-27) | 3.20 row 8 : 8 | "evil reigns once more" | howard parker | "an untold story" "the savior" | september 25, 2016 (2016-09-25) | 2.86 row 9 : 9 | "the final battle begins" | howard parker | "the song in your heart" "the final battle" | may 14, 2017 (2017-05-14) | 2.82
\end{lstlisting}
\textbf{Request Input:} 
\begin{lstlisting}
list of once upon a time (debuted october 23, 2011) episodes consists of 10 specials, one of the narrators alan dale.
\end{lstlisting}
\textbf{Sequence Output:} 
\begin{lstlisting}
refuted
\end{lstlisting}

\subsection{SQL2Text}
\textbf{Structured Input:} 
\begin{lstlisting}
None
\end{lstlisting}
\textbf{Request Input:} 
\begin{lstlisting}
SELECT Fname FROM FACULTY WHERE Rank  =  "Professor" ORDER BY Fname
\end{lstlisting}
\textbf{Sequence Output:} 
\begin{lstlisting}
What are the first names for all faculty professors, ordered by first name?
\end{lstlisting}

\subsection{Logic2Text}
\textbf{Structured Input:} 
\begin{lstlisting}
Caption: list of tallest structures in the world - 300 to 400 metres | planned pinnacle height, expected year of completion, structural type, country, town
\end{lstlisting}
\textbf{Request Input:} 
\begin{lstlisting}
eq { count { filter_eq { all_rows ; expected year of completion ; 2006 } } ; 3 } = true
\end{lstlisting}
\textbf{Sequence Output:} 
\begin{lstlisting}
three of the tallest structures will be completed in 2006 .
\end{lstlisting}

\section{Case study}
\label{app:case-study}
\subsection{Spider case}
\subsubsection{Case 1:}
\textbf{Structured Input:} 
\begin{lstlisting}
| concert_singer | stadium : stadium_id , location , name , capacity , highest , lowest , average | singer : singer_id , name , country , song_name , song_release_year , age , is_male | concert : concert_id , concert_name , theme , stadium_id , year | singer_in_concert : concert_id , singer_id
\end{lstlisting}
\textbf{Request Input:} 
\begin{lstlisting}
What are the names of all stadiums that did not have a concert in 2014?
\end{lstlisting}
\textbf{Sequence Output:} 
\begin{lstlisting}
select name from stadium except select t2.name from concert as t1 join stadium as t2 on t1.stadium_id = t2.stadium_id where t1.year = 2014
\end{lstlisting}
\textbf{T5-base prediction (incorrect):} 
\begin{lstlisting}
select name from stadium except select stadium_name from concert where year = 2014
\end{lstlisting}
\textbf{T5-large prediction (correct):} 
\begin{lstlisting}
select name from stadium except select t2.name from concert as t1 join stadium as t2 on t1.stadium_id = t2.stadium_id where t1.year = 2014
\end{lstlisting}
\textbf{T5-3B prediction (correct):} 
\begin{lstlisting}
select name from stadium except select t2.name from concert as t1 join stadium as t2 on t1.stadium_id = t2.stadium_id where t1.year = 2014
\end{lstlisting}
\subsubsection{Case 2:}
\textbf{Structured Input:} 
\begin{lstlisting}
| concert_singer | stadium : stadium_id , location , name , capacity , highest , lowest , average | singer : singer_id , name , country , song_name , song_release_year , age , is_male | concert : concert_id , concert_name , theme , stadium_id , year | singer_in_concert : concert_id , singer_id
\end{lstlisting}
\textbf{Request Input:} 
\begin{lstlisting}
What is the name and capacity for the stadium with highest average attendance?
\end{lstlisting}
\textbf{Sequence Output:} 
\begin{lstlisting}
select name, capacity from stadium order by average desc limit 1
\end{lstlisting}
\textbf{T5-base prediction (incorrect):} 
\begin{lstlisting}
select name, capacity from stadium order by avg(amount) desc limit 1
\end{lstlisting}
\textbf{T5-large prediction (correct):} 
\begin{lstlisting}
select name, capacity from stadium order by average desc limit 1
\end{lstlisting}
\textbf{T5-3B prediction (correct):} 
\begin{lstlisting}
select name, capacity from stadium order by average desc limit 1
\end{lstlisting}
\subsubsection{Case 3:}
\textbf{Structured Input:} 
\begin{lstlisting}
| pets_1 | student : stuid , lname , fname , age , sex , major , advisor , city_code | has_pet : stuid , petid | pets : petid , pettype ( cat , dog ) , pet_age , weight
\end{lstlisting}
\textbf{Request Input:} 
\begin{lstlisting}
Find the first name of students who have cat or dog pet.
\end{lstlisting}
\textbf{Sequence Output:} 
\begin{lstlisting}
select distinct t1.fname from student as t1 join has_pet as t2 on t1.stuid = t2.stuid join pets as t3 on t3.petid = t2.petid where t3.pettype = 'cat' or t3.pettype = 'dog'
\end{lstlisting}
\textbf{T5-base prediction (incorrect):} 
\begin{lstlisting}
select t1.fname from student as t1 join has_pet as t2 on t1.stuid = t2.stuid where t2.pettype = "cat" or t2.pettype = "dog"
\end{lstlisting}
\textbf{T5-large prediction (incorrect):} 
\begin{lstlisting}
select t1.fname from student as t1 join has_pet as t2 on t1.stuid = t2.stuid where t2.pettype = "cat" or t2.pettype = "dog"
\end{lstlisting}
\textbf{T5-3B prediction (correct):} 
\begin{lstlisting}
select t1.fname from student as t1 join has_pet as t2 on t1.stuid = t2.stuid join pets as t3 on t2.petid = t3.petid where t3.pettype = "cat" or t3.pettype = "dog"
\end{lstlisting}

\subsection{FeTaQA case}
\label{app:fetaqa_case}
\textbf{Structured Input:} 
\begin{lstlisting}
te aroha (new zealand electorate) | 1890 election col : party | party | candidate | votes | % | [non utf-8 token] row 1 : - | independent | william shepherd allen | 786 | 56.34 | - row 2 : - | liberal | william fraser | 609 | 43.65 | - row 3 : majority | majority | majority | 175 | 12.54 | - row 4 : turnout | turnout | turnout | 1,395 | 48.60 | - row 5 : registered electors | registered electors | registered electors | 2,870 | - | -
\end{lstlisting}
\textbf{Request Input:} 
\begin{lstlisting}
what were the voting results in the te aroha election by the candidate?
\end{lstlisting}
\textbf{Sequence Output:} 
\begin{lstlisting}
votes for allen and fraser were 786 and 609 respectively.
\end{lstlisting}
\textbf{T5-base prediction (incorrect):} 
\begin{lstlisting}
the te aroha election was won by william shepherd allen with a majority of 175.
\end{lstlisting}
\textbf{T5-large prediction (incorrect):} 
\begin{lstlisting}
in the te aroha election, allen won with a majority of 175 votes.
\end{lstlisting}
\textbf{T5-3B prediction (correct):} 
\begin{lstlisting}
in the 1890 te aroha election, the result was a victory for the independent candidate, william shepherd allen, who received 786 votes against 609 for the liberal candidate, william fraser.
\end{lstlisting}

\subsection{KVRET case}
\subsubsection{Case 1:}
\label{app:kvret_case1}
\textbf{Structured Input:} 
\begin{lstlisting}
col : location | monday | tuesday | wednesday | thursday | friday | saturday | sunday | today row 1 : danville | clear skies, low of 30f, high of 50f | cloudy, low of 20f, high of 40f | raining, low of 80f, high of 100f | dew, low of 80f, high of 90f | clear skies, low of 70f, high of 80f | frost, low of 40f, high of 60f | overcast, low of 20f, high of 30f | monday row 2 :... row 6 : alhambra | windy, low of 90f, high of 100f | hail, low of 90f, high of 100f | clear skies, low of 80f, high of 90f | rain, low of 70f, high of 80f | stormy, low of 70f, high of 80f | rain, low of 60f, high of 70f | hail, low of 40f, high of 60f | monday ...
\end{lstlisting}
\textbf{Request Input:} 
\begin{lstlisting}
alhambra please. 
\end{lstlisting}
\textbf{Context Input:} 
\begin{lstlisting}
what city are you inquiring about? | is it going to be raining this weekend
\end{lstlisting}
\textbf{Sequence Output:} 
\begin{lstlisting}
it will be raining on saturday and hailing on sunday in alhambra.
\end{lstlisting}
\textbf{T5-base prediction (correct):} 
\begin{lstlisting}
it will be raining on saturday and hailing on sunday in alhambra.
\end{lstlisting}
\textbf{T5-large prediction (incorrect):} 
\begin{lstlisting}
it will not rain in alhambra this weekend.
\end{lstlisting}
\textbf{T5-3B prediction (incorrect):} 
\begin{lstlisting}
there will be no rain this weekend in alhambra.
\end{lstlisting}
\subsubsection{Case 2:}
\label{app:kvret_case2}
\textbf{Structured Input:} 
\begin{lstlisting}
col : event | time | date | room | agenda | party row 1 : tennis activity | 5pm | wednesday | - | - | mother row 2 : taking medicine | 9am | the 3rd | - | - | sister row 3 : swimming activity | 11am | friday | - | - | father row 4 : conference | 6pm | the 8th | 100 conference room | go over budget | infrastructure team row 5 : meeting | 4pm | the 10th | conference room 102 | go over quarterly report | infrastructure team row 6 : dentist appointment | 10am | the 13th | - | - | jeff row 7 : dentist appointment | 11am | the 5th | - | - | jeff row 8 : lab appointment | 2pm | the 4th | - | - | sister
\end{lstlisting}
\textbf{Request Input:} 
\begin{lstlisting}
when is the date and time of my dentist_appointment?
\end{lstlisting}
\textbf{Context Input:} 
\begin{lstlisting}
None 
\end{lstlisting}
\textbf{Sequence Output:} 
\begin{lstlisting}
your dentist_appointment is at 10am on the_13th.
\end{lstlisting}
\textbf{T5-base prediction (correct):} 
\begin{lstlisting}
you have two dentist_appointments scheduled. one on the_5th at 11am and another on the_13th at 10am.
\end{lstlisting}
\textbf{T5-large prediction (correct):} 
\begin{lstlisting}
your dentist_appointment is on the_13th at 10 am.
\end{lstlisting}
\textbf{T5-3B prediction (incorrect):} 
\begin{lstlisting}
you have three dentist_appointments scheduled. one on the_5th at 11am, another on the_13th at 10am, and the last is on the_5th at 11am.
\end{lstlisting}

\subsection{SQL2Text case}
\label{app:sql2text_case}
\textbf{Query: }
\begin{lstlisting}
SELECT company ,  main_industry FROM company WHERE headquarters != 'USA'
\end{lstlisting}
\textbf{Sequence Out: }
\begin{lstlisting}
Show the company name and the main industry for all companies whose headquarters are not from USA.
\end{lstlisting}
\textbf{T5-base prediction (incorrect):} 
\begin{lstlisting}
What are the companies and main industries of companies whose headquarters are not "USA"?
\end{lstlisting}
\textbf{T5-large prediction (correct):} 
\begin{lstlisting}
What are the companies and main industries of the companies whose headquarters are not located in the United States?
\end{lstlisting}
\textbf{T5-3B prediction (incorrect):} 
\begin{lstlisting}
What are the main industries of companies whose headquarters are not in the USA?
\end{lstlisting}

\subsection{DART case}
\label{app:dart_case}
\textbf{Structured Input:} 
\begin{lstlisting}
Cotto : eattype : coffee shop | Cotto : food : English | Cotto : pricerange : more than [non utf-8 token]30 | Cotto : customer rating : high | Cotto : area : city centre | Cotto : near : The Portland Arms
\end{lstlisting}
\textbf{Sequence Output:} 
\begin{lstlisting}
The Cotto is a highly rated English coffee shop in the city centre near The Portland Arms, and prices range more than [non utf-8 token]30.
\end{lstlisting}
\textbf{T5-base prediction (correct):} 
\begin{lstlisting}
Cotto is a coffee shop providing English food in the more than [non utf-8 token]30 price range. It is located in the city centre. It is near The Portland Arms. Its customer rating is high.
\end{lstlisting}
\textbf{T5-large prediction (correct):} 
\begin{lstlisting}
Cotto is a coffee shop providing English food in the more than [non utf-8 token]30 price range. It is located in the city centre. It is near The Portland Arms. Its customer rating is high.
\end{lstlisting}
\textbf{T5-3B prediction (correct):} 
\begin{lstlisting}
Cotto is a coffee shop located in the city centre near The Portland Arms. It serves English food with a price range of more than [non utf-8 token]30 and has a high customer rating.
\end{lstlisting}

\subsection{Logic2Text case}
\label{app:logic2text_case}
\textbf{Structured Input:} 
\begin{lstlisting}
Caption: 2008 san diego chargers season | week, date, time, opponent, result, game site, nfl recap, record
\end{lstlisting}
\textbf{Request Input:} 
\begin{lstlisting}
eq { count { filter_eq { filter_eq { all_rows ; game site ; qualcomm stadium } ; time ; 5:15 pm } } ; 3 } = true
\end{lstlisting}
\textbf{Sequence Output:} 
\begin{lstlisting}
in the 2008 san diego chargers season , among the games that were played in qualcomm stadium , 3 of them started at 5:15 pm .
\end{lstlisting}
\textbf{T5-base prediction (incorrect):} 
\begin{lstlisting}
in the 2008 san diego chargers season, when the game was at qualcomm stadium, there were three times the time was 5:15 pm.
\end{lstlisting}
\textbf{T5-large prediction (incorrect):} 
\begin{lstlisting}
in the 2008 san diego chargers season, when the game was at qualcomm stadium, there were 3 times the time was 5:15 pm.
\end{lstlisting}
\textbf{T5-3B prediction (correct):} 
\begin{lstlisting}
in the 2008 san diego chargers season, among the games played at qualcomm stadium, 3 of them started at 5:15 pm.
\end{lstlisting}

\subsection{ToTTo case}
\label{app:totto_case}
\begin{figure}[!t]
    \centering
    \includegraphics[width=0.75\linewidth]{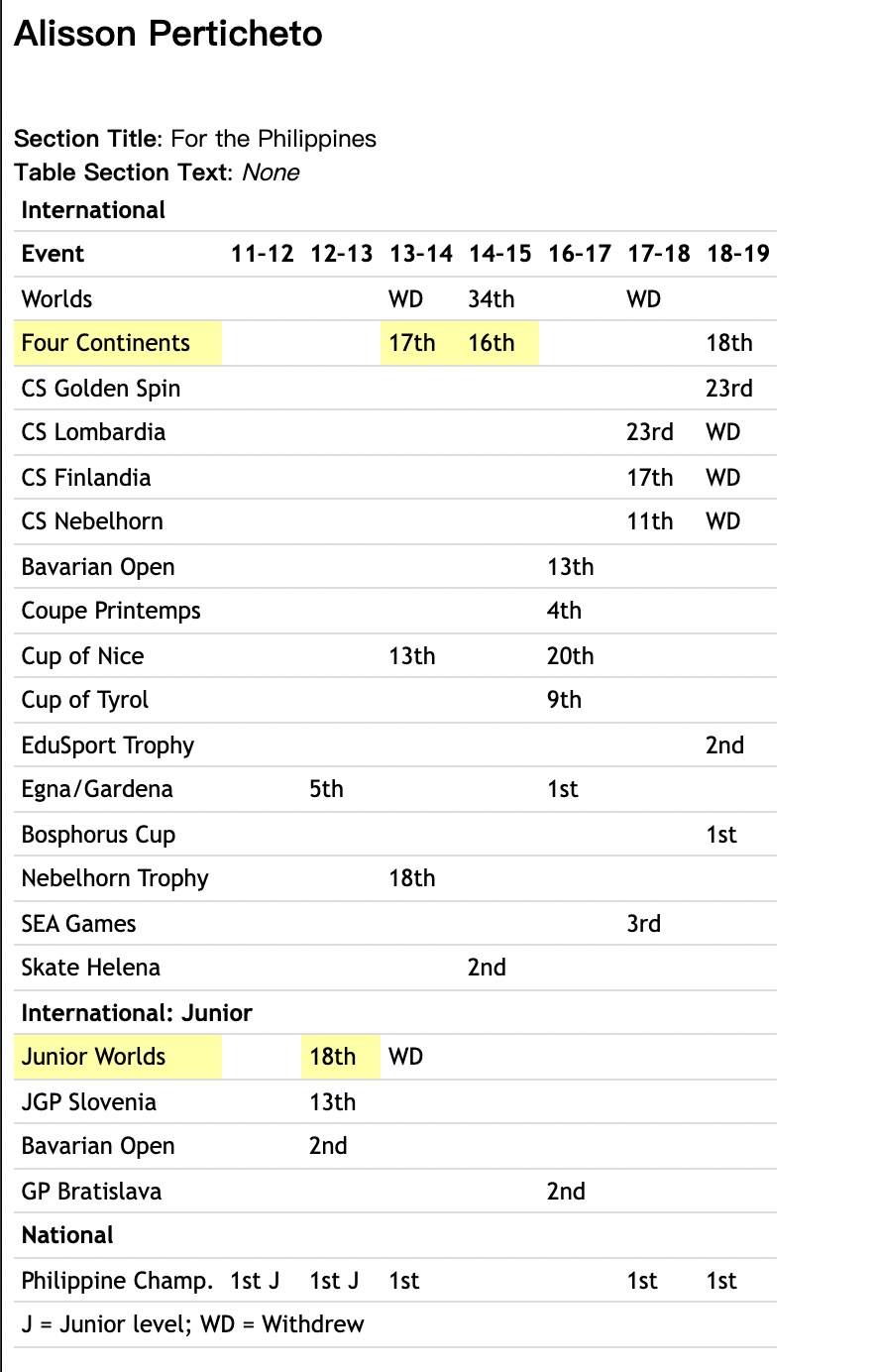}
    \caption{Visualized highted table for ToTTo case 1. }
\label{fig:totto_case1}
\end{figure}

\textbf{Structured Input:} 
See Figure~\ref{fig:totto_case1}.
\begin{lstlisting}
\end{lstlisting}
\textbf{Sequence Output:} 
\begin{lstlisting}
Alisson Perticheto placed 18th at the 2013 Junior Worlds, 17th at the 2014 Four Continents and 16th at the 2015 Four Continents.
\end{lstlisting}
\textbf{T5-base prediction (incorrect):} 
\begin{lstlisting}
Alisson Perticheto finished 18th at the Junior Worlds and 17th at the Four Continents.
\end{lstlisting}
\textbf{T5-large prediction (incorrect):} 
\begin{lstlisting}
Alisson Perticheto placed 17th at the 2014 Four Continents and 16th at the 2015 Junior Worlds.
\end{lstlisting}
\textbf{T5-3B prediction (correct):} 
\begin{lstlisting}
Alisson Perticheto finished 17th at the 2014 Four Continents, 16th at the 2015 Four Continents, and 18th at the 2013 Junior Worlds.
\end{lstlisting}

\section{Natural Language Template Examples}
\label{app:examples_template}
\subsection{Spider Template}

\textbf{Overall Description Template:} 
\begin{lstlisting}
{db id} contains tables such as {table1 name}, {table2 name}
\end{lstlisting}
\textbf{Primary Key Template:}
\begin{lstlisting}
{primary key} is the primary key.
\end{lstlisting}
\textbf{Table Description Template:}
\begin{lstlisting}
Table {table name} has column such as {column 1 name}, {column 2 name}, ...
\end{lstlisting}
\textbf{Foreign Keys Description Template:}
\begin{lstlisting}
The {column1 name} of {table 1} is the foreign key of {column2 name} of {table 2}
\end{lstlisting}

\subsection{TabFact Template}

\textbf{Template Examples:}

\begin{lstlisting}
Table 1-24143253-5:
{name} lost his spouse {deceased spouse} to {cause of death} on {date of spouses death} after {length of marriage} of marriage; they had {children together} together; he is currently {current marital status}
\end{lstlisting}

\begin{lstlisting}
Table 2-14978398-2:
The {version} of song Comme j'ai mal has a length of {length} in album {album} remixed by {remixed by} in year {year}
\end{lstlisting}

\begin{lstlisting}
Table 1-15187735-12:
On {date} in 1936 VFL Season, the home team {home team} and away team {away team} had a game at venue {venue} with a crowd of {crowd}; the home team score is {home team score} and the away team score is {away team score}
\end{lstlisting}

\subsection{WikiSQL Template}

\textbf{Template Example:}

\begin{lstlisting}
Table 1-14240688-1:
in {year} were in division {division}, {league} ranked {regular season}, made it to {playoffs} of the playoffs, made it to <{open cup}> in the open cup, and kept an average attendance of {avg attendance}
\end{lstlisting}

\begin{lstlisting}
Table 2-12997882-1:
On {date} in 2008 European Figure Skating, the home team {home team} and away team {away team} had a game at venue {venue} with a crowd of {crowd}; the home team score is {home team score} and the away team score is {away team score}
\end{lstlisting}

\begin{lstlisting}
Table 1-13740746-1:
Episode number {ep no} of gerry anderson 's new captain scarlet with a title of {title} is directed by {director} and written by {written by}; its original air date is {original air date}; the production number is {production no}
\end{lstlisting}

\end{document}